\documentclass[pdflatex,sn-mathphys]{sn-jnl}

\PassOptionsToPackage{dvipsnames}{xcolor}
\usepackage{amssymb}
\usepackage{makecell}
\usepackage{longtable}
\usepackage[labelformat=empty]{subcaption}
\usepackage{hhline}
\usepackage{lscape}
\usepackage{rotating,graphicx}
\usepackage[dvipsnames]{xcolor}
\renewcommand\appendixname{Extended Data}

\jyear{2021}%

\theoremstyle{thmstyleone}%
\theoremstyle{thmstyletwo}%

\theoremstyle{thmstylethree}%

\begin{document}

\title[Revamping AI Models in Dermatology]{Revamping AI Models in Dermatology: Overcoming Critical Challenges for Enhanced Skin Lesion Diagnosis}


\author[1]{\fnm{Deval} \sur{Mehta}}\email{deval.mehta@monash.edu}

\author[2]{\fnm{Brigid} \sur{Betz-Stablein}}\email{b.betzstablein@uq.edu.au}

\author[3]{\fnm{Toan} \sur{D Nguyen}}\email{Toan.Nguyen@monash.edu}

\author[4]{\fnm{Yaniv} \sur{Gal}}\email{yaniv.gal@molemap.co.nz}

\author[4]{\fnm{Adrian} \sur{Bowling}}\email{adrian.bowling@molemap.co.nz}

\author[4]{\fnm{Martin} \sur{Haskett}}\email{mhaskett06@gmail.com}

\author[5]{\fnm{Maithili} \sur{Sashindranath}}\email{maithili.sashindranath@monash.edu}

\author[6]{\fnm{Paul} \sur{Bonnington}}\email{pvcri@research.uq.edu.au}

\author[7,8]{\fnm{Victoria} \sur{Mar}}\email{victoria.mar@monash.edu.au}

\author[2]{\fnm{H Peter} \sur{Soyer}}\email{p.soyer@uq.edu.au}

\author*[1]{\fnm{Zongyuan} \sur{Ge}}\email{Zongyuan.Ge@monash.edu}

\affil*[1]{\orgdiv{AIM for Health Lab, Faculty of IT}, \orgname{Monash University}, \orgaddress{\street{Wellington Rd}, \city{Melbourne}, \postcode{3800}, \state{Victoria}, \country{Australia}}}

\affil[2]{\orgdiv{Frazer Institute}, \orgname{The University of Queensland, Dermatology Research Centre}, \orgaddress{\street{St Lucia}, \city{Brisbane}, \postcode{4072}, \state{Queensland}, \country{Australia}}}

\affil[3]{\orgdiv{e-Research Center}, \orgname{Monash University}, \orgaddress{\street{Wellington Rd}, \city{Melbourne}, \postcode{3800}, \state{Victoria}, \country{Australia}}}

\affil[4]{\orgdiv{Molemap Pty. Ltd.}, \orgaddress{\street{60 Albert Rd}, \city{Melbourne}, \postcode{3205}, \state{Victoria}, \country{Australia}}}

\affil[5]{\orgdiv{Australian Centre for Blood Diseases}, \orgname{Monash University}, \orgaddress{\street{Wellington Rd}, \city{Melbourne}, \postcode{3800}, \state{Victoria}, \country{Australia}}}

\affil[6]{\orgname{The University of Queensland}, \orgaddress{\street{St Lucia}, \city{Brisbane}, \postcode{4072}, \state{Queensland}, \country{Australia}}}

\affil[7]{\orgdiv{School of Public Health and Preventive Medicine}, \orgname{Monash University}, \orgaddress{\street{Wellington Rd}, \city{Melbourne}, \postcode{3800}, \state{Victoria}, \country{Australia}}}

\affil[8]{\orgdiv{Victorian Melanoma Service}, \orgname{Alfred Health}, \orgaddress{\street{55 Commercial Rd}, \city{Melbourne}, \postcode{3004}, \state{Victoria}, \country{Australia}}}


\abstract{The surge in developing deep learning models for diagnosing skin lesions through image analysis is notable, yet their clinical \textcolor{black}{implementation} faces challenges. Current dermatology AI models have limitations: limited number of possible diagnostic outputs, lack of real-world testing on uncommon skin lesions, inability to detect out-of-distribution images, and over-reliance on dermoscopic images. To address these, we present an All-In-One \textbf{H}ierarchical-\textbf{O}ut of Distribution-\textbf{C}linical Triage (HOT) model. For a clinical image, our model generates three outputs: a hierarchical prediction, an alert for out-of-distribution images, and a recommendation for dermoscopy if clinical image alone is insufficient for diagnosis. When the recommendation is pursued, it integrates both clinical and dermoscopic images to deliver final diagnosis. Extensive experiments on a representative \textcolor{black}{cutaneous} lesion dataset demonstrate the effectiveness and synergy of each component within our framework. Our versatile model provides valuable decision support for lesion diagnosis and sets a promising precedent for medical AI applications.}




\maketitle

\definecolor{level1}{RGB}{255, 192, 0}
\definecolor{level2}{RGB}{126, 100, 158}
\definecolor{level3}{RGB}{192, 80, 70}
\definecolor{green}{RGB}{0, 255, 0}
\definecolor{red}{RGB}{255, 0, 0}
\definecolor{blue}{RGB}{0, 0, 255}

\section{Introduction}\label{sec1}

Skin diseases are a significant global health concern, ranking as the eighteenth leading cause of disability-adjusted life years worldwide. Providing high-quality care is crucial for alleviating the global burden of skin diseases~\cite{seth2017global}. However, the shortage of dermatologists~\cite{kimball2008us}, especially evident during the COVID-19 crisis, has led to their escalating consultation costs~\cite{feng2018comparison,webster2020virtual}. Recent advancements in skin imaging technology and Artificial Intelligence (AI) may assist in increasing the accessibility of dermatology expertise~\cite{liu2020deep}.
The evolution of skin imaging technology such as total body photographs~\cite{ji2021total} and dermoscopy ~\cite{kittler2021evolution}, combined with AI methodologies, has led to rapid progression in image-based skin lesion diagnosis. AI techniques, including deep learning and machine learning models, have demonstrated promising results~\cite{esteva2017dermatologist,haenssle2018man} and even outperformed dermatologists in certain tasks of skin cancer classification~\cite{brinker2019deep,maron2019systematic}.

\begin{figure}[t!]%
\centering
\includegraphics[width=0.87\textwidth]{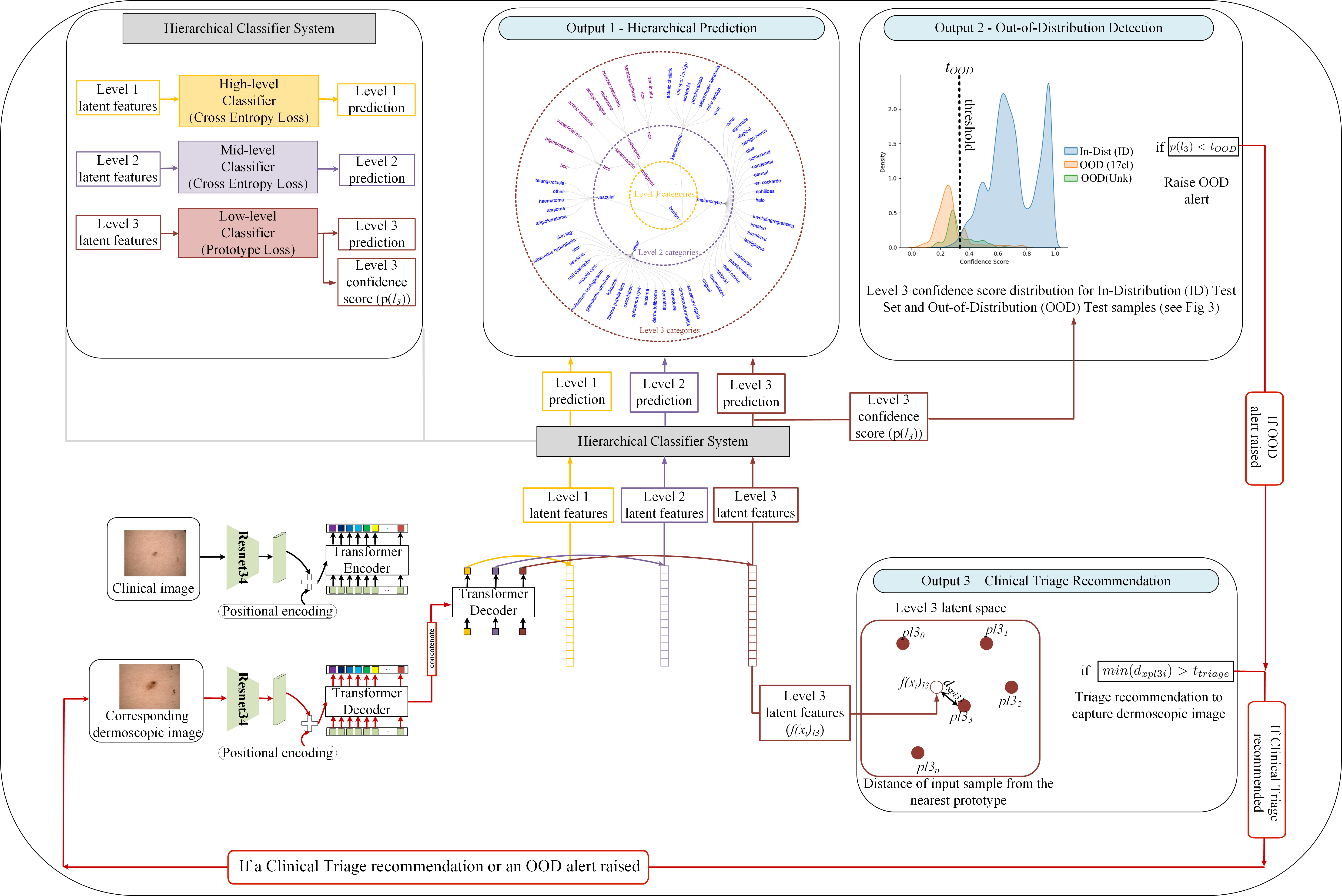}
\caption{Overall framework of our proposed versatile skin lesion diagnosis - \textbf{H}ierarchical-\textbf{O}ut of Distribution- Clinical \textbf{T}riage (HOT) model. When a single clinical skin photograph image of a lesion is analyzed, the HOT model gives three outputs - 1) Hierarchical classification of the skin lesion 2) An Out- of-Distribution (OOD) alert if it thinks the input test image (uncommon or rare skin condition) falls out of its learned distributions. 3) A clinical triage recommendation to capture an additional finer dermoscopic image of the \textcolor{black}{same lesion} for a more accurate diagnosis. If the clinical triage recommendation is considered (depicted by the \textcolor{black}{dark red clinical triage route}), the framework will then provide a more accurate diagnosis for the combined clinical image and dermoscopic images. \textcolor{black}{Our web demonstration with some sample images is available at \url{https://lesion.ai.cloud.monash.edu/}}}\label{fig1}
\end{figure}

Despite these AI advancements~\cite{cruz2013deep,codella2018skin,yuan2017automatic,okuboyejo2013automating,tschandl2019comparison,lopez2017skin,li2018skin,yap2018multimodal}, challenges remain in the adoption of AI for clinical decision-making~\cite{liopyris2022artificial}. We list the critical challenges associated with the dermatology AI models below. 
\begin{itemize}
    \item Existing AI models have typically produced a singular diagnosis for a skin lesion image, thus making it challenging for \textcolor{black}{clinicians} to trust the AI model's decision-making process~\cite{liu2020deep}. Contrary to this, clinicians generally provide a differential diagnosis.
    \item The evaluation of AI models has been limited to datasets~\cite{codella2018skin,tschandl2018ham10000,combalia2019bcn20000} that do not accurately represent real-world scenarios~\cite{daneshjou2022disparities}, lacking diversity \textcolor{black}{and range} in skin diseases. Thus, the reliability of dermatology AI models in detecting out-of-distribution (OOD) lesions is questionable.
    \item AI models primarily use dermoscopic images~\cite{sun2016benchmark}, despite clinical skin photographs often being sufficient or even superior for diagnosing many common skin lesions~\cite{han2018deep}. Dermoscopic imaging is time-consuming and requires specialized equipment, making it impractical outside of dermatology clinics~\cite{liu2020deep}, whereas clinical photographs of skin lesions are \textcolor{black}{more} easily obtained. This inconsistency raises a question of whether to rely solely on \textcolor{black}{an} AI model's diagnosis using a clinical lesion image or to obtain a dermoscopic image of the same lesion, leaving potential triage benefits unexplored~\cite{harkemanneevaluation,cantisani2022melanoma}.
\end{itemize}

Hence, an AI model offering a more nuanced diagnostic capability, such as hierarchical \textcolor{black}{akin to a} differential diagnosis in addition to identifying an OOD image, or providing a triage recommendation, would better align with the clinical thought process, thereby enhancing confidence in the AI model~\cite{du2020ai}.

We present an All-In-One framework called \textbf{H}ierarchical-\textbf{O}ut of Distribution- Clinical\textbf{T}riage (HOT) skin lesion model (illustrated in Fig~\ref{fig1}) as a versatile solution to address the challenges mentioned above. \textcolor{black}{Our} HOT model \textcolor{black}{comprises} of a CNN + Transformer Encoder-Decoder framework, incorporating mixup and prototype learning techniques. For a given clinical skin lesion image, it provides the below details:

\begin{itemize}
    \item A hierarchical classification consisting of a high-level (benign/malignant), mid-level (eight categories), and low-level (\textcolor{black}{44} lesion categories) predictions. This provides a more nuanced understanding compared to a single output diagnosis, particularly for challenging borderline cases such as differentiating benign naevi from melanoma based on the dermoscopic image~\cite{tognetti2021impact}.
    \item Whether the input is an OOD\footnote{In this study, the term `OOD image' refers to an uncommon skin condition or a category not included in the training dataset} image. If an OOD alert is raised, the input image should be treated as an outlier and its original prediction should not be trusted, which is vital for medical AI applications~\cite{zadorozhny2023out} to mitigate potential misleading of clinicians~\cite{tschandl2020human}.
    \item A triage recommendation on whether dermoscopy capture of the same skin lesion is necessary. Clinical images are easily captured and sufficient for AI diagnosis in many cases~\cite{brinker2018skin}. Our framework optimally decides to pursue a dermoscopic image acquisition only when required. If pursued, it integrates both the clinical and dermoscopic image to provide a more accurate diagnosis.
\end{itemize}

In summary, \textcolor{black}{our} HOT model offers a comprehensive \textcolor{black}{framework for diagnosing skin lesions} and \textcolor{black}{its integration has the potential to enhance} end-user confidence and \textcolor{black}{aid} in making \textcolor{black}{well-informed} clinical \textcolor{black}{diagnoses}~\cite{moshi2019evaluation}.


\section{Dataset \& Modules}

\subsection{Dataset}

\begin{figure}[h]%
\centering
\includegraphics[width=0.9\textwidth]{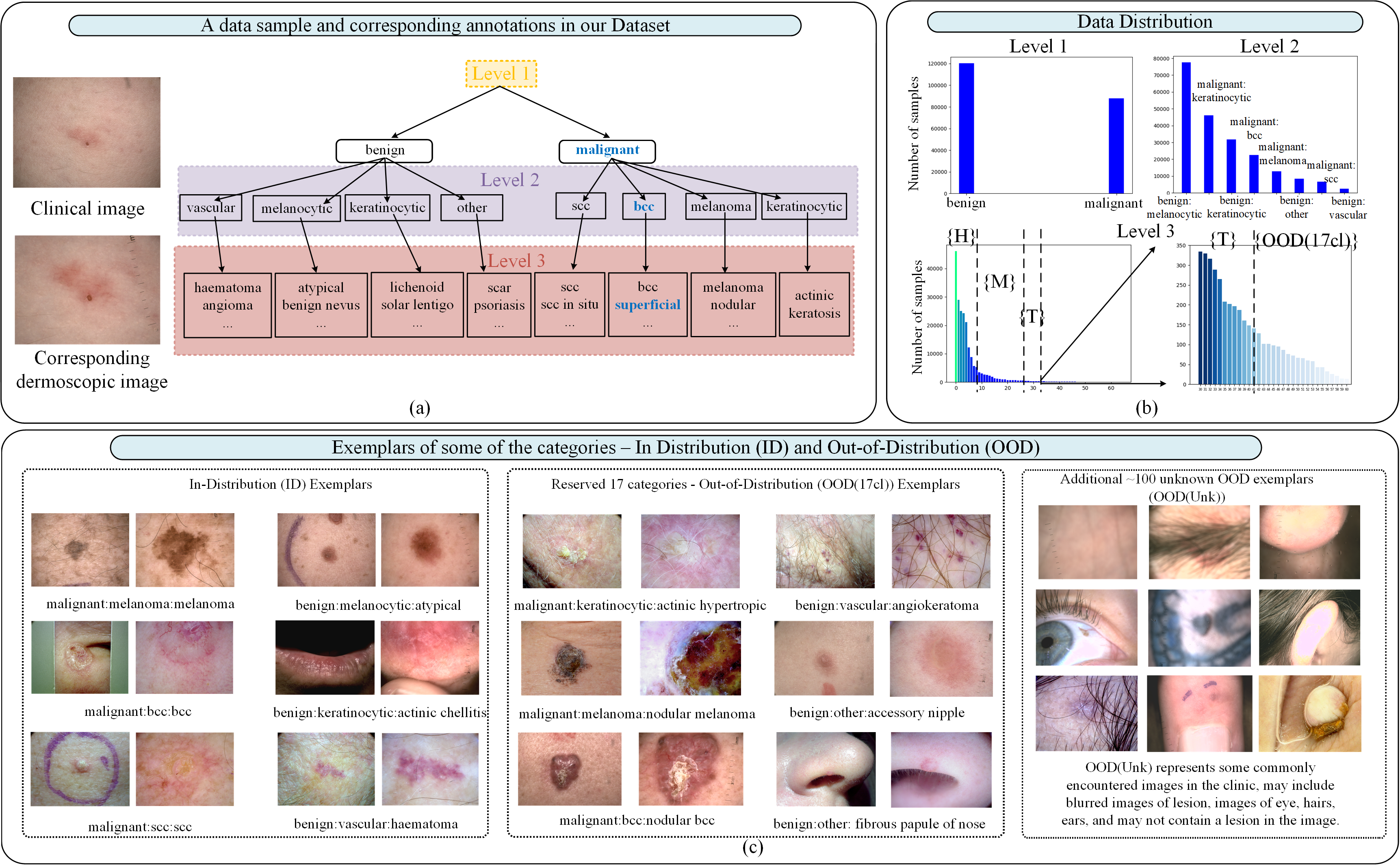}
\caption{Dataset exemplars and statistics.(a) An exemplar image (both clinical and dermoscopic) of malignant:bcc:superficial skin lesion. (b) Data samples distribution across the different categories in the dataset on all the three levels. Note the long-tailed and fine-grained distribution for level 3. (c) Exemplar images of selected skin lesions with their ground truth category, after separating the dataset into ID and OOD.}\label{fig2}
\end{figure}

We utilize the Molemap~\cite{mehta2022out} \textcolor{black}{dataset}, which is a highly representative skin lesion dataset exhibiting skewed and long-tailed \footnote{Long-tailed dataset corresponds to a scenario when there is a large class imbalance in number of images within the lesion types with the majority of the lesion types having less images compared to the minorty of lesion types with the most images} distribution, and is fine-grained in nature. \textcolor{black}{The dataset's imbalance and detailed granularity accurately mirrors the real-world conditions found in clinical settings worldwide}. Molemap contains 208,287 dermoscopic and clinical images, categorized into high-level (benign/malignant), mid-level (eight categories), and low-level (61 categories) hierarchy. \textcolor{black}{Fig~\ref{fig1} illustrates the hierarchical levels, while Fig~\ref{fig2}(b) demonstrates the data distribution among categories across levels 1, 2 and level 3 respectively.} An exemplar skin lesion with the diagnosis malignant:bcc:superficial and its clinical and dermoscopic images are presented in Fig~\ref{fig2}(a). For the skin lesion in Fig~\ref{fig2}(a), high-level (level 1) category is \textbf{malignant}, the mid-level category (level 2) is \textbf{bcc}, and the low-level category (level 3) is \textbf{superficial}. The consistent labeling system assigns a three-level label (level1:level2:level3) to each image in the Molemap dataset. Extension ~\ref{secDataCol} provides further details on the data collection standards, and train and test splits.

\subsubsection{Out-of-Distribution Settings}\label{ood-settings}

To build and test the OOD detection capability, we utilize the fine-grained hierarchical level 3, which consists of 61 categories. \textcolor{black}{Based on the sample distribution shown in Fig~\ref{fig2}(b)}, we split these 61 categories into - 44 In-Distribution (ID) categories and 17 reserved  Out-of-Distribution (OOD) categories\footnote{We used a cut-off of \textcolor{black}{100} images per category and the bottom-most 25\% percentile (based on the number of samples) of the total number of the categories to split them into ID and OOD.}. The hierarchical connection extends this split to level 1 and level 2 samples\footnote{Note that the number of ID categories for level 1 and level 2 remain the same - two categories for level 1 and eight categories for level 2.}. The specific \textcolor{black}{hierarchical levels amongst} the ID and OOD categories are also listed in Table~\ref{taba1} in the appendix. The 17 OOD categories of level 3 are denoted as OOD (17 cl).

To create a more realistic OOD test set, we enhance the split set of 17 reserved categories with $\sim$100 unusual images \textcolor{black}{which are commonly encountered} in clinics, such as blurred or obscured skin lesions. These images represent an additional category in the OOD set, labeled as OOD (Unk). Some sample images from both ID and OOD categories are presented in \textcolor{black}{Figure~\ref{fig2}(c)}. During training, our framework exclusively uses images from the ID categories only.

\subsubsection{Input Image Modality}

We evaluate our framework on three different input image settings to fairly compare the performance of \textcolor{black}{exclusively using clinical and dermoscopic images, and combining both of them together.}

\noindent \textbf{Clinical image} - In this setting, we utilize only the macroscopic clinical image, which represents overall features of the skin lesion and surrounding skin. Our framework analyzes and diagnoses the clinical image alone.

\noindent \textbf{Dermoscopic image} - This pertains to the dermoscopic image, revealing the finer features of the skin lesion. In this setting, only the dermoscopic image is supplied to our framework for analysis and diagnosis.

\noindent \textbf{Clinical+Dermoscopic image} - This setting combines both the clinical and dermoscopic images. \textcolor{black}{This strategy is employed} when our framework recommends clinical triage or raises an OOD alert based on the analysis of the clinical image. The clinical and dermoscopic images are fed into the first half of our framework (CNN+Transformer encoder), and their features are concatenated. The concatenated features are then passed to the second half (Transformer Decoder) for analysis and diagnosis.

\subsection{Modules}

\textcolor{black}{In this section, we provide a brief motivation and overview of the two modules in our HOT skin model. Detailed information about these modules can be found in the Extension sections. The below two modules enable our HOT skin model to achieve the capabilities} of hierarchical prediction, OOD detection, and clinical triage recommendation.
\subsubsection{Hierarchical Module}
\textcolor{black}{Hierarchical models have previously proven to capture complex relationships and dependencies among categories~\cite{salakhutdinov2012learning}, and more importantly leverage prior knowledge and domain expertise from the hierarchical structure of the dataset~\cite{an2021hierarchical}. Our ``Hierarchical" module comprises of a CNN and a Transformer Encoder-Decoder framework. The CNN extracts the image features and these image features are then passed to the Transformer Encoder-Decoder for enhanced learning and specifically to achieve the three-level hierarchical prediction utilizing our unique hierarchical dataset. We chose hierarchical modeling as our foundation because it enables the explicit learning of category hierarchies, facilitating the sharing of relevant abstract knowledge across categories. This, in turn, serves as the base to develop the OOD detection capability of our HOT skin model demonstrated in the following section.}
\subsubsection{Mixup and Prototype Learning (MPL) Module}
\textcolor{black}{Given the fine-grained and long-tailed data distribution for level 3 (61 categories), in contrast to levels 1 and 2, we use level 3 for developing out-of-distribution (OOD) detection and clinical triage recommendation capabilities. To do so we build upon the hierarchical module by combining mixup strategy~\cite{zhang2017mixup} and prototype learning~\cite{yang2018robust}. Both of these established methods have shown to enhance the decision boundaries and robustness of a model~\cite{thulasidasan2019mixup,zhang2020does}. However, instead of naively adopting them for the entire dataset, we only adopt these between the less representative middle and tail categories of our dataset, thus limiting the overbearing influence of the common head categories (subset partitions depicted in Fig~\ref{fig2}(b)). These techniques are integrated with the hierarchical module, forming the ``Hierarchical + MPL'' strategy.} Further details on the training process, experimental settings, hyperparameter selection, hierarchical prediction module, MPL strategy, and thresholds for OOD detection and clinical triage recommendation can be found in the Extension ~\ref{hier-mpl-all}.

\section{Results \& Analysis}\label{sec2}

In this section, we describe the specific impact and benefit of each component in our framework by presenting extensive results and showcasing a multitude of examples.

\subsection{Hierarchical Prediction v/s Single Diagnosis Output}

\begin{table}
\centering
\resizebox{0.95\columnwidth}{!}{%
\begin{tabular}{c|c|c|c}
\hline
Input image and ground truth & \multicolumn{3}{c|}{Hierarchical Predictions} \\
\cline{2-4}
& \textcolor{level1}{Level 1} & \textcolor{level2}{Level 2} & \textcolor{level3}{Level 3} \\
\hline
{\includegraphics[width=0.15\linewidth]{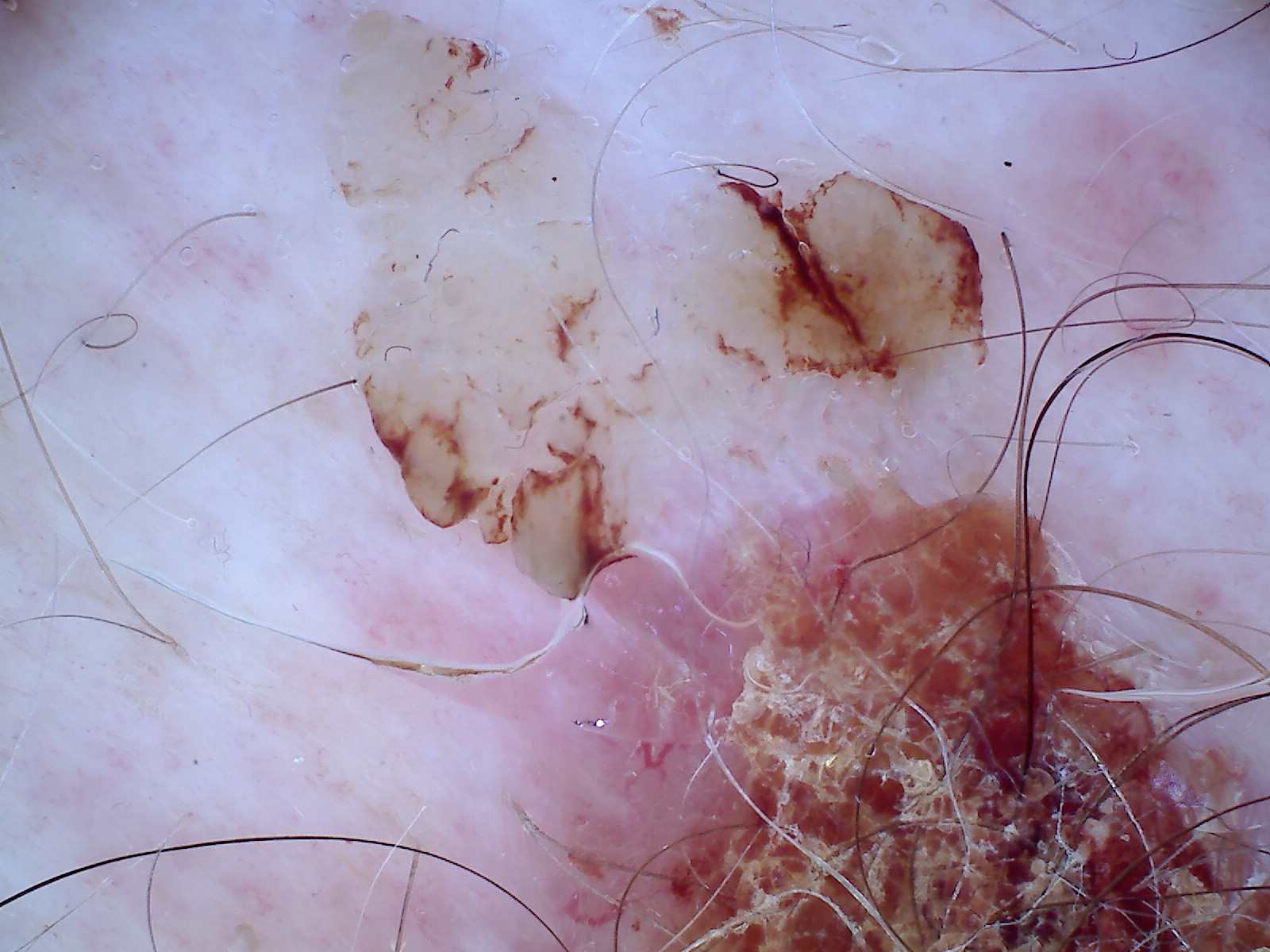}} & \textcolor{blue}{malignant} & \textcolor{blue}{malignant:bcc} & \textcolor{black}{benign:keratinocytic:seborrhoeic keratosis} \\
\multicolumn{1}{c|}{\textcolor{blue}{malignant:bcc:bcc}} & & & \\
\hline
{\includegraphics[width=0.15\linewidth]{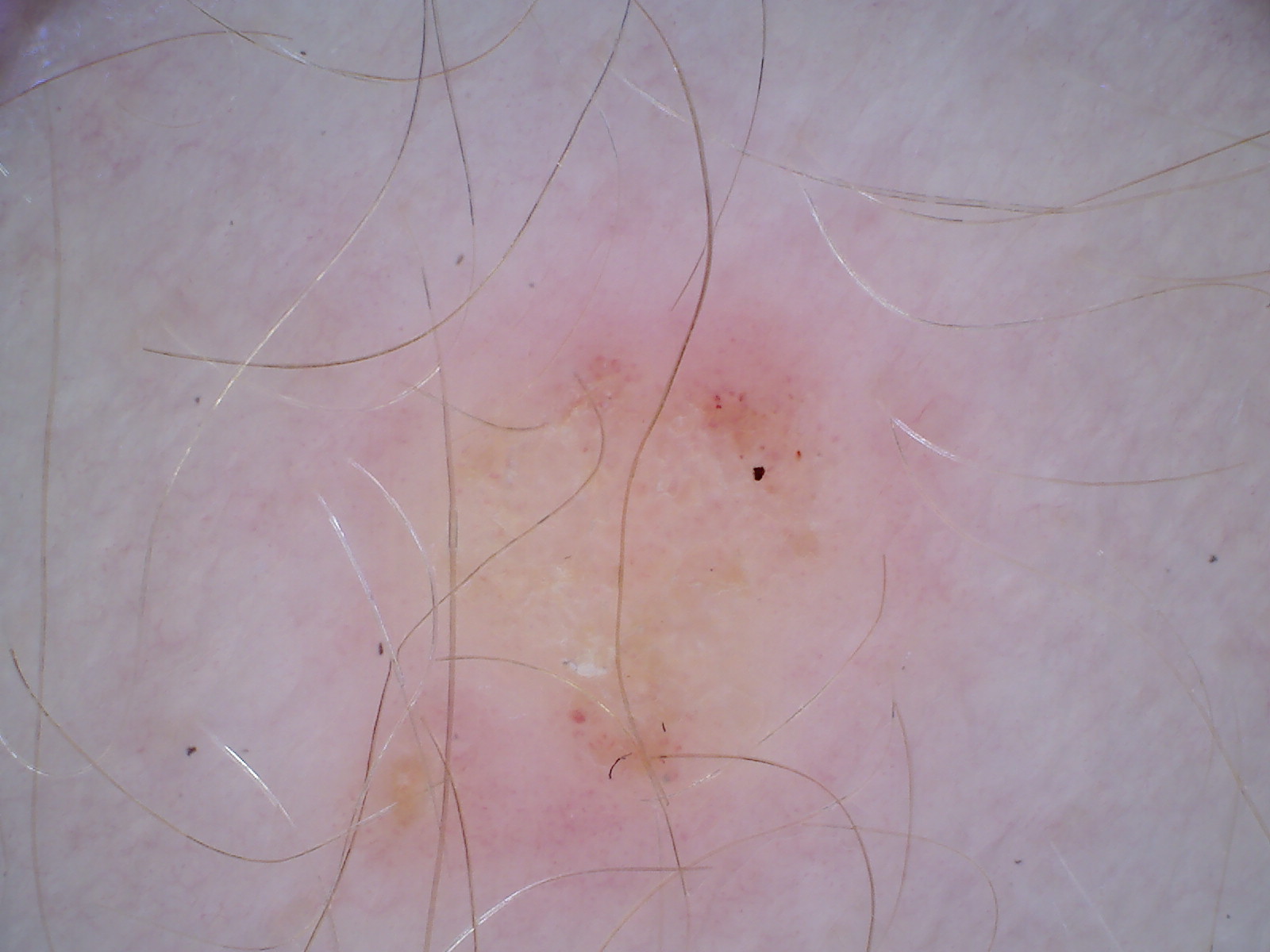}} & \textcolor{blue}{malignant} & \textcolor{blue}{malignant:scc} & \textcolor{black}{benign:keratinocytic:seborrhoeic keratosis} \\
\multicolumn{1}{c|}{\textcolor{blue}{malignant:keratinocytic: actinic keratosis}} & & & \\
\hline
{\includegraphics[width=0.15\linewidth]{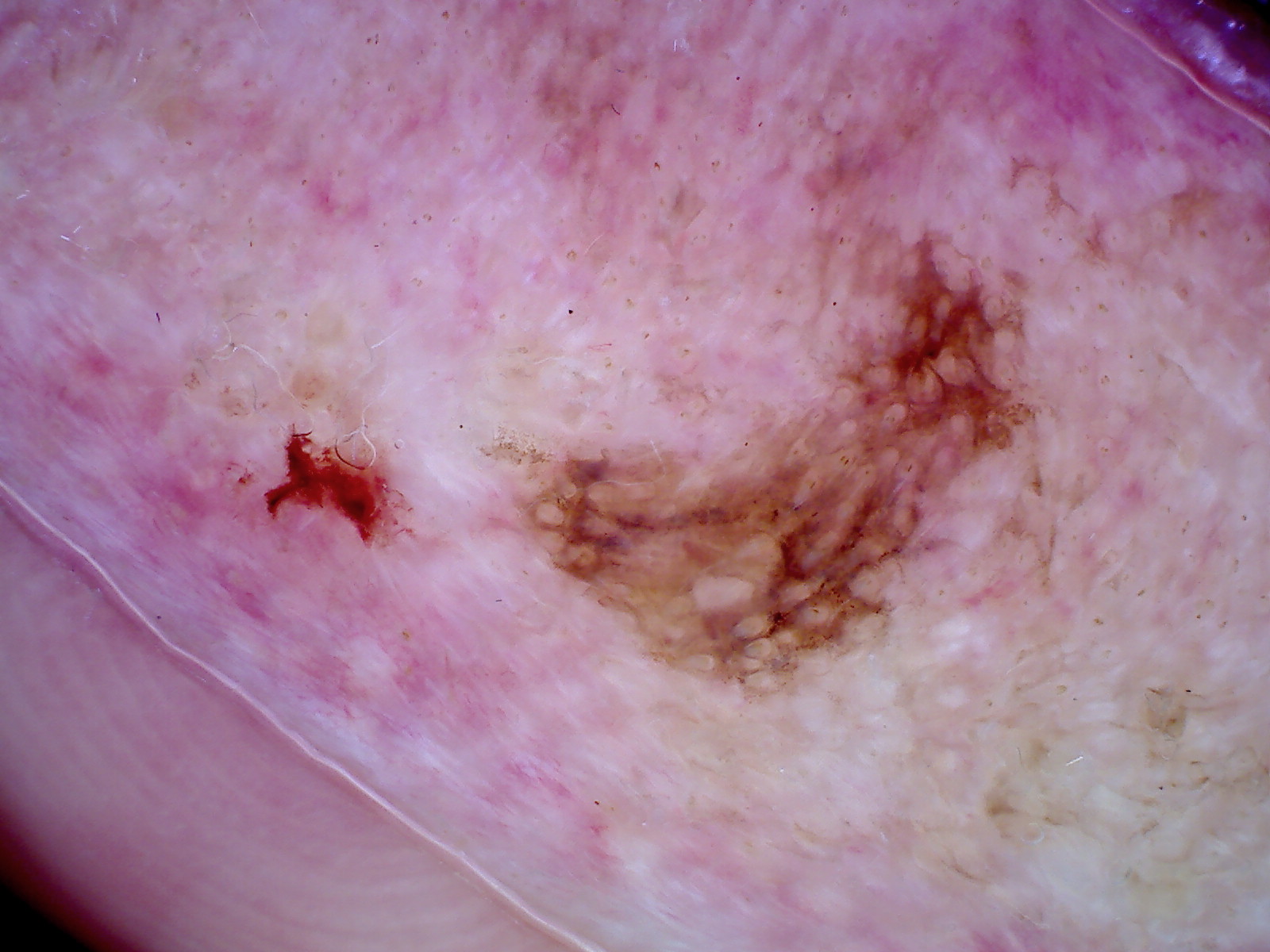}} & \textcolor{blue}{malignant} & \textcolor{blue}{malignant:melanoma} & \textcolor{black}{benign:keratinocytic:solar lentigo} \\
\multicolumn{1}{c|}{\textcolor{blue}{malignant:melanoma:melanoma}} & & & \\
\hline
{\includegraphics[width=0.15\linewidth]{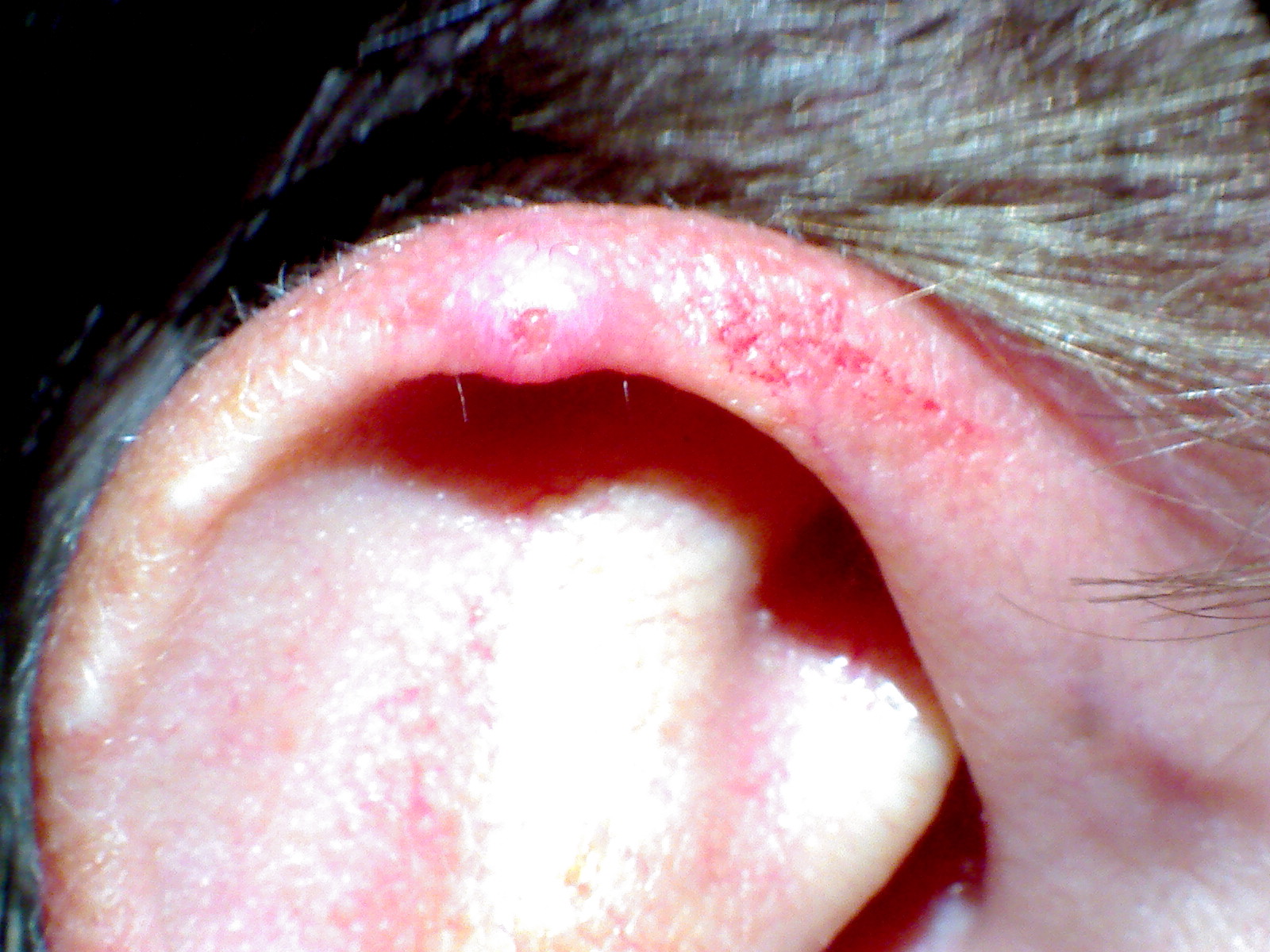}} & \textcolor{blue}{malignant} & \textcolor{blue}{malignant:scc} & \textcolor{black}{benign:other:chondrodermatitis nodularis helicis} \\
\multicolumn{1}{c|}{\textcolor{blue}{malignant:scc:scc}} & & & \\
\hline
{\includegraphics[width=0.15\linewidth]{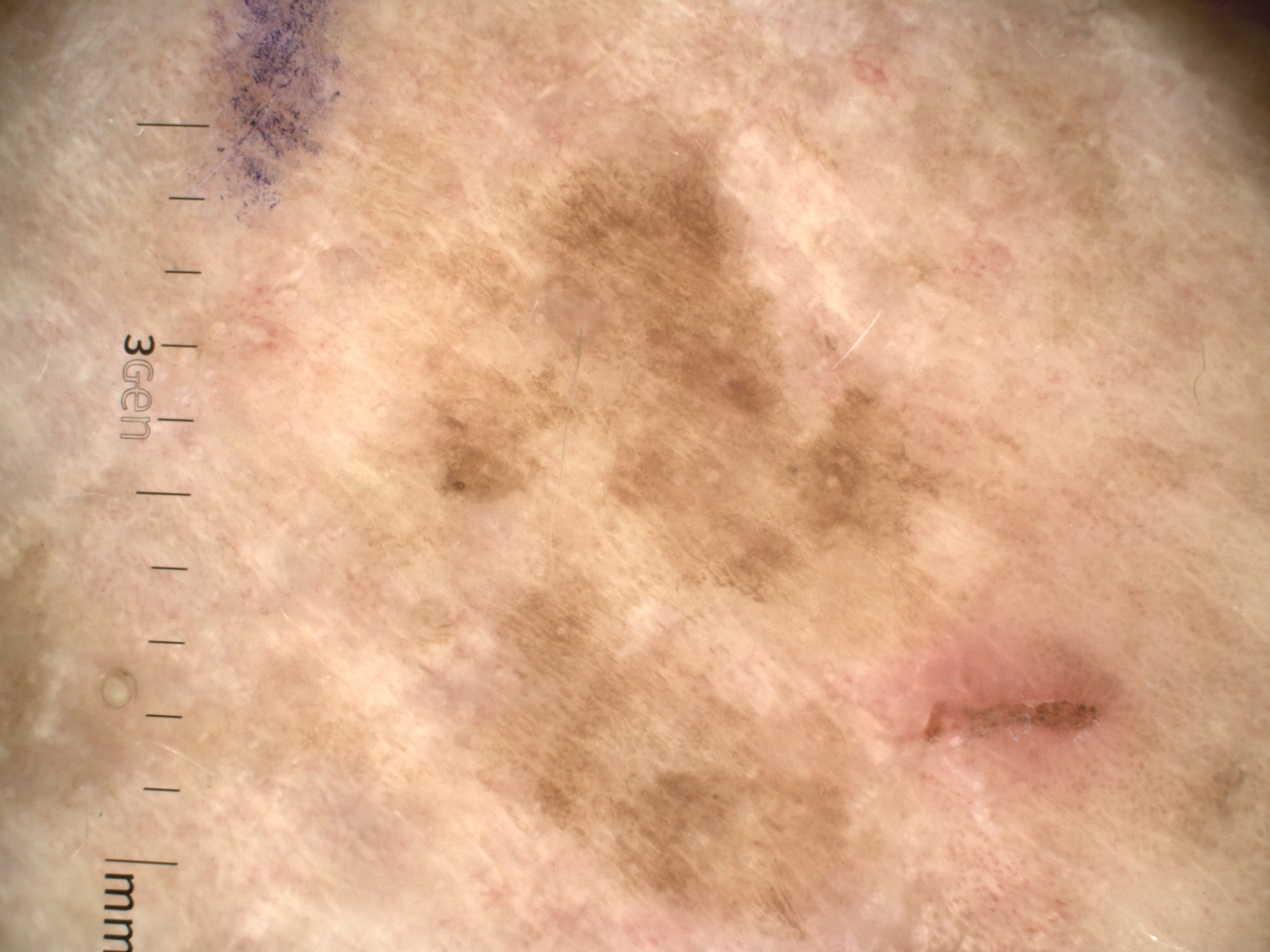}} & \textcolor{blue}{benign} & \textcolor{blue}{benign:keratinocytic} & \textcolor{black}{malignant:keratinocytic:actinic keratosis} \\
\multicolumn{1}{c|}{\textcolor{blue}{benign:keratinocytic:seborrhoeic keratosis}} & & & \\
\hline
{\includegraphics[width=0.15\linewidth]{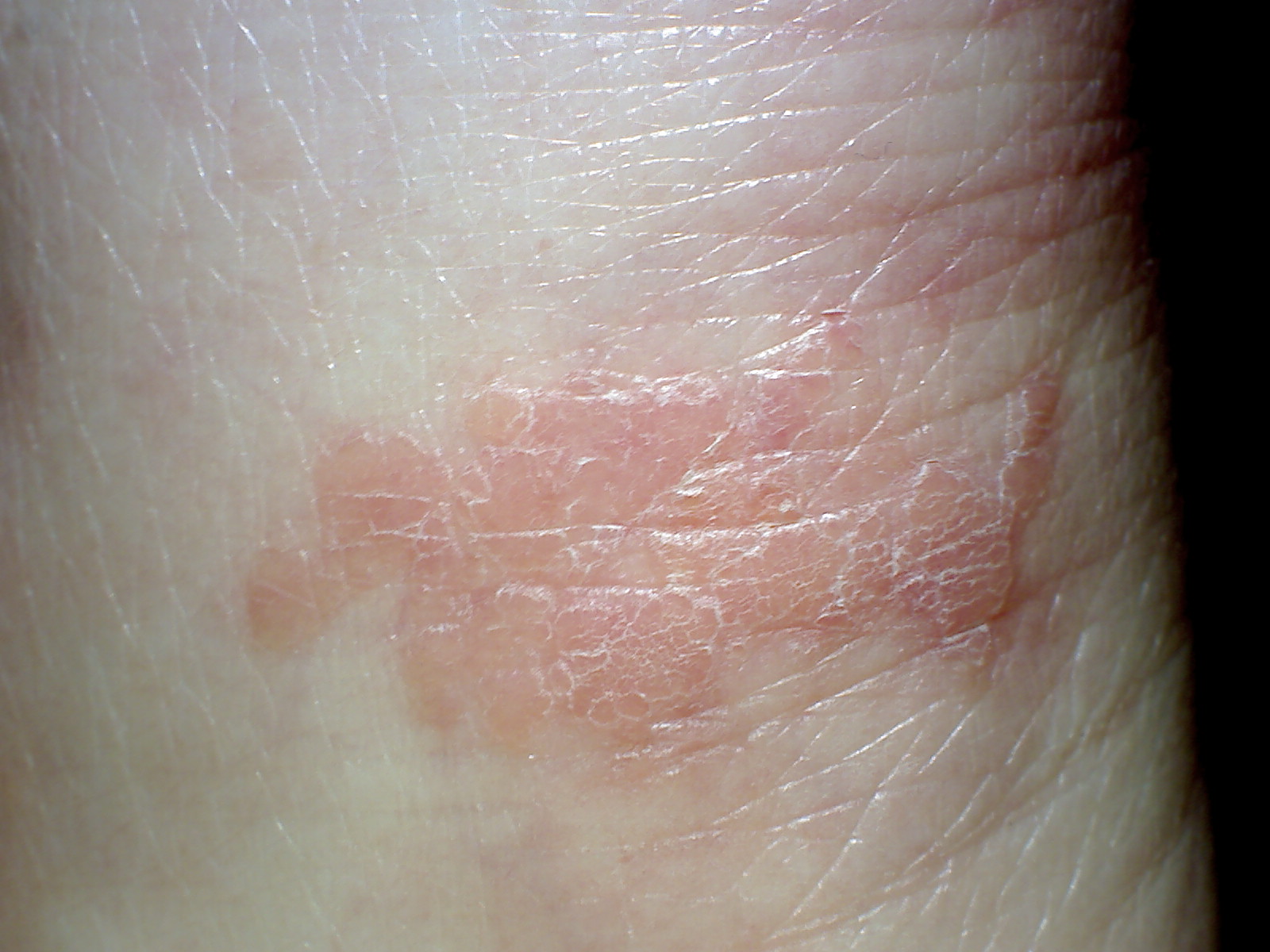}} & \textcolor{blue}{benign} & \textcolor{blue}{benign:keratinocytic} & \textcolor{black}{malignant:scc:scc in situ} \\
\multicolumn{1}{c|}{\textcolor{blue}{benign:keratinocytic:wart}} & & & \\
\hline
{\includegraphics[width=0.15\linewidth]{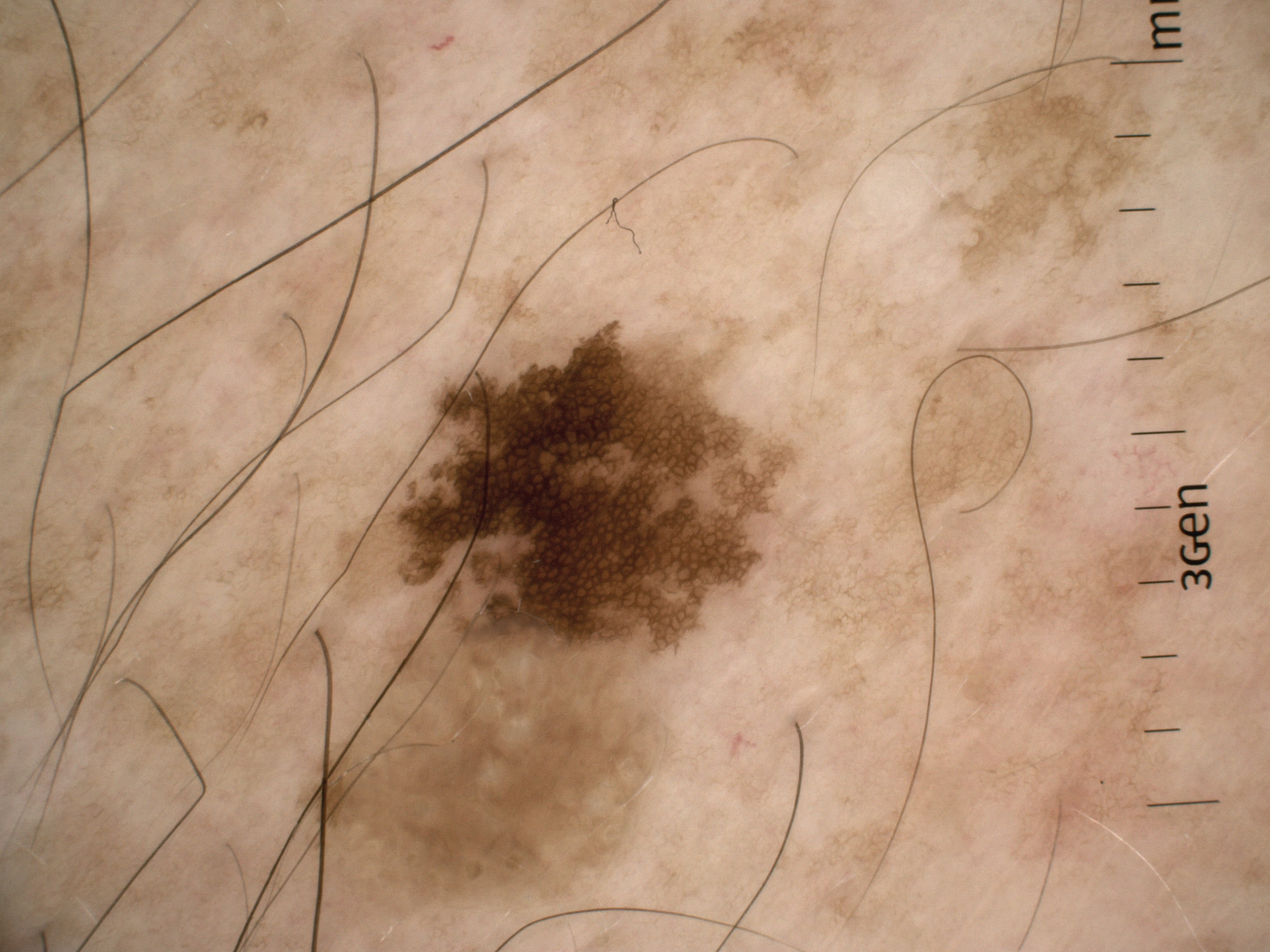}} & \textcolor{blue}{benign} & \textcolor{blue}{benign:keratinocytic} & \textcolor{black}{malignant:melanoma:melanoma} \\
\multicolumn{1}{c|}{\textcolor{blue}{benign:keratinocytic:solar lentigo}} & & & \\
\hline
{\includegraphics[width=0.15\linewidth]{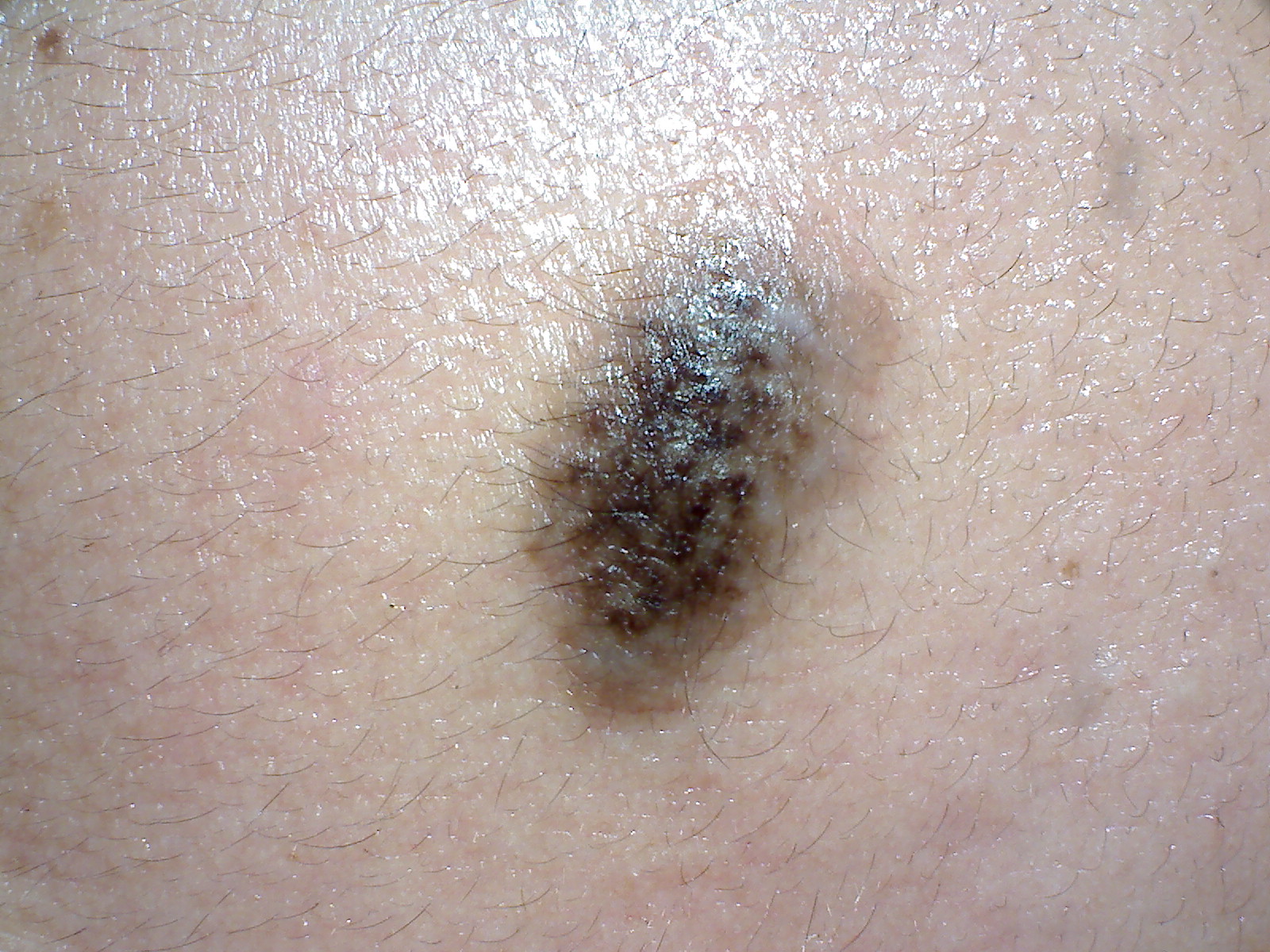}} & \textcolor{blue}{benign} & \textcolor{blue}{benign:melanocytic} & \textcolor{black}{malignant:melanoma:melanoma} \\
\multicolumn{1}{c|}{\textcolor{blue}{benign:melanocytic:congenital}} & & & \\
\hline
\hline
{\includegraphics[width=0.15\linewidth]
{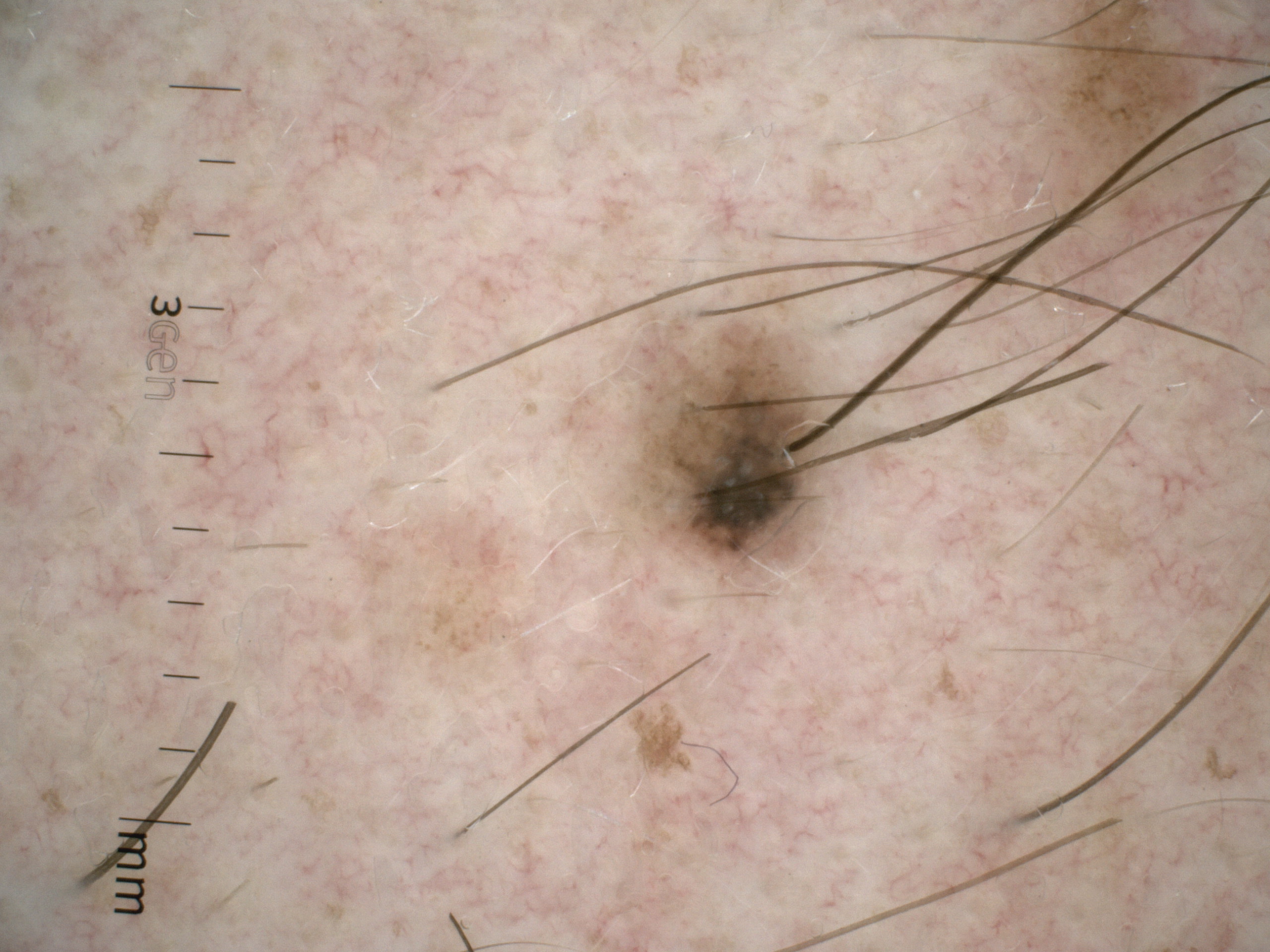}} & \textcolor{blue}{malignant} & \textcolor{blue}{malignant:melanoma} & \textcolor{blue}{malignant:melanoma:melanoma} \\
\multicolumn{1}{c|}{\textcolor{blue}{malignant:melanoma:melanoma}} & & & \\
\hline
{\includegraphics[width=0.15\linewidth]
{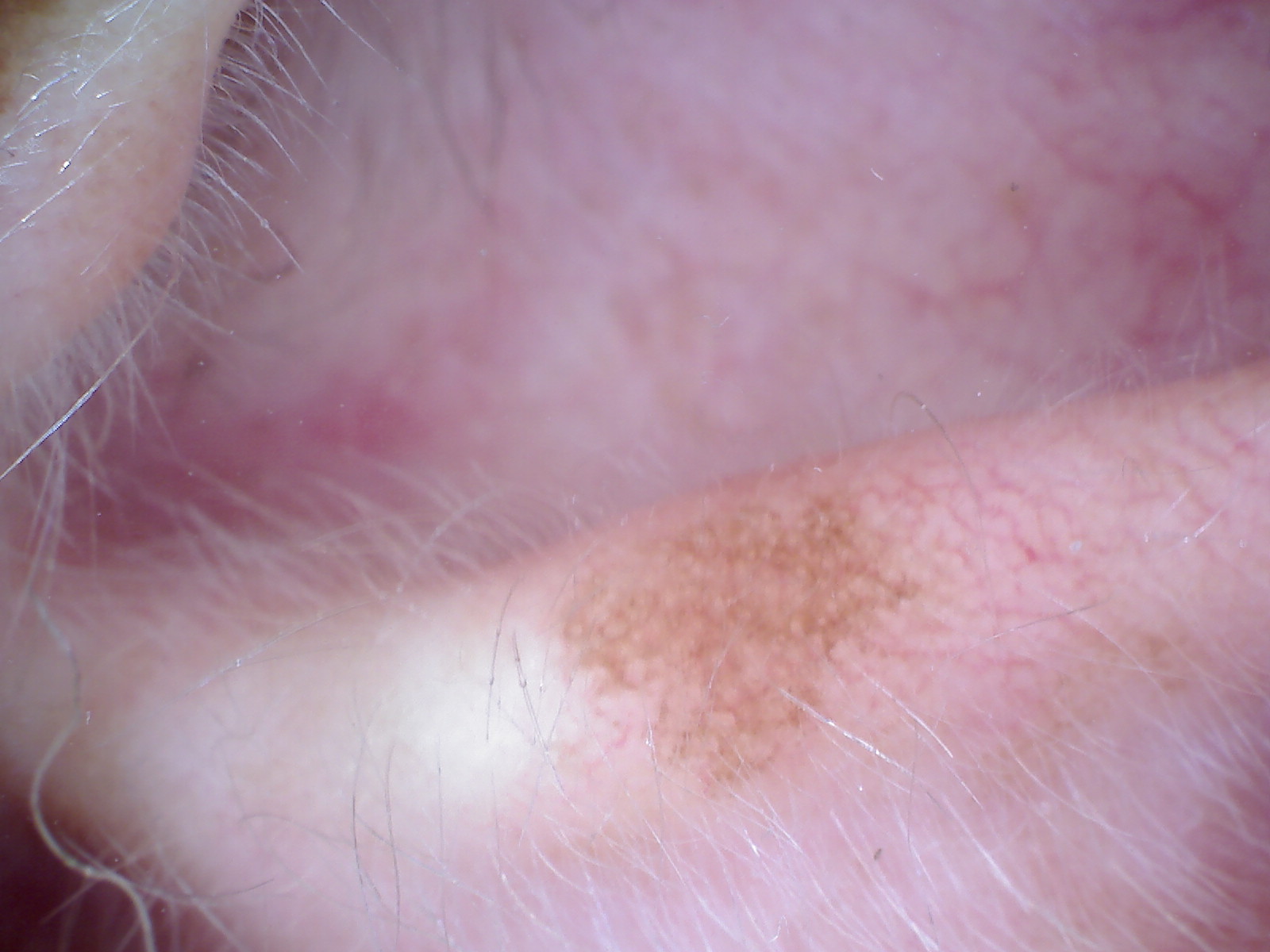}} & \textcolor{blue}{benign} & \textcolor{blue}{benign:keratinocytic} & \textcolor{blue}{benign:keratinocytic:solar lentigo} \\
\multicolumn{1}{c|}{\textcolor{blue}{benign:keratinocytic:solar lentigo}} & & & \\
\hline
{\includegraphics[width=0.15\linewidth]
{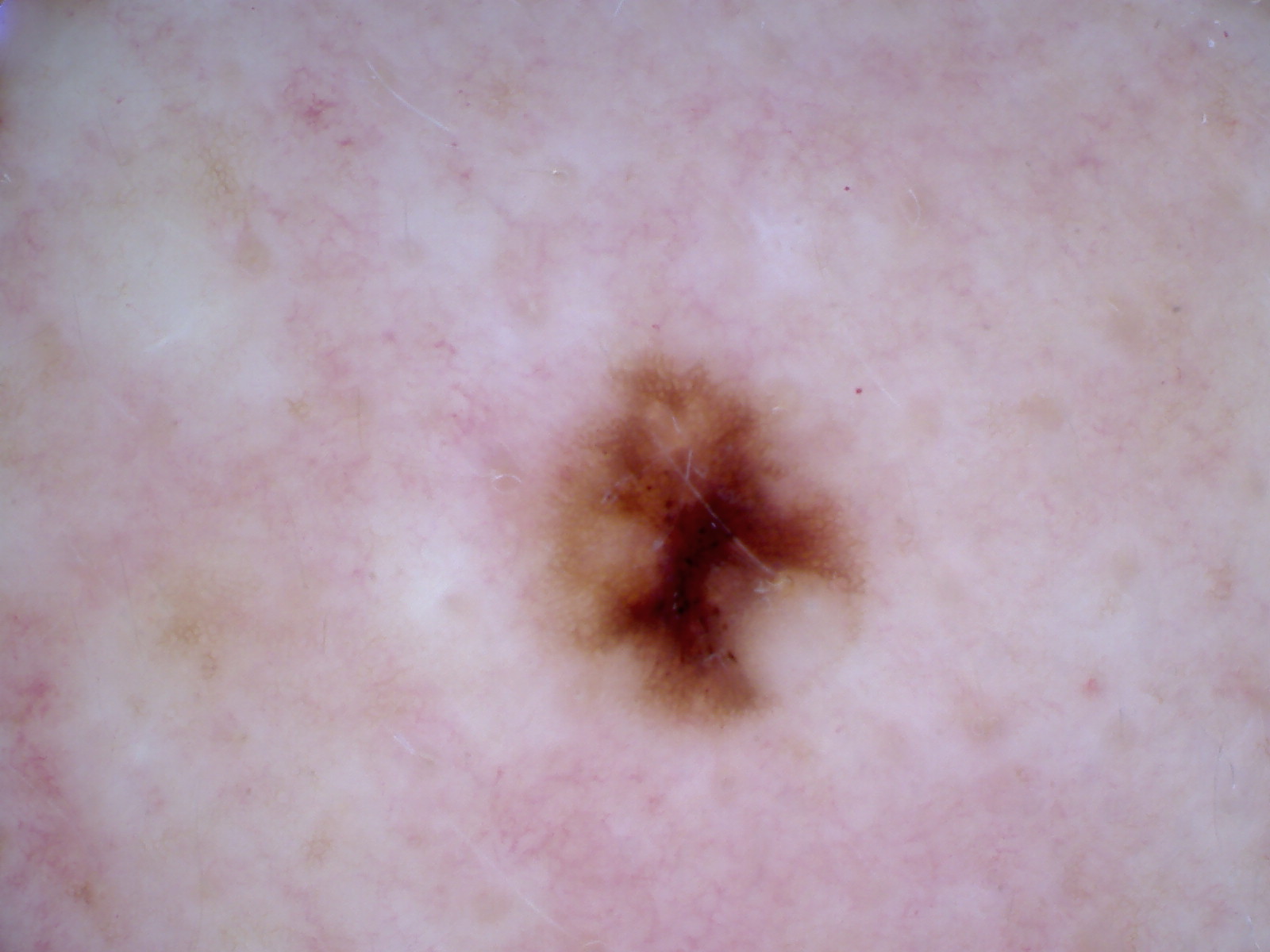}} & \textcolor{blue}{benign} & \textcolor{blue}{benign:melanocytic} & \textcolor{blue}{benign:melanocytic:atypical} \\
\multicolumn{1}{c|}{\textcolor{blue}{benign:melanocytic:atypical}} & & & \\
\hline
{\includegraphics[width=0.15\linewidth]
{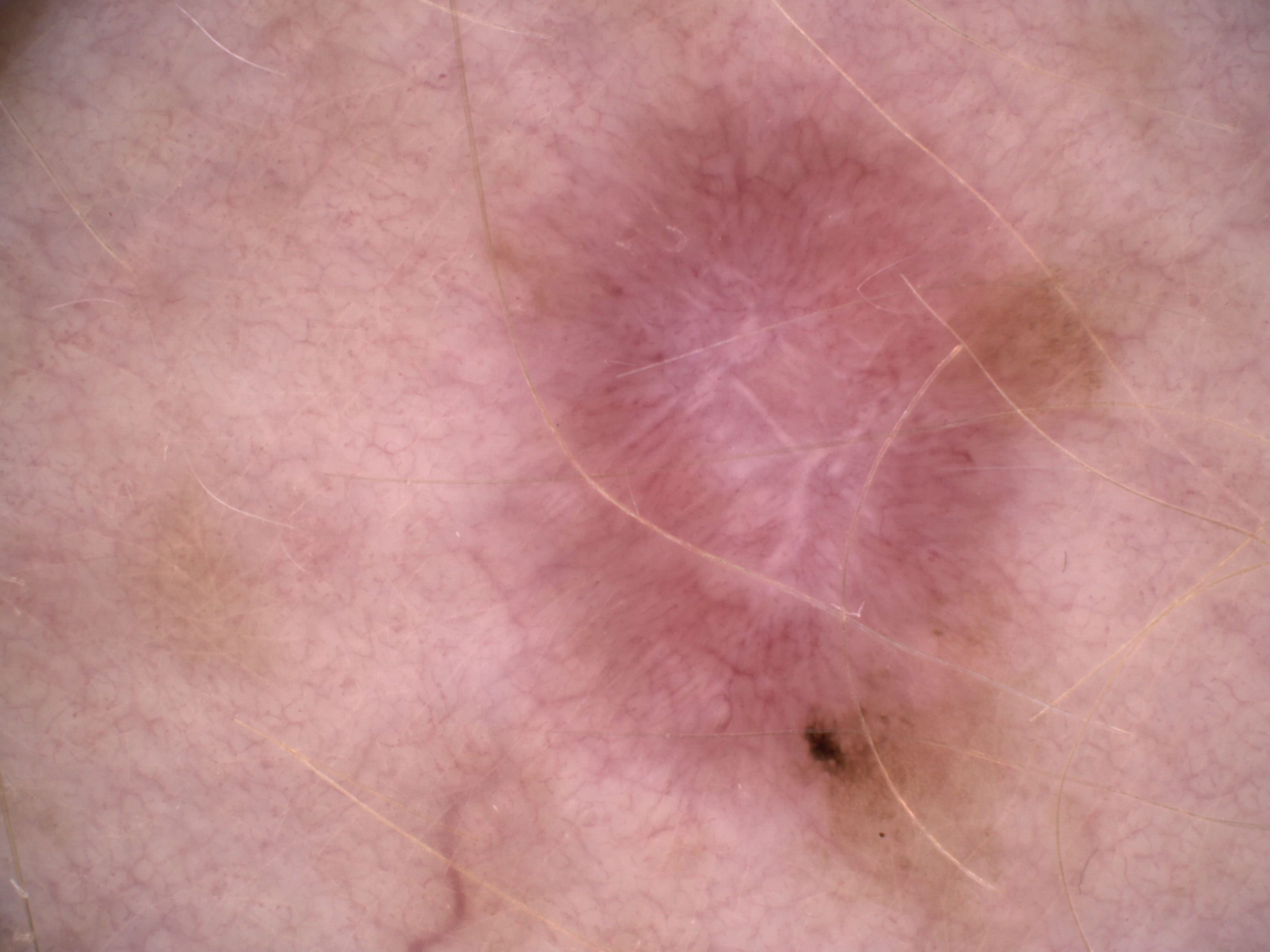}} & \textcolor{blue}{benign} & \textcolor{blue}{benign:other} & \textcolor{blue}{benign:other:dermatofibroma} \\
\multicolumn{1}{c|}{\textcolor{blue}{benign:other:dermatofibroma}} & & & \\
\hline
\end{tabular}}
\caption{Exemplar cases showing the advantage of having the functionality of hierarchical prediction - The leftmost column contains the clinical image and the corresponding ground truth category represented in the format of (level1:level2:level3). The rightmost three columns show the corresponding hierarchical predictions (level1, level2 and level3) by the hierarchical framework. We demonstrate the exemplars where a coarse and intermediate level prediction (level1 \& level2) can help to prevent a potential misdiagnosis coming from the fine-grained level prediction (level3). In all such cases, a clinical triage will be recommended.}
\label{tab:1}
\end{table}

Our hierarchical framework provides a three-level prediction for skin lesion images, which is more informative than the traditional single-output prediction~\cite{esteva2017dermatologist}. Specifically, the three levels have their exclusive classifiers which enable individual-level predictions. In Table ~\ref{tab:1}, \textcolor{black}{through some example images}, we illustrate the importance of hierarchical prediction compared to a singular low-level prediction. The table displays input images with their ground truth labels (colored \textcolor{blue}{blue}) represented as (level1:level2:level3). \textcolor{black}{The right three columns in Table~\ref{tab:1} show the hierarchical predictions for level 1, level 2, and level 3, respectively. We demonstrate two types of cases - 1) When the level 3 singular prediction is incorrect but level 1 and level 2 predictions are correct; 2) When the low-level (level 3) singular prediction is correct along with the high and mid levels - level 1 and level 2. Both the above scenarios are to demonstrate the advantage of hierarchical prediction compared to solely relying on singular prediction.}

For the first four malignant cases in Table~\ref{tab:1}, the low-level (level 3) prediction suggests benignity, while the high-level (level 1) and mid-level (level 2) predictions correctly identify the malignancy. Conversely, in the subsequent four benign cases, the low-level prediction suggests malignancy, but the higher-level predictions classify them correctly as benign. \textcolor{black}{These eight cases specifically demonstrate the potential catastrophic misdiagnosis when relying solely on the inaccurate conventional singular low-level prediction (colored \textcolor{black}{red}). In singular prediction, the first four malignant lesions will be diagnosed as benign whereas the next four benign lesions will be diagnosed as malignant. On the contrary to singular diagnosis, the correct hierarchical predictions of level 1 and level 2 could help to prevent such a misdiagnosis as they correctly represent the high and mid level category of the skin lesion.} 

\textcolor{black}{For the bottom four cases in Table~\ref{tab:1}, the predictions for all three levels are correct and thus align perfectly within the hierarchical connections of the categories for all the three levels. For these lesions, there is no problem relying solely on the singular level 3 prediction. However, even for such lesions, the hierarchical prediction does provide a high to low level overview of the predicted lesion categories, which could further instill clinician trust in the model.}

\textcolor{black}{Furthermore, we compared the classification performance on level 3 of our hierarchical prediction framework and conventional singular diagnosis across three image modalities, which is shown in Table~\ref{tab:2}. We found that hierarchical framework could infact improve the overall classification performance for level 3, while providing a broader context for understanding the categorization of the lesion with the two additional levels. Both qualitative case studies and quantitative analyses underscore the superiority of our hierarchical prediction model over traditional singular diagnostic approaches.}

\begin{table}[t!]
\caption{Comparison of classification performance for level 3 prediction of ID categories represented in terms of \{Precision, Recall and F1-score\} for Singular (Only Level 3) v/s Hierarchical Prediction across the three input image modalities of clinical, dermoscopic and clinical + dermoscopic.} \label{tab:2}
\resizebox{0.95\textwidth}{!}{
\begin{tabular}{c|c|c}
\hline \\
Input Image Modality & Singular & Hierarchical\\
\hline
 Clinical & \{0.544,0.582,0.562\} & \{\textcolor{blue}{0.550},\textcolor{blue}{0.600},\textcolor{blue}{0.574}\}\\
 Dermoscopic & \{0.553,0.597,0.574\} & \{\textcolor{blue}{0.565},\textcolor{blue}{0.604},\textcolor{blue}{0.583}\}\\
 Clinical + Dermoscopic & \{0.558,0.607,0.581\} &  \{\textcolor{blue}{0.584},\textcolor{blue}{0.632},\textcolor{blue}{0.607}\}\\
\hline
\end{tabular}}
\end{table}

\noindent\textbf{Clinical application}: \textcolor{black}{The hierarchical diagnosis is helpful to build clinician trust in the model. If the diagnosis aligns perfectly within the taxonomy of the categories for all the three levels, the clinician can be more confident in its results. Generally, when a clinician is evaluating a lesion from a dermoscopic or clinical image they will have a differential diagnosis. The clinician can think of any contradictory hierarchical diagnosis in this manner. Specifically, when the diagnosis is contradictory the clinician should not trust the model. In this case the clinician has a few options. If the image is substandard, they could take another, and re-run the model. Otherwise, they could consult a colleague for a second opinion. If no consensus can be reached and the model in any way suggested a malignancy, then the clinician should air on the side of caution and recommend excision.}

\subsection{How does the Middle-Tail subsets targeted Mixup and Prototype Learning improve \textcolor{black}{both Out-of-Distribution Detection and In-Distribution Classification Performance}}

\textcolor{black}{We build upon the base of our Hierarchical framework to integrate the Middle-Tail subsets targeted Mixup and Prototype Learning for OOD detection. We evaluated our strategy ``Hierarchical + MPL" against only ``Hierarchical" framework in terms of both OOD detection and In-Distribution classification performance. The quantitative results for Out-of-Distribution (OOD) detection on both OOD(17cl) and OOD(Unk) sets are presented in Table~\ref{tab-ood-eval}. Evidently, integrating the Middle-tail Prototype Learning (MPL) strategy with the Hierarchical approach significantly improves the performance for the OOD(17cl) set and offers a considerable enhancement for the OOD(Unk) set across all three input image modalities.}

\begin{table}[h!]
\caption{Comparison of Out-of-Distribution (OOD) detection represented in terms of AUROC(\%) for Hierarchical v/s (Hierarchical + MPL) methods for OOD(17cl) and OOD(Unk) sets across the three input image modalities of clinical, dermoscopic and clinical + dermoscopic.} \label{tab-ood-eval}
\resizebox{0.95\textwidth}{!}{
\begin{tabular}{c|c|c|c|c|}
\hline \\
\multirow{2}{*}{Input Image Modality}&\multicolumn{2}{c|}{Hierarchical} &\multicolumn{2}{|c}{Hierarchical + MPL}\\
\cline{2-5}
 & OOD(17cl) & OOD(Unk) & OOD(17cl) & OOD(Unk) \\
\hline
 Clinical & 62.66 & 61.97 & \textcolor{blue}{81.804} & \textcolor{blue}{71.89} \\
 Dermoscopic & 58.99 & 66.84 & \textcolor{blue}{83.41} & \textcolor{blue}{71.32} \\
 Clinical + Dermoscopic & 59.24 & 67.73 & \textcolor{blue}{81.25} & \textcolor{blue}{71.79} \\
\hline
\end{tabular}}
\end{table}

\textcolor{black}{In practice, an OOD technique will suffer a trade-off between achieving a high OOD detection performance and retaining the In-Distribution classification performance~\cite{vaze2021open} Thus, we also test our ``Hierarchical+MPL" strategy for In-Distribution classification performance across the three levels and three image modalities. The results are compiled in Table~\ref{tab:4}, which showcase a modest performance enhancement for levels 1 and 2, but a substantial improvement for level 3. These results further substantiate the efficacy of integrating the MPL strategy with the Hierarchical approach for enhancing the decision boundaries of In-Distribution categories.}

\begin{table}[h!]
\caption{Comparison of classification performance represented in terms of \{Precision, Recall and F1-score\} for Hierarchical v/s (Hierarchical + MPL) methods for ID categories across the three input image modalities of clinical, dermoscopic and clinical + dermoscopic.} \label{tab:4}
\resizebox{0.95\textwidth}{!}{
\begin{tabular}{c|c|c|c|c|c|c}
\hline \\
\multirow{2}{*}{Input Image Modality}&\multicolumn{3}{c|}{Hierarchical} &\multicolumn{3}{|c}{Hierarchical + MPL}\\
\cline{2-7}
 & Level 1 & Level 2 & Level 3 & Level 1 & Level 2 & Level 3\\
\hline
 Clinical & \{0.867,0.865,0.866\} & \{0.735,0.748,0.741\} & \{0.550,0.600,0.574\} & \{\textcolor{blue}{0.868},\textcolor{blue}{0.865},\textcolor{blue}{0.866}\} & \{\textcolor{blue}{0.74}, \textcolor{blue}{0.753}, \textcolor{blue}{0.746}\} & \{\textcolor{blue}{0.552}, \textcolor{blue}{0.621},\textcolor{blue}{0.584}\}  \\\
 Dermoscopic & \{0.874,0.87, 0.872\} & \{0.746,0.760,0.752\} & \{0.565,0.604,0.583\} & \{\textcolor{blue}{0.877},\textcolor{blue}{0.875},\textcolor{blue}{0.876}\} & \{\textcolor{blue}{0.758},\textcolor{blue}{0.771},\textcolor{blue}{0.764}\} & \{\textcolor{blue}{0.566},\textcolor{blue}{0.635},\textcolor{blue}{0.598}\}\\\
 Clinical + Dermoscopic & \{0.879,0.878,0.878\} & \{0.768,0.780,0.773\} & \{0.584,0.632,0.607\} & \{\textcolor{blue}{0.881},\textcolor{blue}{0.879},\textcolor{blue}{0.880}\} & \{\textcolor{blue}{0.771},\textcolor{blue}{0.782},\textcolor{blue}{0.776}\} & \{\textcolor{blue}{0.591},\textcolor{blue}{0.646},\textcolor{blue}{0.617}\}\\
\hline
\end{tabular}}
\end{table}

\begin{figure}[b!]%
\centering
\includegraphics[width=0.90\textwidth]{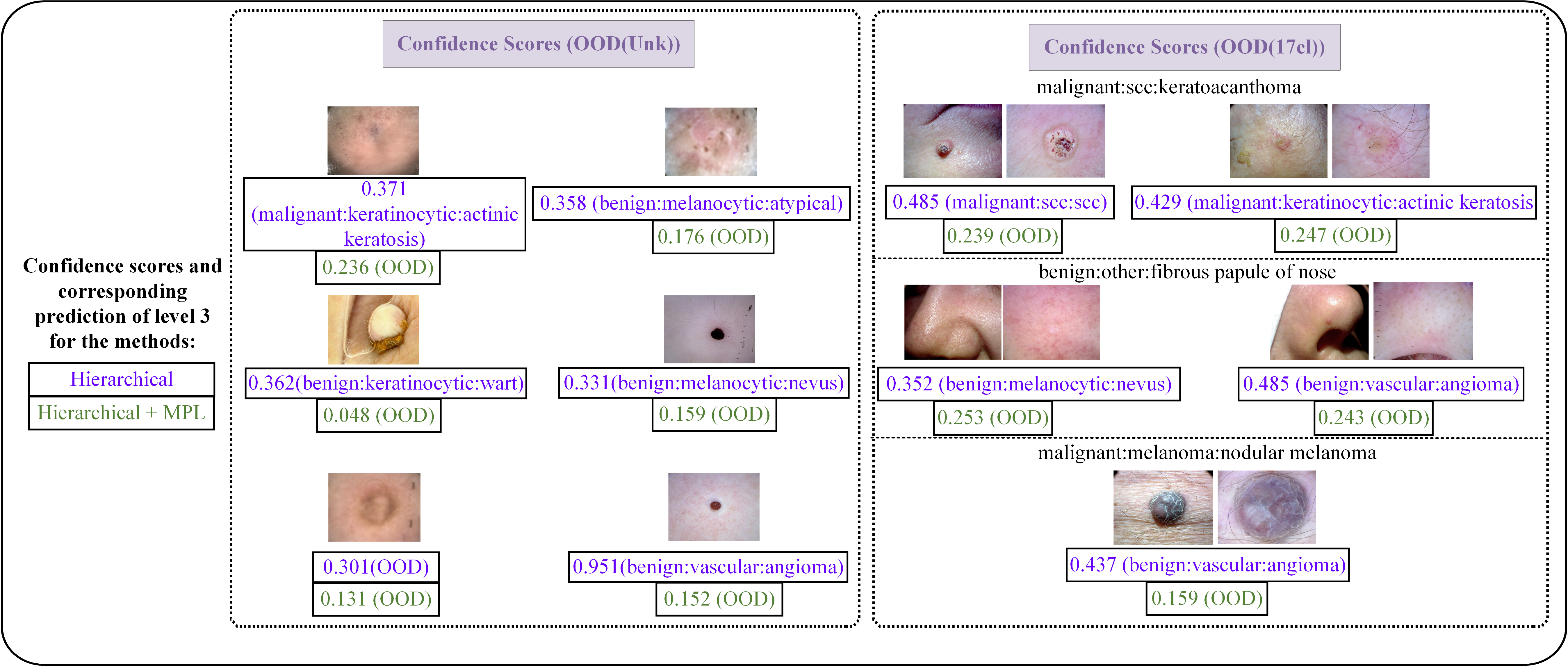}
\caption{Confidence score comparison between Hierarchical module and Hierarchical + MPL strategy for OOD (Unk) [Left] and OOD (17cl) [Right] exemplars . Lower confidence indicates better OOD detection for OOD images.}\label{fig-conf-ood-exemplars}
\end{figure}

\textcolor{black}{To demonstrate the clinical application of OOD detection capability, we show some exemplar images from OOD(17cl) and OOD(Unk) sets and compare the predicted confidence score by ``Hierarchical" and ``Hierarchical + MPL" frameworks in Fig~\ref{fig-conf-ood-exemplars}. If the confidence score of level 3 for an input image exceeds the OOD threshold, we display its confidence score and the corresponding predicted level 3 category. Otherwise, if the confidence score of level 3 is below the OOD threshold, we only present the confidence score and identify it as OOD. The ground truth OOD category for the OOD(17cl) exemplars is also presented in Fig~\ref{fig-conf-ood-exemplars}.
From the cases presented in Fig~\ref{fig-conf-ood-exemplars}, it is evident that the Hierarchical + MPL method can reliably flag OOD images, a task at which the Hierarchical-only method fails. A substantial difference in the confidence scores of the OOD images can be seen between the Hierarchical + MPL and Hierarchical-only methods. The confidence score for the Hierarchical + MPL method is significantly lower compared to the Hierarchical-only method. Thus, by predicting a lower confidence score for the OOD input images, our method is more robust and is more trustworthy in clinical decision making process.}

\noindent\textbf{Clinical application}: The results confirm the reliability of the Hierarchical + MPL strategy in clinical decision-making. This method accurately assigns lower confidence scores to OOD images and higher confidence scores to ID images, ensuring \textcolor{black}{trust} in the diagnostic process. This characteristic is crucial in clinical settings where accurate and dependable diagnoses are vital for patient management. Additionally, the robustness of our strategy improves generalization for diverse and less-represented conditions, leading to overall enhanced model performance.

\subsection{Mixup and Prototype Learning for In-Distribution Performance: \textcolor{black}{Qualitative Analysis of Hierarchical and Hierarchical + MPL Methods in Multiple Modalities}}\label{intraclass-interclass-all}

In this section, \textcolor{black}{we qualitatively assess the impact of incorporating the MPL strategy on intra-class distances (within the same category) and inter-class distances (between different categories) in the latent feature space of level 3. Smaller intra-class distances and larger inter-class distances indicate effective category separation and decision boundaries.}

\subsubsection{Intra-class variation}

\begin{figure}[t!]%
\centering
\includegraphics[width=0.865\textwidth]{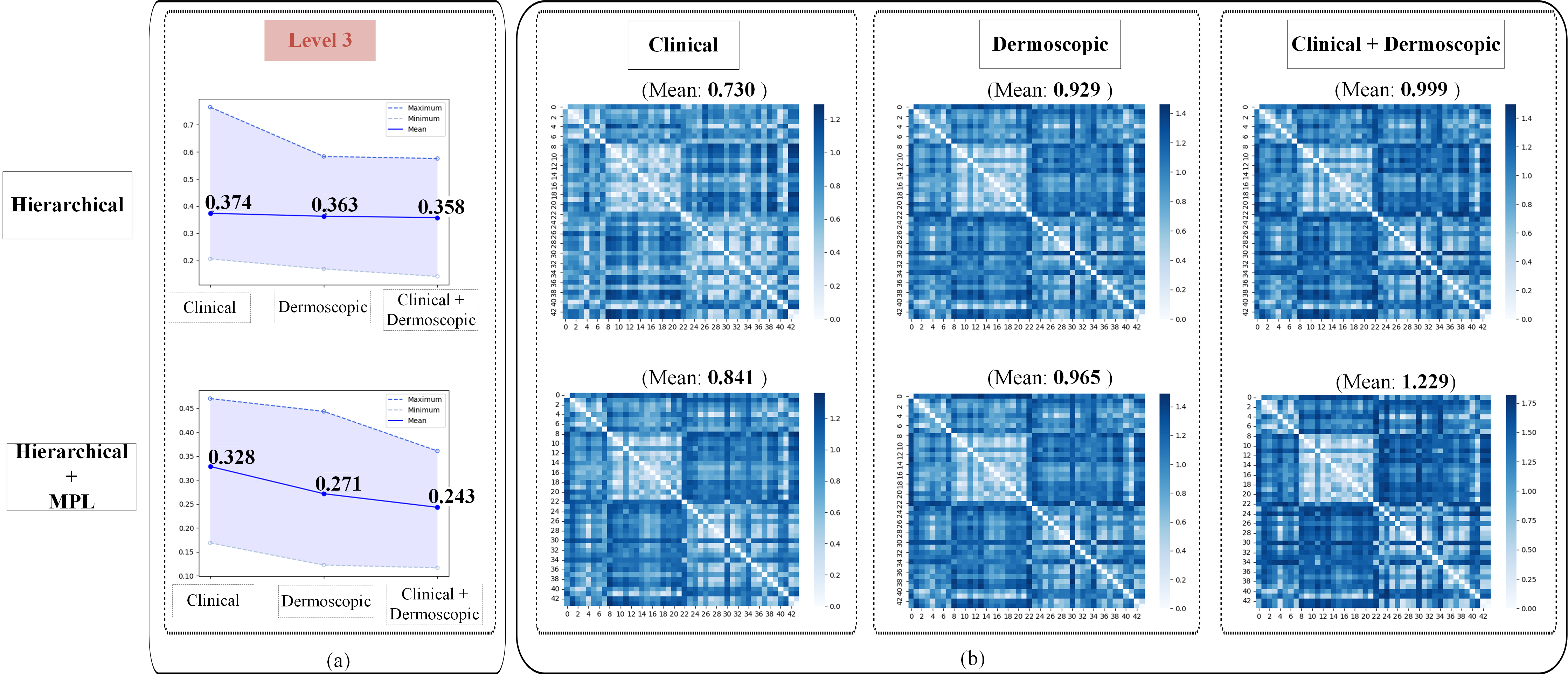}
\caption{(a) Variation of intra-class distances for In-Distribution categories. Maximum, minimum and mean of the intra-class distance of the level 3 ID categories across the three input image modalities for both the Hierarchical and Hierarchical + MPL methods. (b) Inter-class distance matrices between the 44 ID categories for both the Hierarchical and Hierarchical + MPL methods across the three input image modalities - \{clinical, dermoscopic, clinical + dermoscopic\}. The mean value of the inter-class distance is presented on the top of each matrix. A darker color indicates a higher distance value and vice-versa.}\label{fig6}
\end{figure}

In Fig~\ref{fig6}(a), we illustrate the variation (minimum, maximum, and \textbf{mean}) of intra-class distances among the \textcolor{black}{44 ID categories} for both the Hierarchical and Hierarchical + MPL methods across \textcolor{black}{the three input image modalities}. For level 1 and level 2, please refer to Extension~\ref{lev1lev2intraclass}.

\textcolor{black}{We observe that for both the Hierarchical and Hierarchical+MPL methods, the clinical + dermoscopic image has the smallest intra-class distance (\textbf{0.358, 0.243}), followed by the dermoscopic image alone (\textbf{0.363,0.271}), and the clinical image alone (\textbf{0.374,0.328}) has the largest intra-class distance.} This suggests that the dermoscopic image contains more distinct features, leading to smaller variation within the same category. The combination of clinical and dermoscopic features in the clinical + dermoscopic image modality results in the lowest intra-class distance, indicating better differentiation between categories. \textcolor{black}{In addition, we can conclude that MPL strategy reduces the intra-class distance, leading to more tightly packed distributions within categories and enhanced decision boundaries, when integrated with the Hierarchical method.}

\subsubsection{Inter-class distance}

Fig~\ref{fig6}(b) displays the inter-class distance matrices \textcolor{black}{for the Hierarchical and Hierarchical + MPL methods between the 44 ID categories} of level 3 across three input image modalities. Darker colors indicate larger inter-class distances, resulting in the white diagonal of perfect agreement for the same categories. The \textbf{mean} inter-class distances are displayed at the top of each matrix.

Fig~\ref{fig6}(b) reveals two key findings. Firstly, both the Hierarchical and Hierarchical + MPL methods demonstrate that the combined clinical + dermoscopic image has a larger inter-class distance compared to the standalone clinical or dermoscopic image. This highlights the effectiveness of the dermoscopic image in differentiating categories, while also acknowledging the valuable features present in the clinical image when combined with dermoscopy. Secondly, the Hierarchical + MPL strategy consistently exhibits higher inter-class distances across all input modalities compared to the Hierarchical module, indicating that it has inherently improved decision boundary learning between ID categories. Thus, both intra-class and inter-class distances improve with utilizing MPL strategy.

\subsection{Assessing the Efficiency of the Clinical Triage Module in Determining the Need for Dermoscopic Image Acquisition}

\begin{figure}[t!]%
\centering
\includegraphics[width=0.9\textwidth]{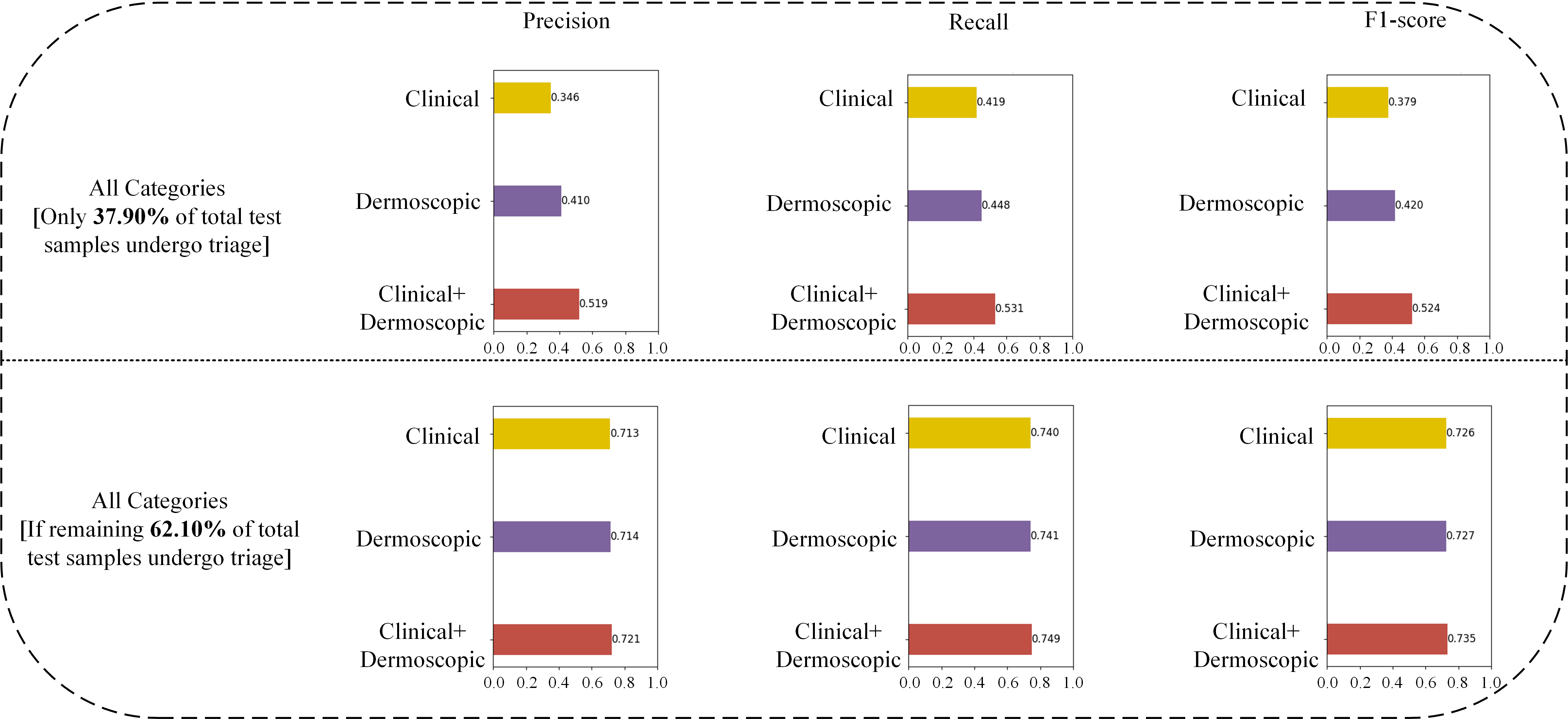}
\caption{Performance Evaluation (Precision, Recall and F1-score) of Clinical Triage strategy for the In-Distribution test images. Comparison of performance increment for all the 44 ID categories on level 3. Top: Represents the ID test images which were recommended for clinical triage and if pursued the corresponding dermoscopic image only based and combined clinical + dermoscopic image-based performance. Bottom: Represents the ID test images which were not recommended for clinical triage, nevertheless if pursued the corresponding dermoscopic image only based and combined clinical + dermoscopic image-based performance.}\label{fig-triage-all}
\end{figure}

In this section, we evaluate the clinical triage strategy for ID images at level 3. For OOD images, we always recommend acquiring a dermoscopic image for confirming the diagnosis. The effectiveness of our clinical triage module is measured by calculating the \textbf{fraction} of test images recommended for dermoscopic image acquisition and the corresponding performance \textbf{increment} at level 3. We also assess the performance increment for the remaining fraction of images. A successful clinical triage strategy will show a significant performance increase for images recommended for dermoscopy, while the remaining fraction should demonstrate marginal improvement.

Fig~\ref{fig-triage-all} presents the evaluation results of the clinical triage strategy on In-Distribution (ID) test images at level 3. In the \textcolor{black}{top} half of the figure, we observe that only \textbf{38\%} of the images are recommended for additional dermoscopy imaging. Employing both clinical and dermoscopic images results in significant improvements in precision, recall, and f1-score compared to using only the dermoscopic image. Specifically, the combination leads to an increase of {\textbf{0.173}, \textbf{0.112}, \textbf{0.145}} in precision, recall, and f1-score, respectively, while the dermoscopic image alone yields enhancements of {0.064, 0.029, 0.041}.

The \textcolor{black}{bottom} half of Fig~\ref{fig-triage-all} represents the marginal performance increment for the remaining \textbf{62\%} of the images if they were to undergo additional imaging. The precision, recall, and f1-score show slight improvements of {0.008, 0.009, 0.009} when using both images, and only {0.001, 0.001, 0.001} with the dermoscopic image alone. \textcolor{black}{It is important to note that these images already exhibit high performance based on the clinical image alone}. This evaluation highlights the effectiveness of our clinical triage module in optimally selecting images requiring additional dermoscopy for a more accurate diagnosis.

\noindent\textcolor{black}{\textbf{Clinical interpretation}: Our analysis of clinical triage module reflects the clinical scenario where a clinical image alone can often provide a high-level differentiation between potentially malignant and non-suspicious benign lesions. Similarly, a dermoscopic image alone is often adequate for a more fine-grained classification. It is the combination, however, of the clinical and dermoscopic image that provides the clinician with both the lesion context and lesion features, allowing for a more comprehensive and more accurate diagnosis for the triage recommended cases. Our clinical triage strategy thus can assist the clinicians in the decision making of whether dermoscopy image acquisition is required for a certain lesion.}

\section{Discussion}
Our All-In-One framework sets a new standard in dermatological AI modeling by integrating \textcolor{black}{multifunctionality and providing comprehensive solutions} which sharply contrasts with the prevailing single-function focus of most existing models. Traditional models tend to fixate on singular tasks such as either singular diagnosis only or out-of-distribution detection, or multi-modal analysis, limiting their clinical applicability. Our model transcends these constraints, \textcolor{black}{by uniting and building upon each one of these essential objectives}, thereby enhancing the scope and reliability of AI models in clinical decision-making processes. A notable innovation is our model's hierarchical prediction capability, \textcolor{black}{which is akin to providing a differential diagnosis for challenging lesions and is a critical necessity in dermatology}. By providing predictions at various levels, especially in scenarios where relying solely on a single prediction could result in diagnostic errors, our approach offers a multi-tiered diagnostic perspective. This enhances the clinical utility of computer-assisted diagnosis and fosters increased trust among healthcare professionals.

In terms of OOD detection, our model takes a more realistic and encompassing approach, going beyond the standard data limitations of existing AI models. Current methodologies rely heavily on select standard datasets and dermoscopic images, resulting in a narrow and potentially unrealistic representation of clinical scenarios. These methods also typically train on a limited range of reserved OOD categories, not adequately reflecting real-world clinical encounters with novel categories. In contrast, our model, backed by an extensive `long-tailed' dataset and \textcolor{black}{minority categories targeted learning}, enhances both OOD detection capability and In-Distribution (ID) classification performance. This strategy creates a more authentic clinical representation by exposing the model to a broad spectrum of known categories while testing its response to previously unseen ones. Importantly, this model is not forced to make a diagnosis, if it has not seen a category during it's training, a low classification probability will trigger the model to return a response of unknown.  

We have also augmented our framework with a unique triage recommendation functionality, addressing the lack of consensus in existing models on when to acquire additional dermoscopic images for a more precise diagnosis. This feature is particularly advantageous in resource-scarce settings, guiding decisions on whether more detailed analysis is necessary. By enabling accurate diagnosis using both dermoscopic and clinical images, our All-In-One framework bridges gaps left by existing models, paving the way for more accessible, efficient, and comprehensive dermatological diagnosis.

\textcolor{black}{While our unique dataset in terms of diversity of lesions and hierarchical taxonomy enables the development of such a versatile model, it has several limitations which may affect the generalizability of the model. Firstly, the dataset consists of images collected from Australia and New Zealand whose populations consist of those of largely European ancestry. While this is not uncommon in dermatology image datasets, it is a concerning issue which may lead to healthcare bias~\cite{daneshjou2022disparities}, and therefore must be addressed in future data collection. Secondly, patient metadata was not available alongside the images for training our framework, and therefore we cannot quantify important clinical aspects of the data such as sex, age, skin tone and ethnicity~\cite{daneshjou2022checklist}. Finally, the ground truth for these models was based on annotation by dermatologists from dermoscopic and clinical images, as opposed to the gold standard histopathology diagnosis. That being said, histopathology diagnosis is also prone to considerable variability~\cite{elmore2017pathologists}, and prospective testing of similar models using dermatologists diagnosis as ground truth have shown to be successful~\cite{felmingham2022improving}. These limitations are further detailed in Extension~\ref{limitations}. Nevertheless, we conduct an extensive evaluation of our framework on various partitions of our dataset including location-based splits, and unseen patients to demonstrate the generalizability of our approach.}

As advancements in AI models for dermatology continue to emerge rapidly, our innovative approach combining Hierarchical Prediction, Out-of-Distribution Detection, and \textcolor{black}{Clinical} Triage Recommendation lays a foundational blueprint for developing future \emph{multi-functional dermatology AI models}. These models, we believe, \emph{will not only lead to more \textcolor{black}{trustworthy} diagnoses but will also significantly streamline clinical workflows}.

\section*{Competing Interests}
\begin{itemize}
    \item H Peter Soyer is a shareholder of MoleMap NZ Limited and e-derm consult GmbH and undertakes regular teledermatological reporting for both companies HPS is a shareholder of MoleMap NZ Limited and e-derm consult GmbH and undertakes regular teledermatological reporting for both companies. HPS is a Medical Consultant for Canfield Scientific Inc, MoleMap Australia Pty Ltd, Blaze Bioscience Inc, and a Medical Advisor for First Derm.

    \item Victoria Mar has received speaker fees from Novartis, Bristol Myers Squibb, Merck and Janssen, Advisory Board fees from MSD and L'Oreal and conference sponsorship from L’Oreal.
	
	\item\textcolor{black}{Martin Haskett is a shareholder of Molemap NZ Limited}
	
	\item\textcolor{black}{Adrian Bowling is a shareholder of Molemap NZ Limited}

\end{itemize}

\begin{appendices}

\section{Data Collection Details}\label{secDataCol}
For the creation of our dataset, we employed a standardized approach for data collection and image acquisition. The skin lesion images included in our dataset were obtained using a specialized skin lesion imaging camera (DermLite) designed by Molemap and manufactured by 3Gen. The Fujinon 12mm camera lens was set at f4, producing a 16mm field of view for dermoscopic images and 35mm for macro images. Calibration of color balance was performed using a specific tile and maintained constant thereafter.

All images were captured by trained healthcare professionals, specifically nurses who underwent three months of specialized instruction for this role. This ensured uniformity in conditions, image quality, and resolution, with each image standing at 5MP using JPEG compression. In terms of artifacts, only hair was a minor factor, appearing in a fraction of the images.

The uniformity in lighting conditions for all the images was ensured through the use of LED lights integrated into the camera lens. For dermoscopic images, the camera lens was placed directly on the skin, while for clinical images, it was positioned 10 centimeters away from the skin surface, a distance maintained precise through a camera structure.

Our dataset comprises images collected from \textcolor{black}{141 dedicated private community clinics in Australia and NewZealand. It has been collected over years from 2005 until 2020, with majority of data collection taking place in between 2010 and 2019. The 208,287 images in the dataset belong to a total of 78,760 patients}. To maintain color uniformity, color calibration was integrated into the camera software, preventing any operator adjustments. As a result, all images maintained consistent color calibration. The dataset includes both polarized and non-polarized dermoscopic images, though we specifically leveraged the polarized ones for model building. \textcolor{black}{The ground truth annotations for the skin lesions were provided by dermatologists using the Molemap system, based on dermoscopoic and clinical images. While meta-data was not encoded in the curated dataset, dermatologists were also provided with sex, age, anatomical location and melanoma history when providing the ground truth lesion diagnosis.}

\begin{sidewaystable}
\centering
\caption{Split of In-Distribution (ID) and Out-of-Distribution categories from the 61 categories of level 3 in our dataset. Categories are grouped according to their respective annotations under the hierarchicary tree. For example, the category of benign:melanocytic:atypical has the annotation labels of benign (level 1), melanocytic (level 2) and atypical (level 3).}\label{taba1}
\resizebox{0.95\columnwidth}{!}{%
\begin{tabular}{c|c|c|c}
\hline
Split& Level 1 & Level 2 & Level 3\\
\hline
\multirow{8}{*}{In-Distribution} & \multirow{4}{*}{benign} & melanocytic  & atypical,  benign naevi,  compound,  junctional, dermal, acral, blue, papillomatous, irritate, congenital, involuting/regressing, halo, nail unit nevus, lentignous \\
\cline{3-4}
& & keratinocytic & seborrhoei keratosis, solar lentigo, lentigo, ink spot lentigo, wart, lichenoid keratosis, actinic chellitis, porokeratosis \\
\cline{3-4}
& & vascular & haematoma, angioma/hemangioma, telangiectasia \\
\cline{3-4}
& & other & dermatofibroma, excpriation, sebaceous hyperlasia, scar, chondrodermatitis nodularis helicis, comedone, eczema, nail dystrophy, dermatitis, folliculitis, other \\
\cline{2-4}
& \multirow{4}{*}{malignant} & keratinocytic & actinic keratosis \\
\cline{3-4}
& & melanoma & melanoma, lentigo maligna  \\
\cline{3-4}
& & basal cell carcinoma  &  superficial, basal cell carcinoma, pigmented bcc  \\
\cline{3-4}
& &  squamous cell carcinoma  &  squamous cell carcinoma, scc in situ (IEC, Bowen's disease) \\

\hline

\multirow{5}{*}{Out-of-Distribution}  & \multirow{3}{*}{benign} & melanocytic & melanosis, spitz naevus, ephilides, reed naevus, agminate, en cockarde \\
\cline{3-4}
& & vascular & angiokeratoma \\
\cline{3-4}
& &  other & psoriasis, skin tag, fibrous papule of nose, myxoid cyst, epidermal cyst, accessory nipple, granuloma annulare, molluscum contagiosum  \\
\cline{2-4}
& \multirow{2}{*}{malignant} & melanoma & nodular melanoma \\
\cline{3-4}
& & squamous cell carcinoma & keratoacanthoma \\

\hline

\end{tabular}}
\end{sidewaystable}

\subsection{Split of ID and OOD classes}\label{appsplit}
We split the 61 categories of level 3 in our dataset into 44 In-Distribution (ID) and 17 Out-of-Distribution (OOD) categories. \textcolor{black}{We recommend reserving the bottom most 25\% percentile (based on the number of samples) of the categories as the OOD for a relatively uniform dataset distribution, as these would represent the most underrepresented categories and less likely to be encountered in clinical settings. If a dataset has highly imbalanced distribution, we would recommend choosing a threshold based on the number of samples to split the ID and OOD categories in addition to considering the bottom most 25\% categories.} In Table~\ref{taba1}, we list the categories which fall into ID and OOD sets for our dataset.

\subsection{\textcolor{black}{Train-Test splits for training and validation}}\label{secA1}

During training, OOD categories are \textbf{not shown} to the network. \textcolor{black}{The data samples belonging to the ID categories were then partitioned into an 85\%-15\% train-test split based on patient IDs}, with the training set further split into 80\%-20\% for training and validation, respectively. \textcolor{black}{With the above train-test split, training data consisted of 68,364 patients distributed amongst the 44 ID categories whereas testing data for the ID categories comprised of 23,208 patients with 13,138 patients being common in training and testing sets while OOD (17cl) and OOD (Unk) sets comprised of 371 unique patients. The 13,138 common patients between train and test sets of ID categories was inevitable as each patient had multiple images in the dataset and in order to make a challenging long-tailed distribution with 44 ID categories, dropping out specific patients was not feasible. Nevertheless, to evaluate the generalizability of our framework across patients, we also evaluated our framework exclusively on the images of the unique 10,025 patients contained only in the ID test set.} Our framework and associated methods were trained exclusively on the training dataset, utilizing the validation set for pinpointing the optimal performing checkpoints during the training process. Final evaluations were carried out on the test set using the saved checkpoint, the results of which were reported subsequently.

\section{Details of Hierarchical Module and MPL Strategy}\label{hier-mpl-all}
\subsection{Hierarchical Prediction Module}\label{hier-pred-sec}
Our hierarchical prediction output is formed by combining a Convolutional Neural Network (CNN) and Transformer Encoder-Decoder modules. Resnet34~\cite{he2016deep} was employed as our CNN architecture, while the Transformer Encoder-Decoder block of End-to-End Detection Transformer (DETR)~\cite{carion2020end} was adopted for encoding and decoding the features learned by the CNN. The Adam optimizer was used for training all strategies and methods with a batch size of 32, an initial learning rate of 1e-4 with exponential decay over 45 epochs. The input skin lesion images were resized to 320x320, and standard data augmentation practices of random cropping and horizontal flipping were adopted.

Given an input image of $x \in \mathbb{R}^{3\times320\times320}$, Resnet34 generates a feature activation map $f \in \mathbb{R}^{512\times7\times7}$. A 1x1 convolution is then applied to reduce the channel dimension of the feature activation map from 512 to 256, creating a new feature map $z_{0} \in \mathbb{R}^{256\times7\times7}$. As the Transformer encoder module expects the input as a sequence, the feature activation map $z_{0}$ is collapsed into one dimension as $256\times49$. Each encoder layer in the Transformer includes a standard architecture comprising a multi-head self-attention module and a feed-forward network (FFN), further supplemented with fixed positional encoding~\cite{parmar2018image,bello2019attention} that is added to the input of each attention layer. The memory embeddings learnt by the Transformer Encoder are then passed to the Transformer Decoder module which also follows the standard architecture of a transformer, transforming the input memory embeddings using multi-headed self attention mechanisms. To produce distinct results, the input embeddings must differ since the decoder is permutation-invariant. These input embeddings are the learnt positional encodings, referred to as \textbf{hierarchical queries}, which are then added to the input of each attention layer in a similar manner as we did in the Transformer Encoder. These hierarchical queries are converted into output embeddings by the Transformer Decoder and are then independently fed to a fully connected network represented as Level 1 Classifier, Level 2 Classifier, and Level 3 Classifier (Fig~\ref{fig1}), which transforms them into the corresponding output number of categories for each hierarchical level (\textbf{trained through standard cross-entropy loss}). \textcolor{black}{It should be noted that the classifiers are unique to each level, allowing for more independence in confidence about the prediction at each level.}

\textcolor{black}{In the multi-modal images scenario (feedback loop due to an OOD alert or clinical triage recommendation)}, the memory embeddings of the dermoscopic image from the Transformer Encoder are concatenated with those of the clinical image before passing the combined embeddings to the Transformer Decoder. In this case, \textcolor{black}{the flow of the framework remains consistent as above}, except that the dimension of the memory embeddings is doubled.

\subsection{\textcolor{black}{OOD Detection and Clinical Triage capabilities}}\label{MPL}

\textcolor{black}{Having built the hierarchical module as our base, we now explain the development of OOD detection and clinical triage capabilities of our framework. Our approach is based on the subset partition strategy for the In-Distribution categories.}

\subsubsection{\textcolor{black}{Subset Split of ID categories}}\label{secA2}

\textcolor{black}{To evaluate OOD detection capability for our framework, we intend to emulate the real world setting where the categories with the least number of samples in the dataset are more likely to be encountered as OOD categories. Thus, we reserve 17 categories (level 3) in our dataset, with each containing less than 100 samples as the Out-of-Distribution categories and denote it as the OOD(17cl) set. Additionally, to develop the OOD detection capability of our framework, we intend to use mixup between different subsets of the In-Distribution categories. Thus, we further split the level 3 ID categories into the following subsets based on their sampling distribution - (i) Head \{H\} - having 6 categories with each containing more than 10,000 samples; (ii) Middle \{M\} - comprising of 21 categories with each having 500 to 10,000 samples; (iii) Tail \{T\} - which has 17 categories with each containing 100 to 500 samples (all partitions shown in Fig~\ref{fig2}(b)). The data distribution amongst the different categories on level 3 is shown in Table~\ref{tab:long-tailed-ID} and Table~\ref{tab:long-tailed-OOD}.} Training is performed only on samples from the $\{H\}$, $\{M\}$, and $\{T\}$ categories; the reserved OOD categories are never shown during training.

\begin{table}[h]
\centering
\caption{Data distribution amongst the 44 ID categories of level 3}
\begin{tabular}{|l|l|}
\hline
Category & Number of samples \\
\hline
malignant:keratinocytic:actinic keratosis & 46126 \\
benign:melanocytic:atypical & 29039 \\
benign:melanocytic:benign nevus & 25018 \\
benign:keratinocytic:seborrhoeic keratosis & 24278 \\
malignant:bcc:basal cell carcinoma & 21243 \\
malignant:melanoma:melanoma & 12180 \\
benign:melanocytic:compound & 8829 \\
benign:keratinocytic:solar lentigo & 5693 \\
benign:other:dermatofibroma & 5226 \\
malignant:scc:squamous cell carcinoma & 3381 \\
malignant:scc:scc in situ & 3104 \\
benign:keratinocytic:lentigo & 2666 \\
benign:melanocytic:junctional & 2515 \\
benign:melanocytic:dermal & 2257 \\
benign:melanocytic:acral & 1888 \\
benign:melanocytic:blue & 1283 \\
benign:vascular:angioma & 1164 \\
benign:melanocytic:papillomatous & 1020 \\
benign:other:sebaceous hyperplasia & 981 \\
malignant:bcc:superficial basal cell carcinoma & 903 \\
benign:melanocytic:irritated & 780 \\
malignant:melanoma:lentigo maligna & 726 \\
benign:vascular:telangiectasia & 675 \\
benign:melanocytic:congenital & 640 \\
benign:keratinocytic:lichenoid & 570 \\
benign:other:scar & 541 \\
benign:melanocytic:involutingregressing & 535 \\
benign:keratinocytic:wart & 486 \\
benign:keratinocytic:actinic cheilitis & 447 \\
benign:keratinocytic:porokeratosis & 359 \\
benign:keratinocytic:ink spot lentigo & 344 \\
benign:vascular:haematoma & 334 \\
benign:other:excoriation & 329 \\
benign:vascular:other & 316 \\
malignant:bcc:pigmented basal cell carcinoma & 265 \\
benign:melanocytic:halo & 208 \\
benign:other:chrondrodermatitis & 202 \\
benign:other:comedone & 187 \\
benign:other:eczema & 161 \\
benign:melanocytic:ungual & 148 \\
benign:melanocytic:lentiginous & 141 \\
benign:other:nail dystrophy & 128 \\
benign:other:foliculitis & 102 \\
benign:other:dermatitis & 102 \\
\hline
\end{tabular}\label{tab:long-tailed-ID}
\end{table}

\begin{table}[h]
\centering
\caption{Data distribution of the reserved OOD (17cl) 17 categories of level 3}
\begin{tabular}{|l|l|}
\hline
Category & Number of samples \\
\hline
benign:melanocytic:melanosis & 98 \\
malignant:scc:keratoacanthoma & 95 \\
benign:other:psoriasis & 87 \\
benign:other:skin tag & 76 \\
benign:other:fibrous papule of nose & 67 \\
benign:melanocytic:spit nevus & 66 \\
benign:other:myxoid cyst & 60 \\
benign:vascular:angiokeratoma & 43 \\
benign:other:epidermal cyst & 43 \\
benign:melanocytic:ephilides & 32 \\
benign:other:accessory nipple & 26 \\
benign:other:granuloma annulare & 21 \\
benign:other:molluscum contagiosum & 14 \\
benign:melanocytic:reed nevus & 13 \\
benign:melanocytic:agminate & 12 \\
benign:melanocytic:en cockarde & 8 \\
malignant:melanoma:nodular melanoma & 5 \\
\hline
\end{tabular}\label{tab:long-tailed-OOD}
\end{table}

\subsubsection{\textcolor{black}{Mixup and Prototype Learning (MPL)}}

\textcolor{black}{It is necessary to learn better decision boundaries to achieve OOD detection and clinical triage recommendation capabilities~\cite{thulasidasan2019mixup,zhang2020does}. Thus, we develop an integration of two established techniques - mixup and prototype learning}. While we maintain the same cross-entropy loss for level 1 and level 2 (as in the stand-alone hierarchical prediction model), we adopt two strategies for level 3 - (i) Mixup of images between the categories of middle $\{M\}$ and tail $\{T\}$ subsets and (ii) Prototype loss for all the subsets $\{H\}$, $\{M\}$, and $\{T\}$. For a mixup image, we develop a custom prototype loss function. Below we explain the Middle-Tail targeted Mixup and Prototype Learning (MPL) strategy. This combination allows for improved decision boundary definition for long-tailed and fine-grained categories of level 3, thereby facilitating the development of OOD detection and clinical triage capabilities on level 3.

\subsubsection{Mixup strategies}
Mixup technique~\cite{zhang2017mixup} combines two different images and their corresponding labels, which can help to create better decision boundaries between the two corresponding categories. However, merely employing the mixup strategy for a long-tailed dataset could lead to an overbearing influence from the head classes (head subset {H} illustrated in Fig~\ref{fig2} (b)). This influence confines its benefits to the most common skin lesion categories in the dataset.

\begin{figure}[t!]%
\centering
\includegraphics[width=0.95\textwidth]{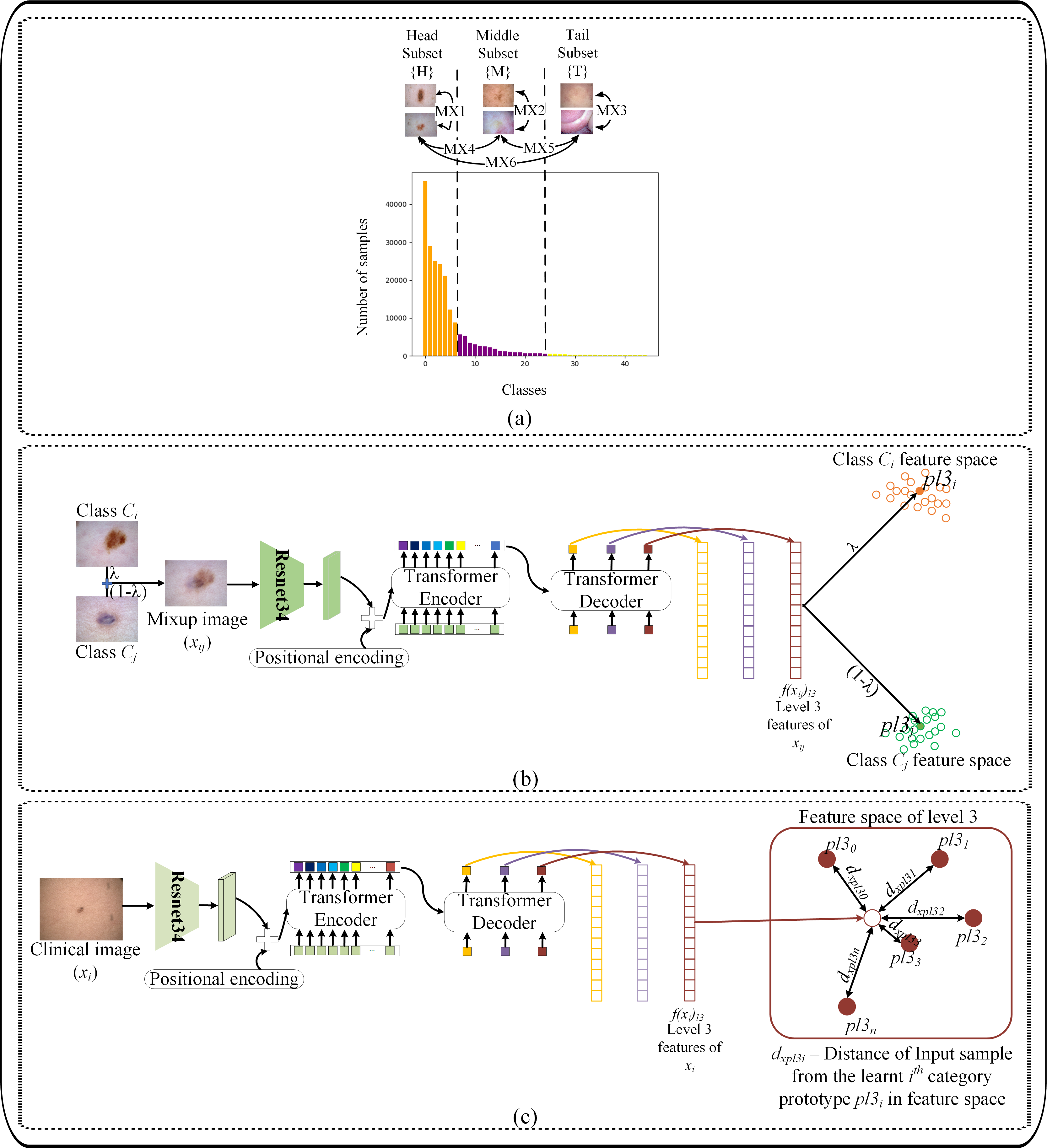}
\caption{Details about the components of our framework: (a) Intra-subset and Inter-subset mixup strategies; (b) Integration of prototype learning and mixup for level 3 categories; (c) Clinical Triage recommendation.}\label{figa1}
\end{figure}

To address this issue, we design \textcolor{black}{mixup strategies} that target specific \textcolor{black}{subsets} of the dataset with the mixup technique. \textcolor{black}{The partition of our In-Distribution (ID) categories into Head ($H \subset C$), Middle ($M \subset C$), and Tail ($T \subset C$) allows us to formulate three intra-subset mixup strategies (both images come from the categories within the same subset) and three inter-subset mixup strategies (each image comes from a category within a different subset).} We denote the intra-subset strategies as MX1, MX2, MX3, and the inter-subset strategies as MX4, MX5, and MX6, as depicted in Fig~\ref{figa1}(a).

\textcolor{black}{We evaluated the different mixup strategies for the OOD detection capability and the results are shown in Table~\ref{mixup-strategies}}. We found that mixup strategies that include only the head subset $\{H\}$ (MX1) perform suboptimally in comparison to not utilizing any mixup strategy \textcolor{black}{at all} (see Table~\ref{mixup-strategies} \textcolor{black}{for OOD detection}). This is mainly attributed to the prevalence of samples coming primarily from the head categories. \textcolor{black}{We discovered that this head subset dominance is effectively mitigated by focusing on the mixup between the categories of the middle and tail subsets (MX5 Middle-Tail mixup strategy), which yields the best result for OOD detection. To further enhance the decision boundaries for OOD detection and more importantly to develop the capability of clinical triage, we integrated the Middle-Tail Mixup strategy (MX5) with prototype learning, discussed further in section~\ref{b13}.}

\begin{table}[t!]
\centering
\caption{Performance evaluation of proposed mixup strategies for OOD detection on OOD(17cl) set for clinical input image modality}\label{tabb1}
\resizebox{0.7\textwidth}{!}{\begin{tabular}{c|c}
\hline
Mixup Strategy&OOD (AUROC\%)\\
\hline
Without Mixup (Only Hierarchical) & 66.84 \\
Standard Mixup & 70.86 \\
\hline
H-H Intrasubset (MX1) & 65.14 \\
M-M Intrasubset (MX2) & 76.16 \\
T-T Intrasubset (MX3) & 76.24 \\
\hline
H-M Intersubset (MX4) & 74.84 \\
\textbf{M-T Intersubset (MX5)} & \textbf{79.47} \\
H-T Intersubset (MX6) & 74.98  \\
\hline
\end{tabular}}\label{mixup-strategies}
\end{table}





\subsubsection{Prototype learning}\label{b12}
Prototype learning~\cite{yang2018robust} has proven to increase the robustness of classifiers by introducing the prototype loss for reducing the intra-class distance variation and increasing the inter-class distance. A conventional prototype loss consists of the mean squared error loss and distance-based cross entropy loss between the latent features from a model (encoder/decoder) and the corresponding prototypes of all the categories. \textcolor{black}{However, for our scenario, as we have inter-subset mixup samples (Middle-Tail Mixup samples), we modify and improve this well-established technique by building a custom loss function that considers the weight of the two categories which make the mixup sample. This integration with mixup strategy and learning with our modified loss function enables further separation of the decision boundaries between the categories as well as making the distrbbution within the categories more compact. We explain the technical details of this integration below}.

\subsubsection{Integration of Mixup and Prototype Learning}\label{b13}
Fig~\ref{figa1}(b) illustrates the integration of prototype learning and mixup in our framework. Firstly, a mixup image is generated from two independent images from different class categories ${C_i,C_j}$, with weights $\lambda$ and $(1-\lambda)$ respectively. This image is then input into our \textcolor{black}{Hierarchical module (CNN-Transformer Encoder-Decoder network)}, which outputs three distinct latent features corresponding to each level - level 1, level 2, and level 3. Given that a mixup image is used as the input for the framework, the corresponding loss functions must adapt accordingly. \textcolor{black}{For levels 1 and 2, we don't employ prototype learning, and thus a weighted cross-entropy loss is applied, based on the weight of the mixup of each image. This  is given by the following equations.}

\begin{equation}
\mathcal{L}_{mixup}=\lambda  \mathcal{L}_{CE}(f(x_{i})_{l1},y_{il1}) +(1-\lambda) \mathcal{L}_{CE}(f(x_{j})_{l1},y_{jl1})
\end{equation}

where $f(x_{i})_{l1}$, $f(x_{j})_{l1}$ represent the latent features of the images $x_i$ and $x_j$ for level 1 respectively; and $y_{il1}$, $y_{jl1}$ represent the ground truth level 1 labels for the images $x_i$ and $x_j$ respectively.

\begin{equation}
\mathcal{L}_{mixup}=\lambda  \mathcal{L}_{CE}(f(x_{i})_{l2},y_{il2}) +(1-\lambda) \mathcal{L}_{CE}(f(x_{j})_{l2},y_{jl2})
\end{equation}

where $f(x_{i})_{l2}$, $f(x_{j})_{l2}$ represent the latent features of the images $x_i$ and $x_j$ for level 2 respectively; and $y_{il2}$, $y_{jl2}$ represent the ground truth level 2 labels for the images $x_i$ and $x_j$ respectively.

\textcolor{black}{For level 3, as we employ protoype learning, we build a custom prototype-mixup loss function which is also based on the weight from each category in the created mixup image.} This is given by the equations below.

\begin{equation}
\mathcal{L}_{mse\|mixup} = \lambda\|f(x_{ij})_{l3}-pl3_{i}\|^{2} + (1-\lambda)\|f(x_{ij})_{l3}-pl3_{j}\|^{2}
\end{equation}

\begin{equation}
\mathcal{L}_{dce\|mixup} = \lambda\mathcal{L}_{CE}(d_{xpl3i},y_{il3}) + (1-\lambda)\mathcal{L}_{CE}(d_{xpl3j},y_{jl3})
\end{equation}

\begin{equation}
\mathcal{L}_{protmix} = \lambda\mathcal{L}_{DCE} + \lambda_{MSE}\mathcal{L}_{MSE}
\end{equation}

where $f(x_{ij})_{l3}$ are the latent features of the mixup image $x_{ij}$; $pl3_{i}$ and $pl3_{j}$ are the learnt prototypes of the $i^{th}$ and the $j^{th}$ categories respectively; $d_{xpl3i}$, $d_{xpl3j}$ is the distance between the input image and the learnt prototypes $pl3_{i}$ and $pl3_{j}$ in the latent space; $y_{il3}$ and $y_{jl3}$ are the ground truth labels for the images $x_i$ and $x_j$ respectively; and $\mathcal{L}_{CE}$ is the standard cross-entropy loss.

Note that we don't employ mixup for the images coming from the head subset $\{H\}$ of the dataset as explained earlier. For the head subset images, a standard cross entropy loss is employed for level 1 and level 2, whereas standard prototype loss is employed for level 3.

\subsection{Selection of Out-of-Distribution Detection Threshold}\label{ood-thresh}

For determining whether an input is an Out-of-Distribution (OOD) image, we compare the confidence score of the image for level 3 to a pre-defined threshold $t_{ood}$. If the confidence score of the image for level 3 is less than the threshold, it is cautioned as an OOD image. We determined the OOD threshold $t_{ood}$ based on the values of FPR and TPR~\cite{kong2020sde}.

\textbf{TPR} corresponds to the True Positive Rate and it is based on the binary classification of whether the input image is an In-Distribution (ID) or OOD based on the confidence score of level 3. On the other hand, \textbf{FPR} corresponds to the False Positive Rate and is calculated contrariwise. Both TPR and FPR are calculated as below.

\begin{equation}
TPR=\dfrac{TP}{TP+FN}
\end{equation}

\begin{equation}
FPR=\dfrac{FP}{FP+TN}
\end{equation}

where $TP$ represents True Positive, $TN$ represents True Negative, $FP$ represents False Positive, and $FN$ represents False Negative for the test images which are either detected as ID or OOD based on their confidence score and the corresponding threshold $t_{ood}$. 

We calculate the TPR on the ID test set images and the FPR on the OOD images by varying the threshold in the range of \{0,1\} with an increment of 0.01. \textcolor{black}{We can then select the OOD operating threshold $t_{ood}$ as the threshold value where a desired trade-off between TPR and FPR is required for clinical decision making.}

\subsection{Selection of Clinical Triage Recommendation Threshold}\label{secA3}

\textcolor{black}{Clinical images are sufficient for diagnosis of many skin lesions. Thus, we develop a clinical triage recommendation functionality which suggests whether to pursue the acquisition of the dermoscopic image for a more accurate diagnosis.} Note that if the input image is detected as an Out-of-Distribution (OOD), we always pursue the additional dermoscopic imaging to confirm the OOD detection. For ID images, our clinical triage strategy utilizes the feature distance between the input image and all the learnt prototypes of the In-Distribution (ID) categories of level 3. This is explained below and shown in Fig~\ref{figa1}(c).

\textcolor{black}{During inference}, once the image is passed through the CNN-Transformer Encoder-Decoder network, it's level 3 latent features are then used to calculate it's distance from all the learnt prototypes (also in the latent space) of the ID categories of level 3. We then select the nearest prototype to the image in the latent space based on the calculated distance. If the distance to the nearest prototype is more than a pre-defined threshold ($t_{triage}$), it implies that the framework is skeptical about the nearest prototype category. Hence, it will be recommended to capture the corresponding dermoscopic image for a more accurate diagnosis. Contrary, if distance to the nearest prototype is less than $t_{triage}$, it is not necessary to capture the dermoscopic image. Below, we describe the process of determining the threshold for triage.

The computation for triage threshold is carried out during the \textbf{inference} on the \textbf{validation} set of the ID-categories after the framework has been trained. First, for each \textbf{correctly predicted image} in the validation ID set, we calculate the \textbf{minimum} distance of the image from all the prototypes i.e. we choose the nearest prototype to this image and calculate it's distance. This is given by the equations below.


\begin{equation}
N_{cvl} = \{ x_{i},if\;\widehat{y}_{il3}= y_{il3}\}
\end{equation}

where $N_{cvl}$ is a subset of the validation ID test set ($N_{cvl} \subset V$) based on whether the predicted level 3 category ($\widehat{y}_{il3}$) of the image ($x_{i}$) matches the ground truth ($y_{il3}$).   


\begin{equation}
D_{xpl3min} = \{ min(d_{xipl30},d_{xipl31},d_{xipl32},...d_{xipl3n}), \forall x_{i} \in N_{cvl}\}
\end{equation}

where $n+1$ is the total number of ID categories; $D_{xpl3min}$ is the set containing the distance of the image to the nearest learnt prototype; and $d_{xipl3j}$ is the distance of an image $x_{i}$ to a prototype of category $j$ in the level 3 latent space, which is given by the equation below.

\begin{equation}
d_{xipl3j} = \|f(x_{i})_{l3}-pl3_{j}\|^{2}
\end{equation}

Lastly, we compute the threshold ($t_{triage}$) as the \textbf{mean} of all the minimum distances.


\begin{equation}
t_{triage} = \dfrac{\sum ^{N}_{k=0}D_{xpl3min}}{N}
\end{equation}

where $N$ is the total number of samples in the $N_{cvl}$ set and $t_{triage}$ is the calculated threshold for recommending clinical triage. The value of $t_{triage}$ is then utilized to make the decision of whether to acquire the additional dermoscopic image which was explained earlier.

\section{\textcolor{black}{Analysis of Confidence Score Distribution for Level 1, Level 2, and Level 3}}\label{lev1lev2-conf}
\textcolor{black}{To evaluate the qualitative effectiveness of our Mixup and Prototype Learning (MPL) strategy on the confidence scores of the ID and OOD distributions, we conducted additional analysis for the confidence score distributions for level 1, level 2, and level 3. For an ideal model, we expect to see a clear separation between the distributions of confidence scores for ID (higher in confidence) and OOD images (lower in confidence). This analysis is shown below for both - ``Hierarchical'' module and ``Hierarchical + MPL'' strategy across the three input image modalities.}

\begin{itemize}
    \item High-level (Level 1): We show the confidence score distribution for ID, OOD(17cl) and OOD(Unk) for level 1 in the left half of Fig~\ref{fig-conf-lev1-lev2}. Since level 1 only comprises two categories, it is observed that both the methods (Hierarchical only and Hierarchical + MPL) have a high degree of confidence, with score distributions greater than 0.5 for any given test image across all three input image settings. Therefore, detecting an Out-of-Distribution image using only the level 1 confidence score is extremely challenging. This mirrors the clinical situation where all skin lesions, regardless of how unknown or rare they are, will either be categorized as benign or malignant.

\begin{figure}%
\centering
\includegraphics[width=0.90\textwidth]{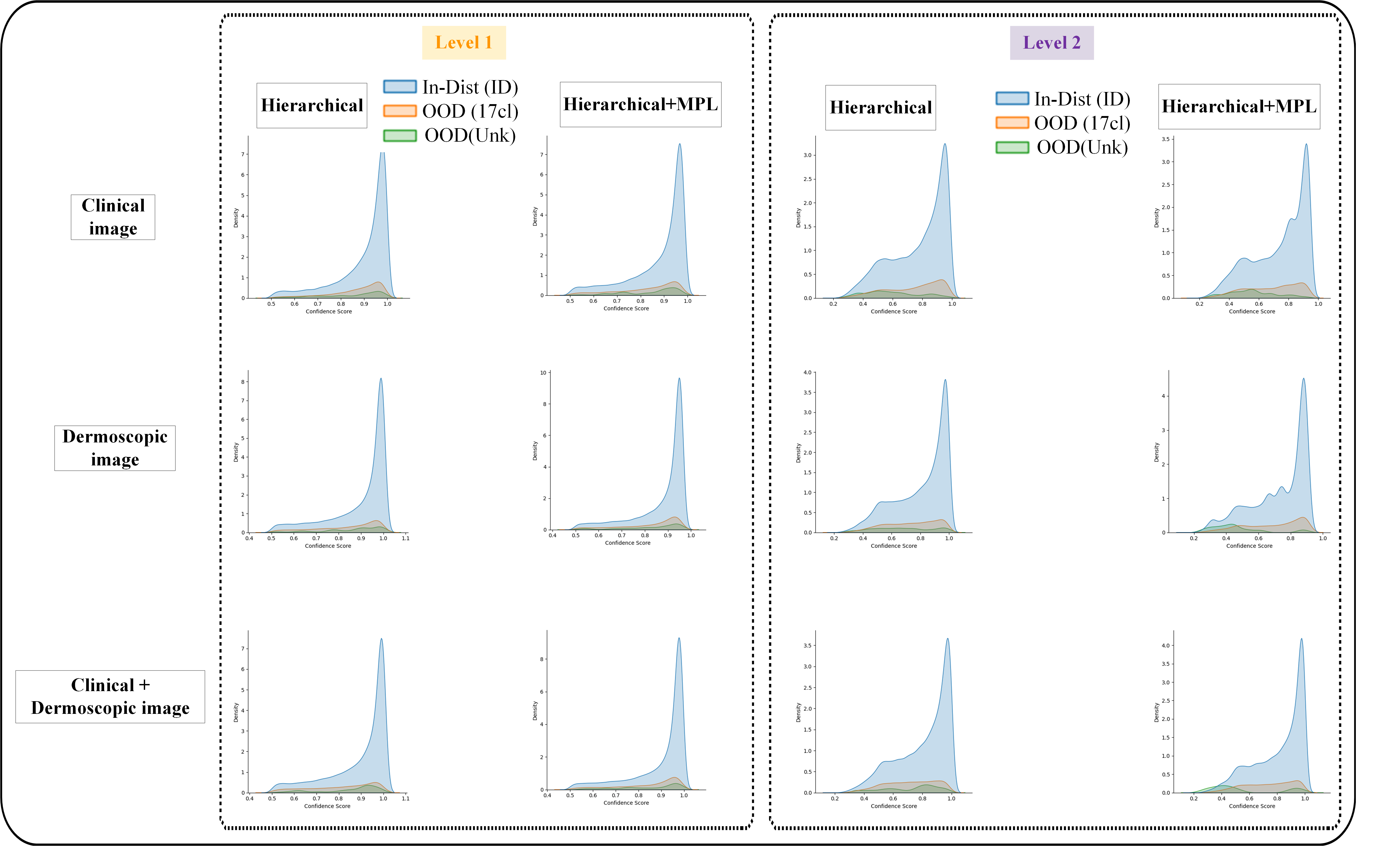}
\caption{Analyzing confidence score distributions of ID, OOD(17cl), and OOD(Unk) for Hierarchical module and Hierarchical + MPL strategy across the two hierarchical levels. Left: Level 1 confidence score distribution; Right: Level 2 confidence score distribution}\label{fig-conf-lev1-lev2}
\end{figure}

    \item Mid-level (Level 2): \textcolor{black}{The confidence score distribution for level 2 is portrayed in the right half of Fig~\ref{fig-conf-lev1-lev2}, where the distribution shows a slight skew towards being less confident (i.e. towards the left) compared to that of level 1.} Nonetheless, a clinical image alone is not sufficient to discern between In-Distribution and Out-of-Distribution images \textcolor{black}{even when based on level 2 distribution}. Examining the distribution for the dermoscopic images in the Hierarchical + MPL method, we note a slight leftward shift in the OOD (Unk) distribution. Its peak now lies somewhere between 0.2 and 0.5, as opposed to the distribution in the Hierarchical prediction method alone, \textcolor{black}{indicating that} the Hierarchical + MPL framework exhibits \textcolor{black}{slightly less} confidence for the OOD (Unk) images.

    As dermoscopic images contain finer and detailed features of the skin lesion, building an OOD detection capability becomes more feasible compared to relying solely on clinical images for \textcolor{black}{level 2}. However, the distribution of OOD (17cl) still lies within a high confidence probability zone even for the Hierarchical + MPL method, \textcolor{black}{reflecting that the majority of OOD images will still fit into one of the level 2 categories.} We also observe that \textcolor{black}{even by employing} the integrated clinical + dermoscopic image, \textcolor{black}{it does not separate the ID and OOD distributions apart} for level 2.

    \item \textcolor{black}{Low-level (Level 3): Left of fig~\ref{fig4} represents the confidence score distributions for ID, OOD (17cl) and OOD (Unk), whereas in the right half we present a more in-depth analysis amongst the Head ${H}$, Middle ${M}$, and Tail ${T}$ subsets of the ID distributions. The integration of MPL strategy with the Hierarchical module improves confidence levels of ID images and enhances OOD detection for OOD(17cl) and OOD(Unk) images across all image modalities (clinical, dermoscopic, clinical+dermoscopic). The ID distribution shifts to the right (higher confidence), while both OOD(17cl) and OOD(Unk) distributions shift to the left (lower confidence) compared to Hierarchical module alone. This validates the effectiveness of integrating the MPL method. Moreover, the OOD(17cl) distribution leans slightly more to the left than the OOD(Unk) distribution, corroborating that the MPL method is more effective in detecting the OOD(17cl) lesions compared to the commonly encountered OOD(Unk) images.}

\begin{figure}%
\centering
\includegraphics[width=0.9\textwidth]{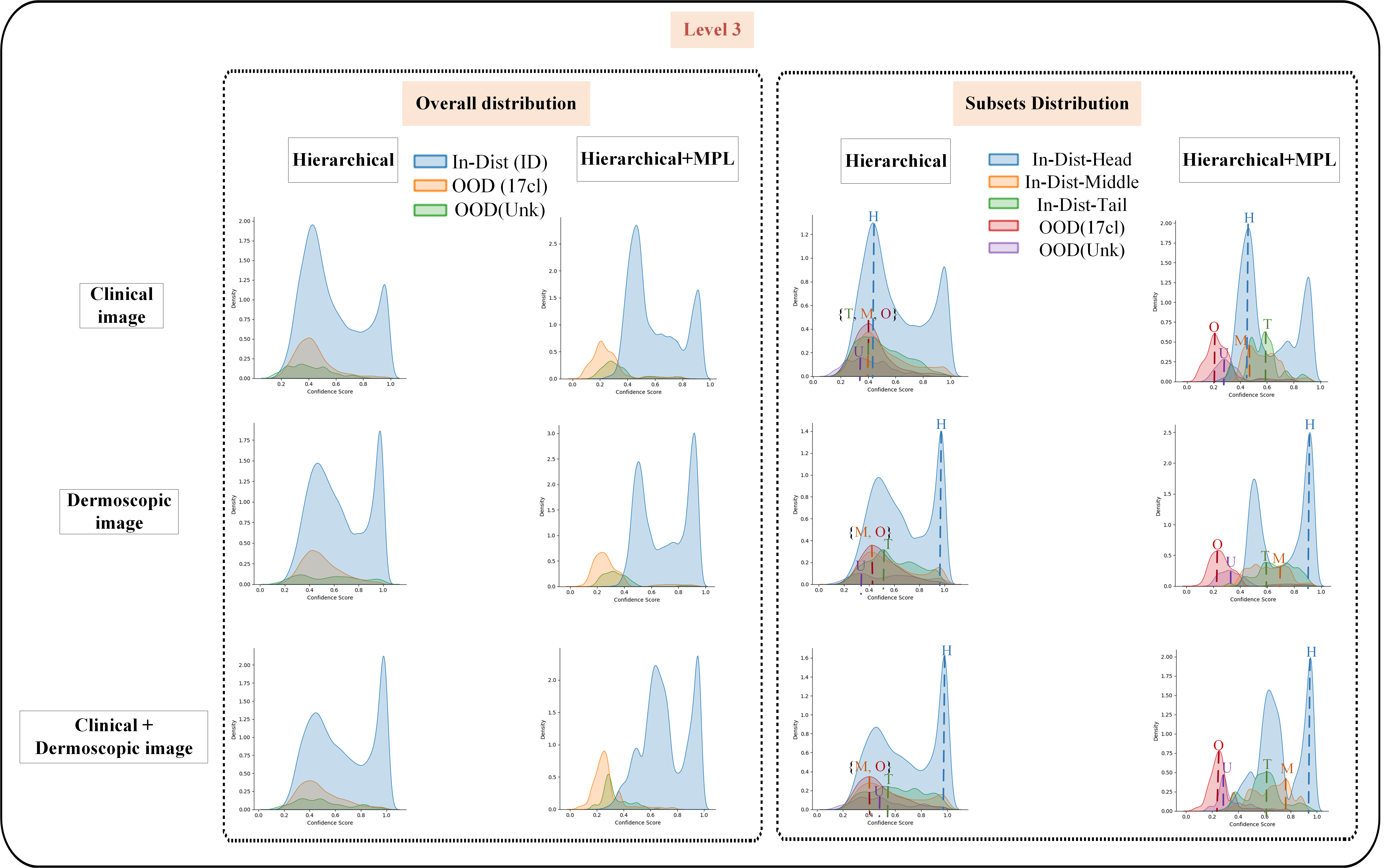}
\caption{Analyzing level 3 confidence score distributions for Hierarchical module and Hierarchical + MPL strategy. Left: Level 3 scores for ID, OOD(17cl) and OOD(Unk) subsets; Right: Detailed Level 3 subset scores (Head, Middle, Tail, OOD(17cl), OOD(Unk))}\label{fig4}
\end{figure}

	In the right half of Fig~\ref{fig4}, we note that the peaks of all three subsets $\{H\}$, $\{M\}$, $\{T\}$ of the ID set shift significantly to right with the Hierarchical + MPL strategy. The $\{M\}$ and $\{T\}$ subsets show particularly notable shifts because MPL strategy specifically focuses on better learning for Middle and Tail subsets. In essence, the MPL method contributes to making the model more confident for these lesser-represented categories of $\{M\}$ and $\{T\}$ subsets, effectively improving OOD detection. \textcolor{black}{This visualization of confidence score distributions for the ID, OOD (17cl) and OOD (Unk) sets further substantiates the high OOD detection performance reported in Table~\ref{tab-ood-eval} for the Hierarchical + MPL strategy.}
 
\end{itemize}

\section{Intra-class distance of Level 1 and Level 2}\label{lev1lev2intraclass}

\begin{figure}[t!]%
\centering
\includegraphics[width=0.865\textwidth]{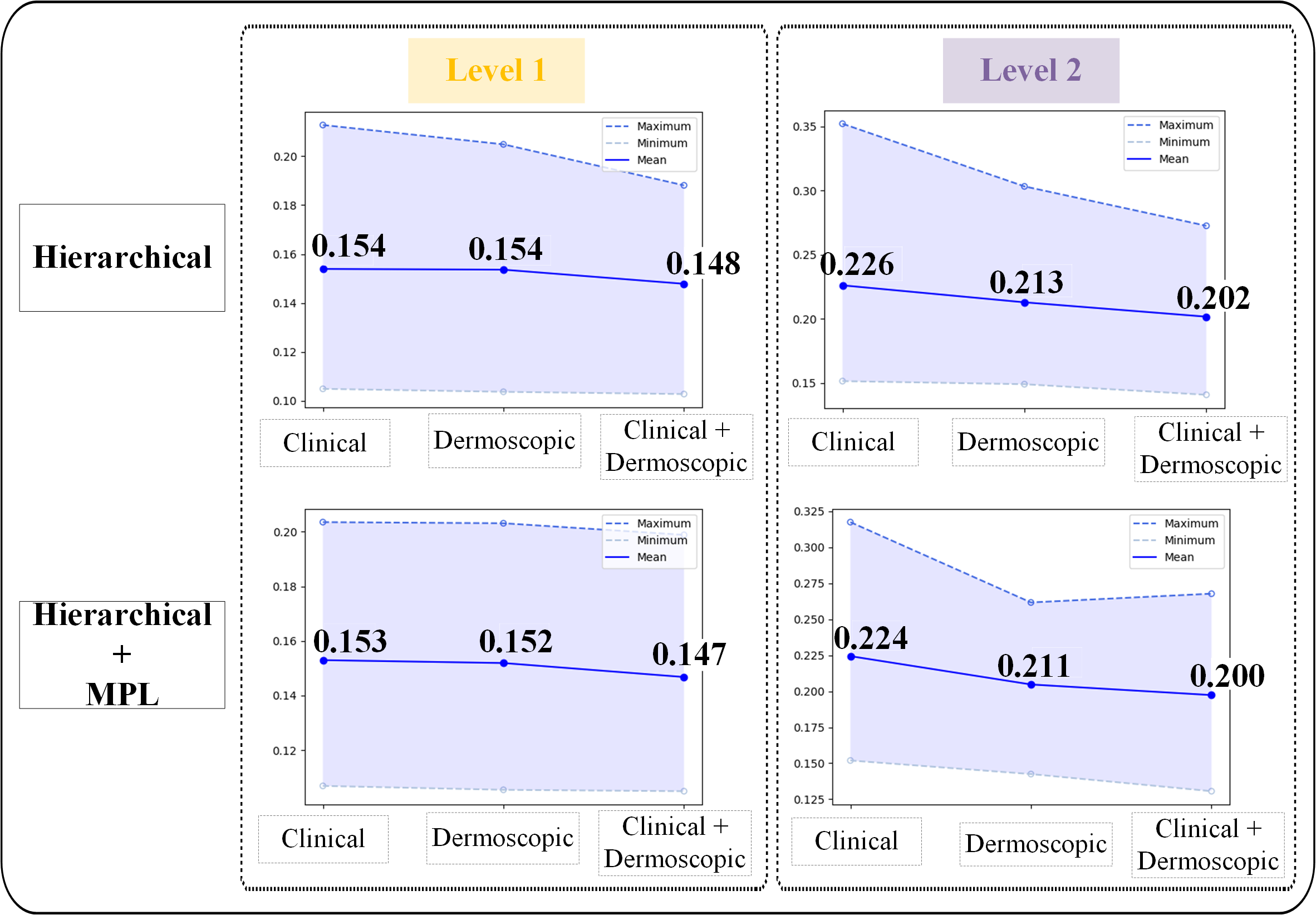}
\caption{Variation of intra-class distances for In-Distribution categories. Maximum, minimum and mean of the categories for hierarchical levels (level 1 and level 2) across the three input image modalities for both the Hierarchical and Hierarchical + MPL methods.}\label{fig-intraclass-lev1lev2}
\end{figure}

\textcolor{black}{In addition to the level 3 analysis which was presented in section~\ref{intraclass-interclass-all}, we also conducted the intra-class distance analysis for level 1 and level 2.} Observations from Fig~\ref{fig-intraclass-lev1lev2} reveal that the \textbf{mean} intra-class distance is lower for level 1 compared to level 2 for both Hierarchical and Hierarchical + MPL methods across the three input image modalities. Additionally, \textbf{mean} intra-class distances for both level 1 and level 2 are lower than that of level 3 (see Fig~\ref{fig6}(a)). Given that level 1 has only two categories (benign and malignant), the images are easier to separate in the feature space, thereby facilitating high-level prediction. With the increase in category count to eight in level 2, images within the same category are distributed over a wider area, thus increasing the intra-class variation. Hence, we observe a rise in mean intra-class distance for level 2.

Across the input image modalities, we notice that the clinical + dermoscopic image has the smallest intra-class distance, followed by the dermoscopic image alone, and the clinical image alone has the largest intra-class distance. \textcolor{black}{This analysis reinforces the observation that the integration of clinical + dermoscopic image is superior to dermoscopic image alone, which in turn are both better than clinical image alone for diagnosis of ID samples.} However, it is noteworthy that the Hierarchical + MPL method diminishes intra-class distance across both the level 1 and level 2, with a moderate impact on level 2, and minimal impact on level 1, owing to MPL's learning focus on level 3.

\section{Melanoma analysis}
We conduct additional analysis specifically for the level 3 category of malignant:melanoma:melanoma, considering the severity and clinical importance of this type of skin cancer. This analysis has provided more nuanced insights that are particularly valuable for diagnosing melanoma. Specifically, for this level 3 category of melanoma, we analyse the intra-class distances, inter-class distances and \textcolor{black}{performance} of the clinical triage strategy shown in Fig~\ref{fig-intraclass-melanoma}(a), Fig~\ref{fig-intraclass-melanoma}(b), and Fig~\ref{fig-triage-melanoma} respectively.

\subsection{Intra-class and Inter-class distances}
\begin{figure}[t!]%
\centering
\includegraphics[width=0.95\textwidth]{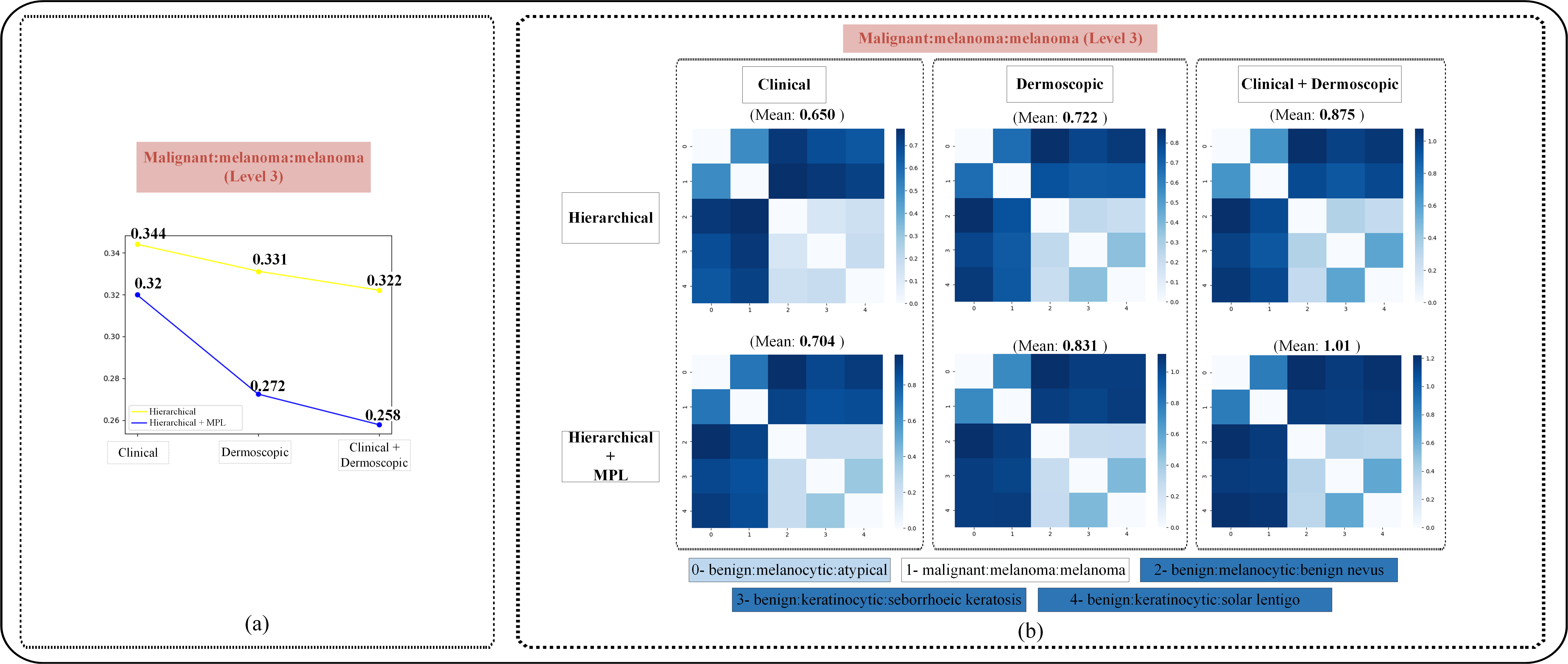}
\caption{Performance of our proposed framework for melanoma classification analysis: (a) Variation of intra-class distance for the malignant:melanoma:melanoma category across the three input image modalities for both the Hierarchical and Hierarchical + MPL methods; (b) Inter-class distance matrices between the top-5 predicted categories (based on the confidence score) for the test samples belonging to the level 3 category of malignant:melanoma:melanoma. A darker color indicates a higher distance.}\label{fig-intraclass-melanoma}
\end{figure}

Fig~\ref{fig-intraclass-melanoma}(a) illustrates the intra-class distances for malignant:melanoma:melanoma across the three input image modalities (clinical, dermoscopic, and a combination of clinical and dermoscopic) for both Hierarchical and Hierarchical + MPL methods. In line with the trend seen for all level 3 categories in Fig~\ref{fig6}(a), the \textcolor{black}{integration of MPL strategy with the Hierarchical method} leads to a considerable reduction in intra-class distance for diagnosis based on the dermoscopic image or a combination of clinical and dermoscopic images. However, the reduction is less pronounced in diagnosis based on clinical images alone, \textcolor{black}{reinforcing the importance of dermoscopic images in diagnosing malignant:melanoma:melanoma}.

\noindent\textbf{Clinical interpretation}: This reflects the clinical reality, where a melanoma is often an ‘ugly duckling’, i.e. it stands out macroscopically from the other lesions. However, dermoscopic images help rule out other ‘ugly duckling’ lesions for example seborrheic keratosis, a benign lesion common in older adults.

\textcolor{black}{Fig~\ref{fig-intraclass-melanoma}(b) shows the inter-class distance matrices between the top 5 predicted categories for all test images of melanoma for both Hierarchical and Hierarchical + MPL methods across all three image modalities. Despite showing similar overall results to the ID categories in Fig~\ref{fig6}(b), several observations can be made from Fig~\ref{fig-intraclass-melanoma}(b). By integrating the MPL strategy, there is a significant increase in the mean inter-class distance between the top 5 categories for diagnoses based on dermoscopic images or a combination of clinical and dermoscopic images, compared to diagnoses based solely on clinical images. This is coherent with earlier observations regarding the inter-class distance for all the ID categories. In addition, the malignant:melanoma:melanoma category is farther distanced from the benign categories, including benign:melanocytic:benign nevus, benign:keratinocytic:seborrheic keratosis, and benign:keratinocytic:solar lentigo \textcolor{black}{compared to atypical naevi. This suggests that on average, it is harder to differentiate between melanoma and atypical naevi}. This distance widens when shifting from only clinical images to dermoscopic images or a combination of both, and further increases when employing the Hierarchical + MPL method, validating the effectiveness of MPL method and utilizing both dermoscopic and clinical images for diagnosis.}

\noindent\textbf{Clinical interpretation}: The fact that the interclass distance of benign:melanocytic:atypical category is relatively closer to malignant:melanoma:melanoma, suggests that benign atypical naevi could be confused as melanoma and vice-versa. It should be noted that the ground truth labels for this dataset are based on images alone, however in practice ground truth would consider the histopathological diagnosis following excision. However, even considering a histopathological diagnosis, there is current debate as to whether melanoma in situ should in fact be classified as atypical or dysplastic as there is little to no evidence of melanoma in situ causing metastasis or death~\cite{semsarian2022we}.

\subsection{Clinical Triage}

\begin{figure}[t!]%
\centering
\includegraphics[width=0.95\textwidth]{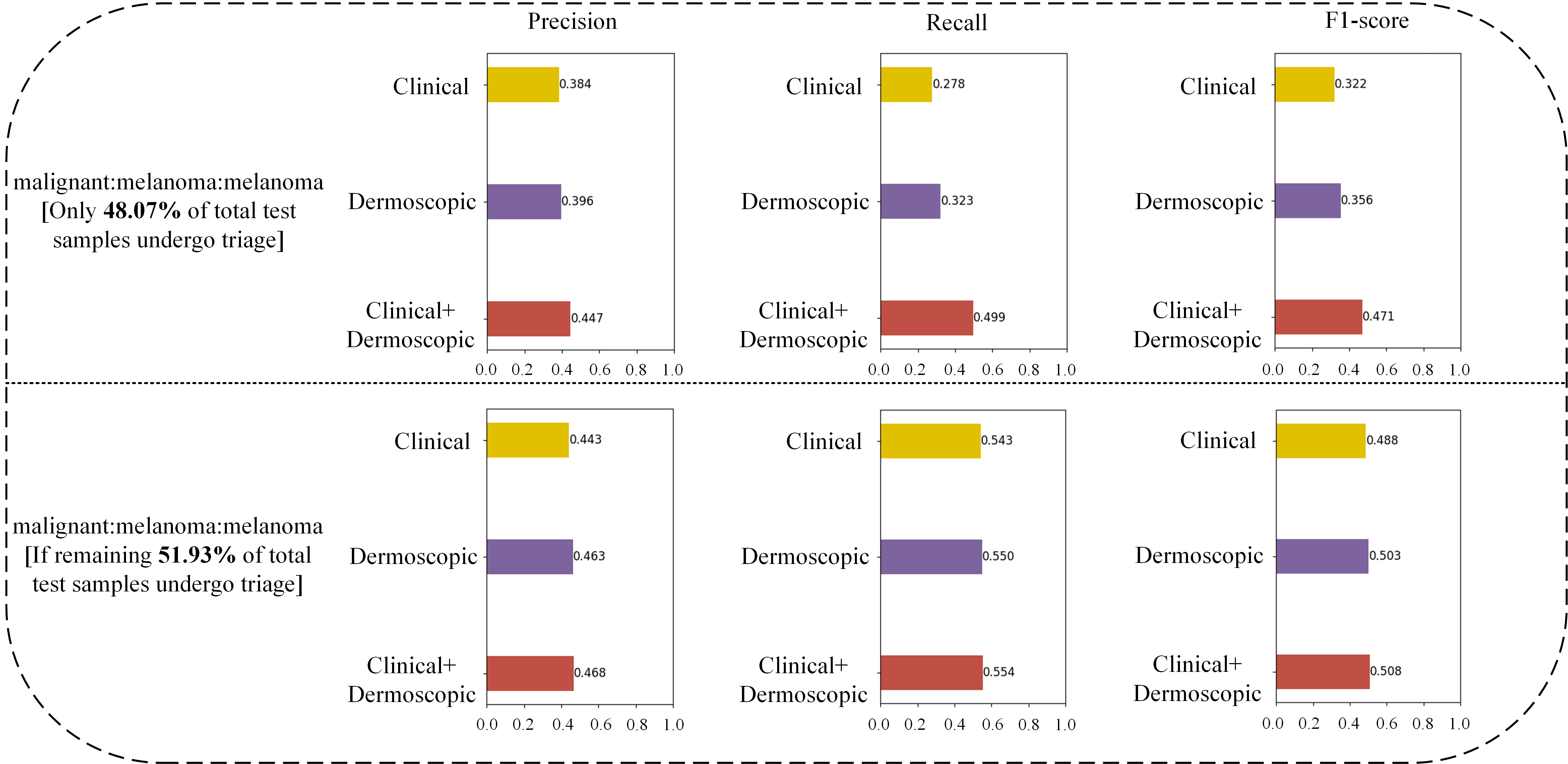}
\caption{Performance Evaluation (Precision, Recall and F1-score) of Clinical Triage strategy for the level 3 category of malignant:melanoma:melanoma. Comparison of performance increment. Top: Represents the melanoma test images which were recommended for clinical triage and if pursued the corresponding dermoscopic image only based and combined clinical + dermoscopic image-based performance. Bottom: Represents melanoma test images which were not recommended for clinical triage, nevertheless if pursued the corresponding dermoscopic image only based and combined clinical + dermoscopic image-based performance.}\label{fig-triage-melanoma}
\end{figure}

Fig~\ref{fig-triage-melanoma} displays the results of the clinical triage module for the specific category of malignant:melanoma:melanoma, which shows a similar trend to the ID test set. However, there are certain important differences that could be noticed for the category of melanoma. Notably, approximately 48\% of melanoma images are recommended for triage, compared to only 38\% for the overall ID test set. This implies that for nearly half of the melanoma cases, a clinical image alone may not provide a sufficient basis for an accurate diagnosis. This also falls in line with the analysis that was made for the intra-class and inter-class distances for melanoma earlier. Furthermore, diagnoses based solely on a dermoscopic image prove to be much more accurate than those based solely on a clinical image, and this accuracy increases significantly when both types of images are used, reflecting results seen in clinical practice. The \textcolor{black}{top} half of Fig~\ref{fig-triage-melanoma} reveals that the precision, recall, and F1-score for melanoma diagnosis improve respectively by {0.012, 0.045, 0.034} and {0.063, 0.221, 0.149} when using only the dermoscopic image and when combining both clinical and dermoscopic images for the triage recommended cases.

Even for the remaining \textbf{52\%} of the not recommended to triage cases, the performance improvement for diagnosing melanoma is slightly higher when diagnosis is based only on the dermoscopic image and on the combination of clinical and dermoscopic images when compared to that of the \textbf{62\%} not recommended cases for the ID categories. In the \textcolor{black}{bottom} half of Fig~\ref{fig-triage-melanoma}, it is evident that the precision, recall, and F1-score improve by {0.002, 0.007, 0.015} and {0.025, 0.011, 0.020} when diagnosis is based only on the dermoscopic image and on the combination of clinical and dermoscopic images, respectively. This further demonstrates that for diagnosing melanoma, acquiring a dermoscopic image is preferred compared to the other ID categories.

\section{Limitations of our proposed system}\label{limitations}
We discuss the limitations associated with our dataset and methods below.

\begin{itemize}
    \item Our current hierarchical framework is inherently confined to three levels, owing to the tri-level hierarchical annotations in our dataset. While we have demonstrated instances where a singular level 3 diagnosis can be misleading and avoided by considering higher and mid-level (level 1 and level 2) predictions, our dataset restricts us from exploring more or fewer levels, and this could be a potential area of exploration. Further, future models could  consider a third ‘precursor’ category for the level one diagnosis which is currently limited to benign or malignant. This could include lesions such as SCC in situ, and would reflect the differences in classification in treatment of lesions such as actinic keratosis, which are considered benign in Australia but malignant in Europe. 
    
    \item The dataset's categorization into head, middle, and tail subsets hinges on the quantity of images in each category and \textcolor{black}{our specific long-tailed distribution.} The cut-off for this partition is hardcoded, reflecting the data distribution in our dataset. Consequently, a distinct dataset distribution would alter the head, middle, and tail segregation, hinging on the specific data distribution. Our methodology's enhanced learning for the middle and tail categories is restricted to datasets with a similar long-tailed distribution. However, for any imbalanced dataset, we propose categorizing subsets such that the least number of categories form the head subset, a moderate amount in the middle subset, and most categories in the tail subset. Future work could also explore splitting based on semantic information, which is beyond the scope of our current research.

    \item The performance of AI models trained on skin lesion images can be influenced by variations in skin tones. \textcolor{black}{Our dataset predominantly originates from local clinics in Australia and New Zealand, thus consisting of individuals of largely European ancestry with an estimated 5\% East Asian skin types and 5-10 per million Indian skin types (largely skin types I-III)}. Our dataset lacks specific skin tone type information (as featured in the Fitzpatrick 17k dataset~\cite{groh2021evaluating}), preventing us from evaluating our model's generalizability  across different skin tones.

    \item Like most methodologies, our model's evaluation and testing are restricted to images captured from the same camera device on which it was trained. AI models trained on images from a specific device can encounter domain shift~\cite{shimodaira2000improving,fogelberg2023domain} issues when tested on external validation image sets from a different camera device. Therefore, our evaluations apply solely to images captured by a specific camera device (as described in the Extension \ref{secDataCol}). \textcolor{black}{However, we do perform a location-based training and testing of our framework, where we found that our framework is able to attend similar results, suggesting generalizability across the locations of captured data.}

    \item The labels in this diagnosis were provided by \textcolor{black}{dermatologists} using the Molemap system, based on dermoscopic and clinical image \textcolor{black}{compared to the gold standard of histopathology analysis. The dermatologists were also provided with sex, age, anatomical location and melanoma history of the patient along with the lesion images for preparing the ground truth}. The hierarchical labelling is somewhat clinician dependent as the clinicians chose their diagnosis based on drop down menus, and have the ability to provide a simple diagnosis i.e. benign naevus v/s bcc, or to provide a more detailed diagnosis i.e. compound naevus v/s pigmented bcc.

    \item The presented methods, results and analysis are based on static images. Due to lack of longitudinal data for each skin lesion in our dataset, it is not possible to integrate the ability to analyse the changes in skin lesion over time which provides more diagnostic information and could accurately track the progression of a lesion.
    
\end{itemize}




\end{appendices}


\bibliography{sn-bibliography}


\begin{thebibliography}{53}
\ifx \bisbn   \undefined \def \bisbn  #1{ISBN #1}\fi
\ifx \binits  \undefined \def \binits#1{#1}\fi
\ifx \bauthor  \undefined \def \bauthor#1{#1}\fi
\ifx \batitle  \undefined \def \batitle#1{#1}\fi
\ifx \bjtitle  \undefined \def \bjtitle#1{#1}\fi
\ifx \bvolume  \undefined \def \bvolume#1{\textbf{#1}}\fi
\ifx \byear  \undefined \def \byear#1{#1}\fi
\ifx \bissue  \undefined \def \bissue#1{#1}\fi
\ifx \bfpage  \undefined \def \bfpage#1{#1}\fi
\ifx \blpage  \undefined \def \blpage #1{#1}\fi
\ifx \burl  \undefined \def \burl#1{\textsf{#1}}\fi
\ifx \doiurl  \undefined \def \doiurl#1{\url{https://doi.org/#1}}\fi
\ifx \betal  \undefined \def \betal{\textit{et al.}}\fi
\ifx \binstitute  \undefined \def \binstitute#1{#1}\fi
\ifx \binstitutionaled  \undefined \def \binstitutionaled#1{#1}\fi
\ifx \bctitle  \undefined \def \bctitle#1{#1}\fi
\ifx \beditor  \undefined \def \beditor#1{#1}\fi
\ifx \bpublisher  \undefined \def \bpublisher#1{#1}\fi
\ifx \bbtitle  \undefined \def \bbtitle#1{#1}\fi
\ifx \bedition  \undefined \def \bedition#1{#1}\fi
\ifx \bseriesno  \undefined \def \bseriesno#1{#1}\fi
\ifx \blocation  \undefined \def \blocation#1{#1}\fi
\ifx \bsertitle  \undefined \def \bsertitle#1{#1}\fi
\ifx \bsnm \undefined \def \bsnm#1{#1}\fi
\ifx \bsuffix \undefined \def \bsuffix#1{#1}\fi
\ifx \bparticle \undefined \def \bparticle#1{#1}\fi
\ifx \barticle \undefined \def \barticle#1{#1}\fi
\bibcommenthead
\ifx \bconfdate \undefined \def \bconfdate #1{#1}\fi
\ifx \botherref \undefined \def \botherref #1{#1}\fi
\ifx \url \undefined \def \url#1{\textsf{#1}}\fi
\ifx \bchapter \undefined \def \bchapter#1{#1}\fi
\ifx \bbook \undefined \def \bbook#1{#1}\fi
\ifx \bcomment \undefined \def \bcomment#1{#1}\fi
\ifx \oauthor \undefined \def \oauthor#1{#1}\fi
\ifx \citeauthoryear \undefined \def \citeauthoryear#1{#1}\fi
\ifx \endbibitem  \undefined \def \endbibitem {}\fi
\ifx \bconflocation  \undefined \def \bconflocation#1{#1}\fi
\ifx \arxivurl  \undefined \def \arxivurl#1{\textsf{#1}}\fi
\csname PreBibitemsHook\endcsname

\bibitem{seth2017global}
\begin{barticle}
\bauthor{\bsnm{Seth}, \binits{D.}},
\bauthor{\bsnm{Cheldize}, \binits{K.}},
\bauthor{\bsnm{Brown}, \binits{D.}},
\bauthor{\bsnm{Freeman}, \binits{E.E.}}:
\batitle{Global burden of skin disease: inequities and innovations}.
\bjtitle{Current dermatology reports}
\bvolume{6}(\bissue{3}),
\bfpage{204}--\blpage{210}
(\byear{2017})
\end{barticle}
\endbibitem

\bibitem{kimball2008us}
\begin{barticle}
\bauthor{\bsnm{Kimball}, \binits{A.B.}},
\bauthor{\bsnm{Resneck~Jr}, \binits{J.S.}}:
\batitle{The us dermatology workforce: a specialty remains in shortage}.
\bjtitle{Journal of the American Academy of Dermatology}
\bvolume{59}(\bissue{5}),
\bfpage{741}--\blpage{745}
(\byear{2008})
\end{barticle}
\endbibitem

\bibitem{feng2018comparison}
\begin{barticle}
\bauthor{\bsnm{Feng}, \binits{H.}},
\bauthor{\bsnm{Berk-Krauss}, \binits{J.}},
\bauthor{\bsnm{Feng}, \binits{P.W.}},
\bauthor{\bsnm{Stein}, \binits{J.A.}}:
\batitle{Comparison of dermatologist density between urban and rural counties in the united states}.
\bjtitle{JAMA dermatology}
\bvolume{154}(\bissue{11}),
\bfpage{1265}--\blpage{1271}
(\byear{2018})
\end{barticle}
\endbibitem

\bibitem{webster2020virtual}
\begin{barticle}
\bauthor{\bsnm{Webster}, \binits{P.}}:
\batitle{Virtual health care in the era of covid-19}.
\bjtitle{The Lancet}
\bvolume{395}(\bissue{10231}),
\bfpage{1180}--\blpage{1181}
(\byear{2020})
\end{barticle}
\endbibitem

\bibitem{liu2020deep}
\begin{barticle}
\bauthor{\bsnm{Liu}, \binits{Y.}},
\bauthor{\bsnm{Jain}, \binits{A.}},
\bauthor{\bsnm{Eng}, \binits{C.}},
\bauthor{\bsnm{Way}, \binits{D.H.}},
\bauthor{\bsnm{Lee}, \binits{K.}},
\bauthor{\bsnm{Bui}, \binits{P.}},
\bauthor{\bsnm{Kanada}, \binits{K.}},
\bauthor{\bparticle{de} \bsnm{Oliveira~Marinho}, \binits{G.}},
\bauthor{\bsnm{Gallegos}, \binits{J.}},
\bauthor{\bsnm{Gabriele}, \binits{S.}}, \betal:
\batitle{A deep learning system for differential diagnosis of skin diseases}.
\bjtitle{Nature medicine}
\bvolume{26}(\bissue{6}),
\bfpage{900}--\blpage{908}
(\byear{2020})
\end{barticle}
\endbibitem

\bibitem{ji2021total}
\begin{barticle}
\bauthor{\bsnm{Ji-Xu}, \binits{A.}},
\bauthor{\bsnm{Dinnes}, \binits{J.}},
\bauthor{\bsnm{Matin}, \binits{R.}}:
\batitle{Total body photography for the diagnosis of cutaneous melanoma in adults: a systematic review and meta-analysis}.
\bjtitle{British Journal of Dermatology}
\bvolume{185}(\bissue{2}),
\bfpage{302}--\blpage{312}
(\byear{2021})
\end{barticle}
\endbibitem

\bibitem{kittler2021evolution}
\begin{botherref}
\oauthor{\bsnm{Kittler}, \binits{H.}}:
Evolution of the clinical, dermoscopic and pathologic diagnosis of melanoma.
Dermatology Practical \& Conceptual
\textbf{11}(Suppl 1)
(2021)
\end{botherref}
\endbibitem

\bibitem{esteva2017dermatologist}
\begin{barticle}
\bauthor{\bsnm{Esteva}, \binits{A.}},
\bauthor{\bsnm{Kuprel}, \binits{B.}},
\bauthor{\bsnm{Novoa}, \binits{R.A.}},
\bauthor{\bsnm{Ko}, \binits{J.}},
\bauthor{\bsnm{Swetter}, \binits{S.M.}},
\bauthor{\bsnm{Blau}, \binits{H.M.}},
\bauthor{\bsnm{Thrun}, \binits{S.}}:
\batitle{Dermatologist-level classification of skin cancer with deep neural networks}.
\bjtitle{nature}
\bvolume{542}(\bissue{7639}),
\bfpage{115}--\blpage{118}
(\byear{2017})
\end{barticle}
\endbibitem

\bibitem{haenssle2018man}
\begin{barticle}
\bauthor{\bsnm{Haenssle}, \binits{H.A.}},
\bauthor{\bsnm{Fink}, \binits{C.}},
\bauthor{\bsnm{Schneiderbauer}, \binits{R.}},
\bauthor{\bsnm{Toberer}, \binits{F.}},
\bauthor{\bsnm{Buhl}, \binits{T.}},
\bauthor{\bsnm{Blum}, \binits{A.}},
\bauthor{\bsnm{Kalloo}, \binits{A.}},
\bauthor{\bsnm{Hassen}, \binits{A.B.H.}},
\bauthor{\bsnm{Thomas}, \binits{L.}},
\bauthor{\bsnm{Enk}, \binits{A.}}, \betal:
\batitle{Man against machine: diagnostic performance of a deep learning convolutional neural network for dermoscopic melanoma recognition in comparison to 58 dermatologists}.
\bjtitle{Annals of oncology}
\bvolume{29}(\bissue{8}),
\bfpage{1836}--\blpage{1842}
(\byear{2018})
\end{barticle}
\endbibitem

\bibitem{brinker2019deep}
\begin{barticle}
\bauthor{\bsnm{Brinker}, \binits{T.J.}},
\bauthor{\bsnm{Hekler}, \binits{A.}},
\bauthor{\bsnm{Enk}, \binits{A.H.}},
\bauthor{\bsnm{Klode}, \binits{J.}},
\bauthor{\bsnm{Hauschild}, \binits{A.}},
\bauthor{\bsnm{Berking}, \binits{C.}},
\bauthor{\bsnm{Schilling}, \binits{B.}},
\bauthor{\bsnm{Haferkamp}, \binits{S.}},
\bauthor{\bsnm{Schadendorf}, \binits{D.}},
\bauthor{\bsnm{Holland-Letz}, \binits{T.}}, \betal:
\batitle{Deep learning outperformed 136 of 157 dermatologists in a head-to-head dermoscopic melanoma image classification task}.
\bjtitle{European Journal of Cancer}
\bvolume{113},
\bfpage{47}--\blpage{54}
(\byear{2019})
\end{barticle}
\endbibitem

\bibitem{maron2019systematic}
\begin{barticle}
\bauthor{\bsnm{Maron}, \binits{R.C.}},
\bauthor{\bsnm{Weichenthal}, \binits{M.}},
\bauthor{\bsnm{Utikal}, \binits{J.S.}},
\bauthor{\bsnm{Hekler}, \binits{A.}},
\bauthor{\bsnm{Berking}, \binits{C.}},
\bauthor{\bsnm{Hauschild}, \binits{A.}},
\bauthor{\bsnm{Enk}, \binits{A.H.}},
\bauthor{\bsnm{Haferkamp}, \binits{S.}},
\bauthor{\bsnm{Klode}, \binits{J.}},
\bauthor{\bsnm{Schadendorf}, \binits{D.}}, \betal:
\batitle{Systematic outperformance of 112 dermatologists in multiclass skin cancer image classification by convolutional neural networks}.
\bjtitle{European Journal of Cancer}
\bvolume{119},
\bfpage{57}--\blpage{65}
(\byear{2019})
\end{barticle}
\endbibitem

\bibitem{cruz2013deep}
\begin{bchapter}
\bauthor{\bsnm{Cruz-Roa}, \binits{A.A.}},
\bauthor{\bsnm{Arevalo~Ovalle}, \binits{J.E.}},
\bauthor{\bsnm{Madabhushi}, \binits{A.}},
\bauthor{\bsnm{Gonz{\'a}lez~Osorio}, \binits{F.A.}}:
\bctitle{A deep learning architecture for image representation, visual interpretability and automated basal-cell carcinoma cancer detection}.
In: \bbtitle{International Conference on Medical Image Computing and Computer-assisted Intervention},
pp. \bfpage{403}--\blpage{410}
(\byear{2013}).
\bcomment{Springer}
\end{bchapter}
\endbibitem

\bibitem{codella2018skin}
\begin{bchapter}
\bauthor{\bsnm{Codella}, \binits{N.C.}},
\bauthor{\bsnm{Gutman}, \binits{D.}},
\bauthor{\bsnm{Celebi}, \binits{M.E.}},
\bauthor{\bsnm{Helba}, \binits{B.}},
\bauthor{\bsnm{Marchetti}, \binits{M.A.}},
\bauthor{\bsnm{Dusza}, \binits{S.W.}},
\bauthor{\bsnm{Kalloo}, \binits{A.}},
\bauthor{\bsnm{Liopyris}, \binits{K.}},
\bauthor{\bsnm{Mishra}, \binits{N.}},
\bauthor{\bsnm{Kittler}, \binits{H.}}, \betal:
\bctitle{Skin lesion analysis toward melanoma detection: A challenge at the 2017 international symposium on biomedical imaging (isbi), hosted by the international skin imaging collaboration (isic)}.
In: \bbtitle{2018 IEEE 15th International Symposium on Biomedical Imaging (ISBI 2018)},
pp. \bfpage{168}--\blpage{172}
(\byear{2018}).
\bcomment{IEEE}
\end{bchapter}
\endbibitem

\bibitem{yuan2017automatic}
\begin{barticle}
\bauthor{\bsnm{Yuan}, \binits{Y.}},
\bauthor{\bsnm{Chao}, \binits{M.}},
\bauthor{\bsnm{Lo}, \binits{Y.-C.}}:
\batitle{Automatic skin lesion segmentation using deep fully convolutional networks with jaccard distance}.
\bjtitle{IEEE transactions on medical imaging}
\bvolume{36}(\bissue{9}),
\bfpage{1876}--\blpage{1886}
(\byear{2017})
\end{barticle}
\endbibitem

\bibitem{okuboyejo2013automating}
\begin{bchapter}
\bauthor{\bsnm{Okuboyejo}, \binits{D.A.}},
\bauthor{\bsnm{Olugbara}, \binits{O.O.}},
\bauthor{\bsnm{Odunaike}, \binits{S.A.}}:
\bctitle{Automating skin disease diagnosis using image classification}.
In: \bbtitle{Proceedings of the World Congress on Engineering and Computer Science},
vol. \bseriesno{2},
pp. \bfpage{850}--\blpage{854}
(\byear{2013})
\end{bchapter}
\endbibitem

\bibitem{tschandl2019comparison}
\begin{barticle}
\bauthor{\bsnm{Tschandl}, \binits{P.}},
\bauthor{\bsnm{Codella}, \binits{N.}},
\bauthor{\bsnm{Akay}, \binits{B.N.}},
\bauthor{\bsnm{Argenziano}, \binits{G.}},
\bauthor{\bsnm{Braun}, \binits{R.P.}},
\bauthor{\bsnm{Cabo}, \binits{H.}},
\bauthor{\bsnm{Gutman}, \binits{D.}},
\bauthor{\bsnm{Halpern}, \binits{A.}},
\bauthor{\bsnm{Helba}, \binits{B.}},
\bauthor{\bsnm{Hofmann-Wellenhof}, \binits{R.}}, \betal:
\batitle{Comparison of the accuracy of human readers versus machine-learning algorithms for pigmented skin lesion classification: an open, web-based, international, diagnostic study}.
\bjtitle{The lancet oncology}
\bvolume{20}(\bissue{7}),
\bfpage{938}--\blpage{947}
(\byear{2019})
\end{barticle}
\endbibitem

\bibitem{lopez2017skin}
\begin{bchapter}
\bauthor{\bsnm{Lopez}, \binits{A.R.}},
\bauthor{\bsnm{Giro-i-Nieto}, \binits{X.}},
\bauthor{\bsnm{Burdick}, \binits{J.}},
\bauthor{\bsnm{Marques}, \binits{O.}}:
\bctitle{Skin lesion classification from dermoscopic images using deep learning techniques}.
In: \bbtitle{2017 13th IASTED International Conference on Biomedical Engineering (BioMed)},
pp. \bfpage{49}--\blpage{54}
(\byear{2017}).
\bcomment{IEEE}
\end{bchapter}
\endbibitem

\bibitem{li2018skin}
\begin{barticle}
\bauthor{\bsnm{Li}, \binits{Y.}},
\bauthor{\bsnm{Shen}, \binits{L.}}:
\batitle{Skin lesion analysis towards melanoma detection using deep learning network}.
\bjtitle{Sensors}
\bvolume{18}(\bissue{2}),
\bfpage{556}
(\byear{2018})
\end{barticle}
\endbibitem

\bibitem{yap2018multimodal}
\begin{barticle}
\bauthor{\bsnm{Yap}, \binits{J.}},
\bauthor{\bsnm{Yolland}, \binits{W.}},
\bauthor{\bsnm{Tschandl}, \binits{P.}}:
\batitle{Multimodal skin lesion classification using deep learning}.
\bjtitle{Experimental dermatology}
\bvolume{27}(\bissue{11}),
\bfpage{1261}--\blpage{1267}
(\byear{2018})
\end{barticle}
\endbibitem

\bibitem{liopyris2022artificial}
\begin{botherref}
\oauthor{\bsnm{Liopyris}, \binits{K.}},
\oauthor{\bsnm{Gregoriou}, \binits{S.}},
\oauthor{\bsnm{Dias}, \binits{J.}},
\oauthor{\bsnm{Stratigos}, \binits{A.J.}}:
Artificial intelligence in dermatology: Challenges and perspectives.
Dermatology and Therapy,
1--15
(2022)
\end{botherref}
\endbibitem

\bibitem{tschandl2018ham10000}
\begin{barticle}
\bauthor{\bsnm{Tschandl}, \binits{P.}},
\bauthor{\bsnm{Rosendahl}, \binits{C.}},
\bauthor{\bsnm{Kittler}, \binits{H.}}:
\batitle{The ham10000 dataset, a large collection of multi-source dermatoscopic images of common pigmented skin lesions}.
\bjtitle{Scientific data}
\bvolume{5}(\bissue{1}),
\bfpage{1}--\blpage{9}
(\byear{2018})
\end{barticle}
\endbibitem

\bibitem{combalia2019bcn20000}
\begin{botherref}
\oauthor{\bsnm{Combalia}, \binits{M.}},
\oauthor{\bsnm{Codella}, \binits{N.C.}},
\oauthor{\bsnm{Rotemberg}, \binits{V.}},
\oauthor{\bsnm{Helba}, \binits{B.}},
\oauthor{\bsnm{Vilaplana}, \binits{V.}},
\oauthor{\bsnm{Reiter}, \binits{O.}},
\oauthor{\bsnm{Carrera}, \binits{C.}},
\oauthor{\bsnm{Barreiro}, \binits{A.}},
\oauthor{\bsnm{Halpern}, \binits{A.C.}},
\oauthor{\bsnm{Puig}, \binits{S.}}, et al.:
Bcn20000: Dermoscopic lesions in the wild.
arXiv preprint arXiv:1908.02288
(2019)
\end{botherref}
\endbibitem

\bibitem{daneshjou2022disparities}
\begin{barticle}
\bauthor{\bsnm{Daneshjou}, \binits{R.}},
\bauthor{\bsnm{Vodrahalli}, \binits{K.}},
\bauthor{\bsnm{Novoa}, \binits{R.A.}},
\bauthor{\bsnm{Jenkins}, \binits{M.}},
\bauthor{\bsnm{Liang}, \binits{W.}},
\bauthor{\bsnm{Rotemberg}, \binits{V.}},
\bauthor{\bsnm{Ko}, \binits{J.}},
\bauthor{\bsnm{Swetter}, \binits{S.M.}},
\bauthor{\bsnm{Bailey}, \binits{E.E.}},
\bauthor{\bsnm{Gevaert}, \binits{O.}}, \betal:
\batitle{Disparities in dermatology ai performance on a diverse, curated clinical image set}.
\bjtitle{Science advances}
\bvolume{8}(\bissue{31}),
\bfpage{6147}
(\byear{2022})
\end{barticle}
\endbibitem

\bibitem{sun2016benchmark}
\begin{bchapter}
\bauthor{\bsnm{Sun}, \binits{X.}},
\bauthor{\bsnm{Yang}, \binits{J.}},
\bauthor{\bsnm{Sun}, \binits{M.}},
\bauthor{\bsnm{Wang}, \binits{K.}}:
\bctitle{A benchmark for automatic visual classification of clinical skin disease images}.
In: \bbtitle{European Conference on Computer Vision},
pp. \bfpage{206}--\blpage{222}
(\byear{2016}).
\bcomment{Springer}
\end{bchapter}
\endbibitem

\bibitem{han2018deep}
\begin{barticle}
\bauthor{\bsnm{Han}, \binits{S.S.}},
\bauthor{\bsnm{Park}, \binits{G.H.}},
\bauthor{\bsnm{Lim}, \binits{W.}},
\bauthor{\bsnm{Kim}, \binits{M.S.}},
\bauthor{\bsnm{Na}, \binits{J.I.}},
\bauthor{\bsnm{Park}, \binits{I.}},
\bauthor{\bsnm{Chang}, \binits{S.E.}}:
\batitle{Deep neural networks show an equivalent and often superior performance to dermatologists in onychomycosis diagnosis: Automatic construction of onychomycosis datasets by region-based convolutional deep neural network}.
\bjtitle{PloS one}
\bvolume{13}(\bissue{1}),
\bfpage{0191493}
(\byear{2018})
\end{barticle}
\endbibitem

\bibitem{harkemanneevaluation}
\begin{botherref}
\oauthor{\bsnm{Harkemanne}, \binits{E.}},
\oauthor{\bsnm{Legrand}, \binits{C.}},
\oauthor{\bsnm{Sawadogo}, \binits{K.}},
\oauthor{\bparticle{van} \bsnm{Maanen}, \binits{A.}},
\oauthor{\bsnm{Vossaert}, \binits{K.}},
\oauthor{\bsnm{Argenziano}, \binits{G.}},
\oauthor{\bsnm{Braun}, \binits{R.}},
\oauthor{\bsnm{Thomas}, \binits{L.}},
\oauthor{\bsnm{Baeck}, \binits{M.}},
\oauthor{\bsnm{Tromme}, \binits{I.}}:
Evaluation of primary care physicians’ competence in selective skin tumor triage after short versus long dermoscopy training a randomized non-inferiority trial.
Journal of the European Academy of Dermatology and Venereology
\end{botherref}
\endbibitem

\bibitem{cantisani2022melanoma}
\begin{barticle}
\bauthor{\bsnm{Cantisani}, \binits{C.}},
\bauthor{\bsnm{Ambrosio}, \binits{L.}},
\bauthor{\bsnm{Cucchi}, \binits{C.}},
\bauthor{\bsnm{Meznerics}, \binits{F.A.}},
\bauthor{\bsnm{Kiss}, \binits{N.}},
\bauthor{\bsnm{B{\'a}nv{\"o}lgyi}, \binits{A.}},
\bauthor{\bsnm{Rega}, \binits{F.}},
\bauthor{\bsnm{Grignaffini}, \binits{F.}},
\bauthor{\bsnm{Barbuto}, \binits{F.}},
\bauthor{\bsnm{Frezza}, \binits{F.}}, \betal:
\batitle{Melanoma detection by non-specialists: An untapped potential for triage?}
\bjtitle{Diagnostics}
\bvolume{12}(\bissue{11}),
\bfpage{2821}
(\byear{2022})
\end{barticle}
\endbibitem

\bibitem{du2020ai}
\begin{barticle}
\bauthor{\bsnm{Du-Harpur}, \binits{X.}},
\bauthor{\bsnm{Watt}, \binits{F.}},
\bauthor{\bsnm{Luscombe}, \binits{N.}},
\bauthor{\bsnm{Lynch}, \binits{M.}}:
\batitle{What is ai? applications of artificial intelligence to dermatology}.
\bjtitle{British Journal of Dermatology}
\bvolume{183}(\bissue{3}),
\bfpage{423}--\blpage{430}
(\byear{2020})
\end{barticle}
\endbibitem

\bibitem{tognetti2021impact}
\begin{barticle}
\bauthor{\bsnm{Tognetti}, \binits{L.}},
\bauthor{\bsnm{Cartocci}, \binits{A.}},
\bauthor{\bsnm{Cinotti}, \binits{E.}},
\bauthor{\bsnm{Moscarella}, \binits{E.}},
\bauthor{\bsnm{Farnetani}, \binits{F.}},
\bauthor{\bsnm{Lallas}, \binits{A.}},
\bauthor{\bsnm{Tiodorovic}, \binits{D.}},
\bauthor{\bsnm{Carrera}, \binits{C.}},
\bauthor{\bsnm{Longo}, \binits{C.}},
\bauthor{\bsnm{Puig}, \binits{S.}}, \betal:
\batitle{The impact of anatomical location and sun exposure on the dermoscopic recognition of atypical nevi and early melanomas: usefulness of an integrated clinical-dermoscopic method (idscore)}.
\bjtitle{Journal of the European Academy of Dermatology and Venereology}
\bvolume{35}(\bissue{3}),
\bfpage{650}--\blpage{657}
(\byear{2021})
\end{barticle}
\endbibitem

\bibitem{zadorozhny2023out}
\begin{bchapter}
\bauthor{\bsnm{Zadorozhny}, \binits{K.}},
\bauthor{\bsnm{Thoral}, \binits{P.}},
\bauthor{\bsnm{Elbers}, \binits{P.}},
\bauthor{\bsnm{Cin{\`a}}, \binits{G.}}:
\bctitle{Out-of-distribution detection for medical applications: Guidelines for practical evaluation}.
In: \bbtitle{Multimodal AI in Healthcare},
pp. \bfpage{137}--\blpage{153}.
\bpublisher{Springer}, \blocation{???}
(\byear{2023})
\end{bchapter}
\endbibitem

\bibitem{tschandl2020human}
\begin{barticle}
\bauthor{\bsnm{Tschandl}, \binits{P.}},
\bauthor{\bsnm{Rinner}, \binits{C.}},
\bauthor{\bsnm{Apalla}, \binits{Z.}},
\bauthor{\bsnm{Argenziano}, \binits{G.}},
\bauthor{\bsnm{Codella}, \binits{N.}},
\bauthor{\bsnm{Halpern}, \binits{A.}},
\bauthor{\bsnm{Janda}, \binits{M.}},
\bauthor{\bsnm{Lallas}, \binits{A.}},
\bauthor{\bsnm{Longo}, \binits{C.}},
\bauthor{\bsnm{Malvehy}, \binits{J.}}, \betal:
\batitle{Human--computer collaboration for skin cancer recognition}.
\bjtitle{Nature Medicine}
\bvolume{26}(\bissue{8}),
\bfpage{1229}--\blpage{1234}
(\byear{2020})
\end{barticle}
\endbibitem

\bibitem{brinker2018skin}
\begin{barticle}
\bauthor{\bsnm{Brinker}, \binits{T.J.}},
\bauthor{\bsnm{Hekler}, \binits{A.}},
\bauthor{\bsnm{Utikal}, \binits{J.S.}},
\bauthor{\bsnm{Grabe}, \binits{N.}},
\bauthor{\bsnm{Schadendorf}, \binits{D.}},
\bauthor{\bsnm{Klode}, \binits{J.}},
\bauthor{\bsnm{Berking}, \binits{C.}},
\bauthor{\bsnm{Steeb}, \binits{T.}},
\bauthor{\bsnm{Enk}, \binits{A.H.}},
\bauthor{\bsnm{Von~Kalle}, \binits{C.}}:
\batitle{Skin cancer classification using convolutional neural networks: systematic review}.
\bjtitle{Journal of medical Internet research}
\bvolume{20}(\bissue{10}),
\bfpage{11936}
(\byear{2018})
\end{barticle}
\endbibitem

\bibitem{moshi2019evaluation}
\begin{barticle}
\bauthor{\bsnm{Moshi}, \binits{M.R.}},
\bauthor{\bsnm{Parsons}, \binits{J.}},
\bauthor{\bsnm{Tooher}, \binits{R.}},
\bauthor{\bsnm{Merlin}, \binits{T.}}:
\batitle{Evaluation of mobile health applications: is regulatory policy up to the challenge?}
\bjtitle{International Journal of Technology Assessment in Health Care}
\bvolume{35}(\bissue{4}),
\bfpage{351}--\blpage{360}
(\byear{2019})
\end{barticle}
\endbibitem

\bibitem{mehta2022out}
\begin{bchapter}
\bauthor{\bsnm{Mehta}, \binits{D.}},
\bauthor{\bsnm{Gal}, \binits{Y.}},
\bauthor{\bsnm{Bowling}, \binits{A.}},
\bauthor{\bsnm{Bonnington}, \binits{P.}},
\bauthor{\bsnm{Ge}, \binits{Z.}}:
\bctitle{Out-of-distribution detection for long-tailed and fine-grained skin lesion images}.
In: \bbtitle{International Conference on Medical Image Computing and Computer-Assisted Intervention},
pp. \bfpage{732}--\blpage{742}
(\byear{2022}).
\bcomment{Springer}
\end{bchapter}
\endbibitem

\bibitem{salakhutdinov2012learning}
\begin{barticle}
\bauthor{\bsnm{Salakhutdinov}, \binits{R.}},
\bauthor{\bsnm{Tenenbaum}, \binits{J.B.}},
\bauthor{\bsnm{Torralba}, \binits{A.}}:
\batitle{Learning with hierarchical-deep models}.
\bjtitle{IEEE transactions on pattern analysis and machine intelligence}
\bvolume{35}(\bissue{8}),
\bfpage{1958}--\blpage{1971}
(\byear{2012})
\end{barticle}
\endbibitem

\bibitem{an2021hierarchical}
\begin{barticle}
\bauthor{\bsnm{An}, \binits{G.}},
\bauthor{\bsnm{Akiba}, \binits{M.}},
\bauthor{\bsnm{Omodaka}, \binits{K.}},
\bauthor{\bsnm{Nakazawa}, \binits{T.}},
\bauthor{\bsnm{Yokota}, \binits{H.}}:
\batitle{Hierarchical deep learning models using transfer learning for disease detection and classification based on small number of medical images}.
\bjtitle{Scientific reports}
\bvolume{11}(\bissue{1}),
\bfpage{4250}
(\byear{2021})
\end{barticle}
\endbibitem

\bibitem{zhang2017mixup}
\begin{botherref}
\oauthor{\bsnm{Zhang}, \binits{H.}},
\oauthor{\bsnm{Cisse}, \binits{M.}},
\oauthor{\bsnm{Dauphin}, \binits{Y.N.}},
\oauthor{\bsnm{Lopez-Paz}, \binits{D.}}:
mixup: Beyond empirical risk minimization.
arXiv preprint arXiv:1710.09412
(2017)
\end{botherref}
\endbibitem

\bibitem{yang2018robust}
\begin{bchapter}
\bauthor{\bsnm{Yang}, \binits{H.-M.}},
\bauthor{\bsnm{Zhang}, \binits{X.-Y.}},
\bauthor{\bsnm{Yin}, \binits{F.}},
\bauthor{\bsnm{Liu}, \binits{C.-L.}}:
\bctitle{Robust classification with convolutional prototype learning}.
In: \bbtitle{Proceedings of the IEEE Conference on Computer Vision and Pattern Recognition},
pp. \bfpage{3474}--\blpage{3482}
(\byear{2018})
\end{bchapter}
\endbibitem

\bibitem{thulasidasan2019mixup}
\begin{botherref}
\oauthor{\bsnm{Thulasidasan}, \binits{S.}},
\oauthor{\bsnm{Chennupati}, \binits{G.}},
\oauthor{\bsnm{Bilmes}, \binits{J.A.}},
\oauthor{\bsnm{Bhattacharya}, \binits{T.}},
\oauthor{\bsnm{Michalak}, \binits{S.}}:
On mixup training: Improved calibration and predictive uncertainty for deep neural networks.
Advances in Neural Information Processing Systems
\textbf{32}
(2019)
\end{botherref}
\endbibitem

\bibitem{zhang2020does}
\begin{botherref}
\oauthor{\bsnm{Zhang}, \binits{L.}},
\oauthor{\bsnm{Deng}, \binits{Z.}},
\oauthor{\bsnm{Kawaguchi}, \binits{K.}},
\oauthor{\bsnm{Ghorbani}, \binits{A.}},
\oauthor{\bsnm{Zou}, \binits{J.}}:
How does mixup help with robustness and generalization?
arXiv preprint arXiv:2010.04819
(2020)
\end{botherref}
\endbibitem

\bibitem{vaze2021open}
\begin{botherref}
\oauthor{\bsnm{Vaze}, \binits{S.}},
\oauthor{\bsnm{Han}, \binits{K.}},
\oauthor{\bsnm{Vedaldi}, \binits{A.}},
\oauthor{\bsnm{Zisserman}, \binits{A.}}:
Open-set recognition: A good closed-set classifier is all you need?
arXiv preprint arXiv:2110.06207
(2021)
\end{botherref}
\endbibitem

\bibitem{daneshjou2022checklist}
\begin{barticle}
\bauthor{\bsnm{Daneshjou}, \binits{R.}},
\bauthor{\bsnm{Barata}, \binits{C.}},
\bauthor{\bsnm{Betz-Stablein}, \binits{B.}},
\bauthor{\bsnm{Celebi}, \binits{M.E.}},
\bauthor{\bsnm{Codella}, \binits{N.}},
\bauthor{\bsnm{Combalia}, \binits{M.}},
\bauthor{\bsnm{Guitera}, \binits{P.}},
\bauthor{\bsnm{Gutman}, \binits{D.}},
\bauthor{\bsnm{Halpern}, \binits{A.}},
\bauthor{\bsnm{Helba}, \binits{B.}}, \betal:
\batitle{Checklist for evaluation of image-based artificial intelligence reports in dermatology: Clear derm consensus guidelines from the international skin imaging collaboration artificial intelligence working group}.
\bjtitle{JAMA dermatology}
\bvolume{158}(\bissue{1}),
\bfpage{90}--\blpage{96}
(\byear{2022})
\end{barticle}
\endbibitem

\bibitem{elmore2017pathologists}
\begin{botherref}
\oauthor{\bsnm{Elmore}, \binits{J.G.}},
\oauthor{\bsnm{Barnhill}, \binits{R.L.}},
\oauthor{\bsnm{Elder}, \binits{D.E.}},
\oauthor{\bsnm{Longton}, \binits{G.M.}},
\oauthor{\bsnm{Pepe}, \binits{M.S.}},
\oauthor{\bsnm{Reisch}, \binits{L.M.}},
\oauthor{\bsnm{Carney}, \binits{P.A.}},
\oauthor{\bsnm{Titus}, \binits{L.J.}},
\oauthor{\bsnm{Nelson}, \binits{H.D.}},
\oauthor{\bsnm{Onega}, \binits{T.}}, et al.:
Pathologists’ diagnosis of invasive melanoma and melanocytic proliferations: observer accuracy and reproducibility study.
bmj
\textbf{357}
(2017)
\end{botherref}
\endbibitem

\bibitem{felmingham2022improving}
\begin{barticle}
\bauthor{\bsnm{Felmingham}, \binits{C.}},
\bauthor{\bsnm{MacNamara}, \binits{S.}},
\bauthor{\bsnm{Cranwell}, \binits{W.}},
\bauthor{\bsnm{Williams}, \binits{N.}},
\bauthor{\bsnm{Wada}, \binits{M.}},
\bauthor{\bsnm{Adler}, \binits{N.R.}},
\bauthor{\bsnm{Ge}, \binits{Z.}},
\bauthor{\bsnm{Sharfe}, \binits{A.}},
\bauthor{\bsnm{Bowling}, \binits{A.}},
\bauthor{\bsnm{Haskett}, \binits{M.}}, \betal:
\batitle{Improving skin cancer management with artificial intelligence (smarti): protocol for a preintervention/postintervention trial of an artificial intelligence system used as a diagnostic aid for skin cancer management in a specialist dermatology setting}.
\bjtitle{BMJ open}
\bvolume{12}(\bissue{1}),
\bfpage{050203}
(\byear{2022})
\end{barticle}
\endbibitem

\bibitem{he2016deep}
\begin{bchapter}
\bauthor{\bsnm{He}, \binits{K.}},
\bauthor{\bsnm{Zhang}, \binits{X.}},
\bauthor{\bsnm{Ren}, \binits{S.}},
\bauthor{\bsnm{Sun}, \binits{J.}}:
\bctitle{Deep residual learning for image recognition}.
In: \bbtitle{Proceedings of the IEEE Conference on Computer Vision and Pattern Recognition},
pp. \bfpage{770}--\blpage{778}
(\byear{2016})
\end{bchapter}
\endbibitem

\bibitem{carion2020end}
\begin{bchapter}
\bauthor{\bsnm{Carion}, \binits{N.}},
\bauthor{\bsnm{Massa}, \binits{F.}},
\bauthor{\bsnm{Synnaeve}, \binits{G.}},
\bauthor{\bsnm{Usunier}, \binits{N.}},
\bauthor{\bsnm{Kirillov}, \binits{A.}},
\bauthor{\bsnm{Zagoruyko}, \binits{S.}}:
\bctitle{End-to-end object detection with transformers}.
In: \bbtitle{European Conference on Computer Vision},
pp. \bfpage{213}--\blpage{229}
(\byear{2020}).
\bcomment{Springer}
\end{bchapter}
\endbibitem

\bibitem{parmar2018image}
\begin{bchapter}
\bauthor{\bsnm{Parmar}, \binits{N.}},
\bauthor{\bsnm{Vaswani}, \binits{A.}},
\bauthor{\bsnm{Uszkoreit}, \binits{J.}},
\bauthor{\bsnm{Kaiser}, \binits{L.}},
\bauthor{\bsnm{Shazeer}, \binits{N.}},
\bauthor{\bsnm{Ku}, \binits{A.}},
\bauthor{\bsnm{Tran}, \binits{D.}}:
\bctitle{Image transformer}.
In: \bbtitle{International Conference on Machine Learning},
pp. \bfpage{4055}--\blpage{4064}
(\byear{2018}).
\bcomment{PMLR}
\end{bchapter}
\endbibitem

\bibitem{bello2019attention}
\begin{bchapter}
\bauthor{\bsnm{Bello}, \binits{I.}},
\bauthor{\bsnm{Zoph}, \binits{B.}},
\bauthor{\bsnm{Vaswani}, \binits{A.}},
\bauthor{\bsnm{Shlens}, \binits{J.}},
\bauthor{\bsnm{Le}, \binits{Q.V.}}:
\bctitle{Attention augmented convolutional networks}.
In: \bbtitle{Proceedings of the IEEE/CVF International Conference on Computer Vision},
pp. \bfpage{3286}--\blpage{3295}
(\byear{2019})
\end{bchapter}
\endbibitem

\bibitem{kong2020sde}
\begin{botherref}
\oauthor{\bsnm{Kong}, \binits{L.}},
\oauthor{\bsnm{Sun}, \binits{J.}},
\oauthor{\bsnm{Zhang}, \binits{C.}}:
Sde-net: Equipping deep neural networks with uncertainty estimates.
arXiv preprint arXiv:2008.10546
(2020)
\end{botherref}
\endbibitem

\bibitem{semsarian2022we}
\begin{barticle}
\bauthor{\bsnm{Semsarian}, \binits{C.R.}},
\bauthor{\bsnm{Ma}, \binits{T.}},
\bauthor{\bsnm{Nickel}, \binits{B.}},
\bauthor{\bsnm{Scolyer}, \binits{R.A.}},
\bauthor{\bsnm{Ferguson}, \binits{P.M.}},
\bauthor{\bsnm{Soyer}, \binits{H.P.}},
\bauthor{\bsnm{Parker}, \binits{L.}},
\bauthor{\bsnm{Barratt}, \binits{A.}},
\bauthor{\bsnm{Thompson}, \binits{J.F.}},
\bauthor{\bsnm{Bell}, \binits{K.J.}}:
\batitle{Do we need to rethink the diagnoses melanoma in situ and severely dysplastic naevus?}
\bjtitle{British Journal of Dermatology}
\bvolume{186}(\bissue{6}),
\bfpage{1030}--\blpage{1032}
(\byear{2022})
\end{barticle}
\endbibitem

\bibitem{groh2021evaluating}
\begin{bchapter}
\bauthor{\bsnm{Groh}, \binits{M.}},
\bauthor{\bsnm{Harris}, \binits{C.}},
\bauthor{\bsnm{Soenksen}, \binits{L.}},
\bauthor{\bsnm{Lau}, \binits{F.}},
\bauthor{\bsnm{Han}, \binits{R.}},
\bauthor{\bsnm{Kim}, \binits{A.}},
\bauthor{\bsnm{Koochek}, \binits{A.}},
\bauthor{\bsnm{Badri}, \binits{O.}}:
\bctitle{Evaluating deep neural networks trained on clinical images in dermatology with the fitzpatrick 17k dataset}.
In: \bbtitle{Proceedings of the IEEE/CVF Conference on Computer Vision and Pattern Recognition},
pp. \bfpage{1820}--\blpage{1828}
(\byear{2021})
\end{bchapter}
\endbibitem

\bibitem{shimodaira2000improving}
\begin{barticle}
\bauthor{\bsnm{Shimodaira}, \binits{H.}}:
\batitle{Improving predictive inference under covariate shift by weighting the log-likelihood function}.
\bjtitle{Journal of statistical planning and inference}
\bvolume{90}(\bissue{2}),
\bfpage{227}--\blpage{244}
(\byear{2000})
\end{barticle}
\endbibitem

\bibitem{fogelberg2023domain}
\begin{botherref}
\oauthor{\bsnm{Fogelberg}, \binits{K.}},
\oauthor{\bsnm{Chamarthi}, \binits{S.}},
\oauthor{\bsnm{Maron}, \binits{R.C.}},
\oauthor{\bsnm{Niebling}, \binits{J.}},
\oauthor{\bsnm{Brinker}, \binits{T.J.}}:
Domain shifts in dermoscopic skin cancer datasets: Evaluation of essential limitations for clinical translation.
New Biotechnology
(2023)
\end{botherref}
\endbibitem

\end{thebibliography}



\end{document}